\documentclass{article}

\usepackage[main, final]{neurips_2026}

\usepackage[utf8]{inputenc} % allow utf-8 input
\usepackage[T1]{fontenc}    % use 8-bit T1 fonts
\usepackage{hyperref}       % hyperlinks
\usepackage{url}            % simple URL typesetting
\usepackage{booktabs}       % professional-quality tables
\usepackage{amsfonts}       % blackboard math symbols
\usepackage{nicefrac}       % compact symbols for 1/2, etc.
\usepackage{microtype}      % microtypography
\usepackage{xcolor}         % colors
\usepackage{xspace}
\usepackage{graphicx}
\usepackage{multirow}
\usepackage{subcaption}
\usepackage{amsmath}
\usepackage{tcolorbox}
\usepackage[ruled,vlined,linesnumbered]{algorithm2e}
\usepackage{wrapfig}
\usepackage{tabularx}
\usepackage{float}
\usepackage{listings}
\usepackage{xcolor}
\usepackage{inconsolata}

\newtcolorbox{promptbox}{
    colback=gray!5,       % 背景色：极浅灰色
    colframe=gray!50,     % 边框色：中灰色
    arc=1mm,              % 圆角
    boxrule=0.5pt,        % 边框厚度
    left=1.5mm,           % 左内边距
    right=1.5mm,          % 右内边距
    top=1mm,              % 上内边距
    bottom=1mm,           % 下内边距
    boxsep=0pt,           % 额外间距归零，实现最 tight 效果
    fontupper=\small\ttfamily, % 使用小号等宽字体
}

\newcommand{\techname}{HEAL\xspace}

\title{Demystifying Numerical Instability in LLM Inference: Achieving Reproducible Inference for Mission-Critical Tasks with \underline{\techname}}

\author{
Zhenting Zhu$^{1}$,\enspace
Lucas Thai$^{1}$,\enspace
Shan Yu$^{1}$,\enspace
Yicheng Liu$^{1}$,\enspace
Yifan Qiao$^{2}$,\enspace
Chenxi Wang$^{3}$,\enspace \\
\textbf{Harry Xu}$^{1}$,\enspace
\textbf{Junyi Shu}$^{1}$\\[0.5ex]
$^1$UCLA \quad
$^2$UC Berkeley \quad
$^3$University of Chinese Academy of Sciences 
}
\newif\ifdraft
\draftfalse

\microtypecontext{spacing=nonfrench}
\newcommand{\squishlist}{
   \begin{list}{$\bullet$}
    { \setlength{\itemsep}{0pt}      \setlength{\parsep}{3pt}
      \setlength{\topsep}{3pt}       \setlength{\partopsep}{0pt}
      \setlength{\leftmargin}{1.0em} \setlength{\labelwidth}{1em}
      \setlength{\labelsep}{0.5em} } }
\newcommand{\squishend}{
    \end{list}  }
    
\definecolor{javared}{rgb}{0.6,0,0} % for strings
\definecolor{javagreen}{rgb}{0.25,0.5,0.35} % comments
\definecolor{javapurple}{rgb}{0.5,0,0.35} % keywords
\definecolor{javadocblue}{rgb}{0.25,0.35,0.75} % javadoc

\newcommand{\naive}[0]{na\"{i}ve\xspace}

\newcommand{\codeIn}[1]{{\small\texttt{#1}}}

\newcommand{\MyPara}[1]{\vspace{.1em}\noindent\textbf{#1}~}
\newcommand{\captionfonts}{\small}

\makeatletter  % Allow the use of @ in command names
\long\def\@makecaption#1#2{%
  \vskip\abovecaptionskip
  \sbox\@tempboxa{{\captionfonts #1: #2}}%
  \ifdim \wd\@tempboxa >\hsize
    {\captionfonts #1: #2\par}
  \else
    \hbox to\hsize{\hfil\box\@tempboxa\hfil}%
  \fi
  \vskip\belowcaptionskip}
\makeatother   % Cancel the effect of \makeatletter

\newcommand{\squishlistree}{
   \begin{list}{$\bullet$}
    { \setlength{\itemsep}{0pt}      \setlength{\parsep}{0pt}
      \setlength{\topsep}{3pt}       \setlength{\partopsep}{0pt}
      \setlength{\leftmargin}{1em} \setlength{\labelwidth}{1em}
      \setlength{\labelsep}{0.5em} } }

\newcommand{\squishlisttwo}{
   \begin{list}{$\bullet$}
    { \setlength{\itemsep}{0pt}    \setlength{\parsep}{0pt}
      \setlength{\topsep}{0pt}     \setlength{\partopsep}{0pt}
      \setlength{\leftmargin}{2em} \setlength{\labelwidth}{1.5em}   
      \setlength{\labelsep}{0.5em} } }

\newcommand{\eg}{\hbox{\emph{e.g.}}\xspace}

\makeatletter
\newcommand*{\circled}{\@ifstar\circledstar\circlednostar}
\makeatother
 
\newcommand*\circledstar[1]{%
   \tikz[baseline=(C.base)]
     \node[%
       fill=black!20,
       circle,
       minimum size=1em,
       text=black,
       font=\footnotesize,
       inner sep=0.3pt
     ](C) {#1};%
}
\newcommand*\circlednostar[1]{%
   \tikz[baseline=(C.base) - .6em]
     \node[%
       fill=black,
       text=white,
       circle,
       minimum size=.8em,
       font={\bf \footnotesize},
       inner sep=0.2pt
     ](C) {#1};%
}

\definecolor{softred}{RGB}{240,70,70}
\definecolor{softgreen}{RGB}{130,220,130}
\definecolor{softyellow}{RGB}{255, 230, 120}
\begin{document}

\maketitle

\begin{abstract}
As Large Language Models~(LLMs) deploy into mission-critical domains~(\eg, finance, medicine, and law), output reproducibility has become a strict system requirement. While practitioners use greedy decoding to eliminate algorithmic stochasticity, empirical deployments with 16-bit precisions still exhibit catastrophic output divergence across heterogeneous GPUs. Through SASS-level profiling, we reveal that this inconsistency is fundamentally driven by truncation errors introduced during downcasting at kernel boundaries. 
% Because extreme numerical overflows are rare during inference, output divergence is almost entirely caused by these microscopic losses of fractional precision; thus, protecting the significand is the true key to deterministic outputs.
However, achieving reproducibility via a global FP32 pipeline incurs prohibitive system penalties: bypassing 16-bit hardware accelerators hurts compute efficiency, while upcasting the KV cache doubles memory overhead. To bridge this gap, we propose \textbf{H}ybrid \textbf{E}rror \textbf{AL}leviation~(\textbf{\techname}), a targeted intervention that approximates FP32 precision while resolving hardware constraints through two targeted mechanisms. 
First, recognizing that floating-point formats underutilize their bit-width for Q, K, V tensors, \techname applies INT16 quantization that preserves numerical stability without expanding the KV cache footprint.
Second, \techname synthesizes high-precision matrix multiplications via an algebraic error compensation strategy, executing entirely on high-throughput 16-bit Tensor Cores.
To evaluate our approach practically, we introduce MCR-Bench, a benchmark targeting reproducibility in mission-critical tasks. \techname achieves the same level of reproducibility on downstream tasks as the FP32 baseline 
% \zt{with as low as 13\% (to-verify) overhead} \junyi{compare both reproducibility and overhead to FP32} 
while reducing the performance overhead by up to 6.0$\times$.
\end{abstract}
\section{Introduction}
\label{sec:intro}

As Large Language Models~(LLMs) transition from creative assistants to the backbone of mission-critical applications\textemdash such as enterprise-grade services~\cite{ibm-redbooks, coveo, xu2024llm4workflow}, finance~\cite{star, khatchadourian2025llmoutputdriftcrossprovider}, medical diagnosis~\cite{cardamone2025classifying, gim2026llm, ullah2024challenges}, and regulatory legal compliance~\cite{blairstanek2025llmsprovideunstableanswers,  song2026evaluatingimplicitregulatorycompliance, yue2024lawllm}\textemdash the primary requirement shifts from mere fluency to strict system reliability. In these high-stakes workflows, a fundamental expectation is reproducibility: a fixed input must yield a functionally identical outcome. To achieve this, practitioners typically employ greedy decoding~(temperature=0), a setting universally assumed to remove stochasticity from the generation process.

However, the infrastructure supporting these models makes this guarantee exceedingly difficult to uphold. Modern datacenters are inherently heterogeneous, frequently mixing different generations and vendors of accelerators due to rapid procurement cycles and massive compute demands. For instance, older generations of NVIDIA GPUs remain ubiquitous~\cite{nvidia-a100, aws-a100}, while leading model providers increasingly integrate non-NVIDIA hardware to scale capacity and mitigate vendor lock-in~\cite{ anthropic-aws, anthropic-google, openai-amd, openai-aws}. Consequently, LLM inference remains non-deterministic even under greedy decoding. When identical inference code executes across varied hardware, the non-associativity of floating-point arithmetic\textemdash combined with dynamically shifting software configurations~(\eg, varying attention backends or batch sizes)\textemdash causes generated tokens to diverge~\cite{he2025defeatingnondeterminism, coveo, lmsys2025sglang}. In domains requiring exact numerical outputs~(\eg, specific ICD-10 medical codes~\cite{icd}, or boolean legal determinations), a single divergent token can invalidate the entire semantic pipeline. 
% \shan{Shall we highlight the cross-hardware reproducibility as the problem setting of this paper? I think it is a unique problem that our paper focuses on.}

To secure the practical, task-level reproducibility required by mission-critical applications, existing mitigation strategies attempt to enforce strict bit-wise determinism\textemdash ensuring every generated token matches exactly across runs. These approaches generally fall into two categories, both of which present an unacceptable dilemma for live production environments. One line of work enforces deterministic floating-point reduction orders, through framework-level deterministic modes, hardware-specific features such as CUDA backward compatibility, or specialized deterministic kernels and inference engines~\cite{demmel2016efficient, he2025defeatingnondeterminism, ingonyama2024llmreproducibility, pytorch-reproducibility, lmsys2025sglang, zhang2025deterministicinferencetensorparallel}. While these methods achieve bit-level consistency, they restrict deployment to narrow execution paths, disable modern hardware features, sacrifice throughput, and fail to generalize across disparate GPU vendors or generations. An alternative approach seeks to suppress numerical instability through higher-precision computation, performing all calculations in FP32~\cite{yuan2025understanding}. Although this approach effectively eliminates downcasting noise and is theoretically generalizable, practical barriers make its adoption prohibitively expensive. Serving entirely in FP32 severely degrades generation speed, increasing the Time Per Output Token (TPOT) by up to 13.1$\times$, rendering it economically impractical.

% Serving entirely in FP32 reduces inference throughput to roughly one-ninth of peak performance, rendering it economically impractical.

\MyPara{Key Insight.} This work challenges the assumption that the all-or-nothing cost of full high-precision pipelines is truly required to achieve robust, deployment-ready reproducibility. To systematically investigate the root causes of non-reproducibility and the structural barriers to efficient mitigation, we establish comprehensive baselines across standard precision formats. Through SASS-level profiling, we find that instability is not driven by internal kernel arithmetic~(which predominantly operates in FP32), but by \emph{the truncation of high-precision registers during downcasting at kernel boundaries}. Because extreme numerical overflows are rare during inference, it is these microscopic losses of fractional precision that accumulate over the network depth. Eventually, this end-to-end truncation error shifts the final log probability enough to alter the top-1 token selection, directly causing non-reproducibility.

Crucially, our analysis reveals that we do not need strict bitwise parity to achieve reproducibility.
% \shan{We could define functional reproducibility more clearly at the beginning. Reviewers may confuse bitwise determinism, token-level determinism, answer-level reproducibility, and semantic/task-level reproducibility.}; 
Instead, we only need to bound numerical noise below the threshold that triggers result flipping. However, \naive attempts to enforce this bounded precision through FP32 upcasting run into two fatal hardware walls. 
% \shan{Logic gap: bounding numerical noise does not directly imply FP32 upcasting.}
First, for \codeIn{Attention}, scaling precision introduces a memory bottleneck. Upcasting the Key-Value~(KV) cache to 32-bit strictly doubles its memory footprint, severely degrading the performance.
Second, for \codeIn{GEMM}, scaling up to higher precision triggers a compute bottleneck. Modern AI accelerators achieve high throughput by relying on specialized hardware units~(\eg, Tensor Cores) that structurally require 16-bit inputs. Upcasting inputs to FP32 to preserve precision forces the hardware to fall back to standard execution units, causing a massive performance regression.

\MyPara{\techname.} To resolve this fundamental tradeoff between reproducibility and hardware efficiency, we propose \textbf{H}ybrid \textbf{E}rror \textbf{AL}leviation~(\textbf{\techname}), a targeted intervention that achieves FP32-level reproducibility while effectively bypassing both the \codeIn{Attention} and \codeIn{GEMM} bottlenecks. 
To resolve the \codeIn{Attention} bottleneck, \techname avoids full FP32 cache upcast. Recognizing that floating-point formats underutilize their bit-width for Q, K, V tensors, \techname applies a targeted INT16 quantization. This achieves the necessary fractional precision to prevent divergence without expanding the physical KV cache footprint.
To address the \codeIn{GEMM} bottleneck, \techname synthesizes high-precision matrix multiplications without bypassing 16-bit hardware accelerators. It employs an algebraic error compensation strategy, decomposing the input activations into a primary 16-bit value and a residual error term. By evaluating the matrix multiplication for both components using standard 16-bit inputs, the system captures the truncation error while executing entirely on high-throughput Tensor Cores.

% First, to resolve the compute bottleneck~(Observation 2), \techname synthesizes high-precision matrix multiplications without bypassing 16-bit hardware accelerators. It employs an algebraic error compensation strategy, decomposing the input activations into a primary 16-bit value and a residual error term. By evaluating the matrix multiplication for both components using standard 16-bit inputs, the system captures the truncation error while executing entirely on high-throughput Tensor Cores. 
% \ztnote{Do we need to mention this also make the FP16 LLM pipeline more reliable?} \junyi{you can elaborate around this in the main body but not here} 
% Second, to resolve the memory bottleneck~(Observation 3), \techname avoids an FP32 cache upcast entirely. Recognizing that the error-sensitive V-tensor contains significantly fewer outliers than other activations, \techname applies a targeted INT16 quantization. This achieves the necessary precision to prevent divergence without expanding the KV cache footprint.

To evaluate our approach practically, we introduce MCR-Bench~(Mission-Critical Reproducibility Benchmark), a benchmark targeting reproducibility in domains where outcome consistency is paramount. Our evaluation across a diverse spectrum of modern LLMs demonstrates that \techname achieves the same level of downstream task reproducibility as the FP32 baseline. With our hybrid error alleviation strategy, \techname incurs only a 70\%--120\% TPOT overhead, a stark contrast to the severe 4.2--13.1$\times$ slowdown exhibited by the FP32 pipeline~\cite{yuan2025understanding} on various GPU and model types.

% To evaluate our approach practically, we introduce MCR-Bench~(Mission-Critical Reproducibility Benchmark), a benchmark targeting reproducibility in domains where outcome consistency is paramount. Our evaluation across a diverse spectrum of modern LLMs demonstrates that \techname achieves the same level of downstream task reproducibility as the FP32 baseline. With our hybrid error alleviation strategy, \techname incurs only a 91.5\% \todoinline{update with range} TPOT overhead, a stark contrast to the nearly 16.8$\times$ \todoinline{update with range} performance degradation of the FP32 pipeline~\cite{yuan2025understanding} on various GPU types.

% To evaluate SPU beyond theoretical numerical errors, we introduce \textit{MCR-Bench}~(Mission-Critical Reproducibility Benchmark), targeting domains where outcome consistency is paramount. Our comprehensive evaluation across a diverse spectrum of modern LLMs\textemdash including Qwen3, Llama-3.1, and DeepSeek-R1-Distill, ranging from 7B to 32B parameters\textemdash demonstrates that SPU achieves practically equivalent downstream stability and prevents semantic divergence.

In summary, this paper makes the following contributions:
% \begin{itemize}
% \item We demonstrate that intrinsic non-reproducibility across heterogeneous hardware is driven by BF16 downcasting and super-linearly amplified by structural ``heavy-hitter'' outliers.
% \item We introduce Selective Precision Upcasting~(SPU), which applies FP32 exclusively to <1\% of sensitivity-critical values, establishing the optimal Pareto frontier between numerical stability and hardware efficiency.
% \item We introduce MCR-Bench, demonstrating that SPU achieves near-FP32 reliability for mission-critical downstream tasks across varied hardware architectures with only a 24.7\% performance overhead.
% \end{itemize}
\squishlist
\item We demonstrate that the intrinsic non-reproducibility is fundamentally driven by boundary truncation errors, and establish that robust reproducibility requires bounding this noise below a result flipping threshold~(\S\ref{sec:anatomy}).
\item We introduce Hybrid Error ALleviation~(\techname), a two-pronged solution that bypasses hardware penalties by safeguarding \codeIn{Attention} via INT16 quantization and synthesizing high-precision \codeIn{GEMM} via algebraic error compensation~(\S\ref{sec:solution}).
\item We introduce MCR-Bench, demonstrating that \techname achieves FP32-level reproducibility for mission-critical downstream tasks across varied hardware types while reducing the performance overhead by up to 6.0$\times$~(\S\ref{sec:eval}).
\squishend

The source code of \techname and the MCR-Bench data are publicly available at \url{https://github.com/ucla-systems/HEAL}.
\section{The Anatomy of Non-reproducibility}
\label{sec:anatomy}

\begin{wrapfigure}{R}{0.5\linewidth}
%\vspace{-1.5em}
\centering
% \begin{figure}% \vskip 0.2in
% \begin{center}
% \vspace{-2em}
\includegraphics[width=\linewidth]{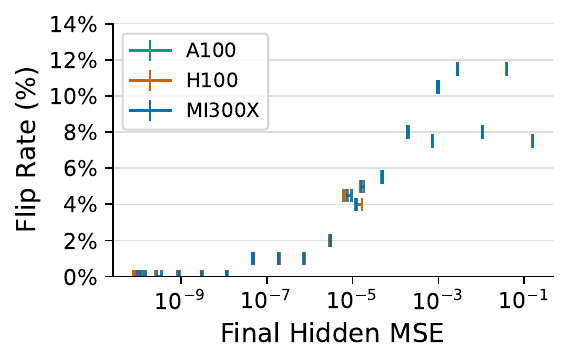}
% \vspace{-2em}
\caption{Answer flip rate vs. total numerical error.}
\label{fig:error-vs-flip}
% \end{center}
% \end{figure}
% \vspace{-1em}
\end{wrapfigure}

To understand why standard 16-bit deployments~(\eg, BF16) fail to provide the deterministic reproducibility required by mission-critical applications, we must isolate the structural mechanisms driving end-to-end divergence. In this section, we systematically trace the lifecycle of numerical errors in reverse: starting from the macroscopic failure where final log-probabilities shift enough to change the top-1 token, tracing back to observe how these errors accumulate and propagate across network layers, and finally isolating their microscopic origin at the hardware level.

To ensure our findings represent fundamental systemic behaviors rather than artifacts of a specific architecture, our diagnostic analysis spans five widely deployed open-source models\footnote{This section presents the results for Qwen3-14B. The results for the remaining models are detailed in Appendix~\ref{app:study}.}: Qwen3-8B, 14B and 32B~\cite{yang2025qwen3}, Llama-3.1-8B-Instruct~\cite{grattafiori2024llama}, and DeepSeek-R1-Distill-Qwen-14B~\cite{guo2025deepseek}. Furthermore, to measure numerical error accurately without the confounding effects of autoregressive branching, we enforce strict execution path alignment via Teacher Forcing~\cite{williams1989learning}. Specifically, we utilize an NVIDIA H100 executing a full FP32 pipeline as our deterministic anchor, tracking the layer-by-layer divergence of low-precision executions against this anchor.\footnote{For more details on the experiment setup, please refer to Appendix~\ref{app:setup} and \ref{app:study}.}

\subsection{The Macroscopic Failure: Result Flipping}

We begin at the end of the inference pipeline. Because strict reproducibility is a baseline requirement for mission-critical applications, these systems naturally default to greedy decoding. In this deterministic setting, the model's final decision relies on a simple argmax operation over the output logits: $\text{token} = \arg\max(p(t))$. Therefore, numerical error only manifests as non-reproducibility if it accumulates enough magnitude to bridge the probability margin between the top-1 and top-2 tokens.

Crucially, this structural reliance on argmax creates a highly non-linear relationship between the accumulated numerical noise and the likelihood of result flipping. The network exhibits a distinct \textit{divergence threshold}. To quantify the accumulated noise just before the final decision, we measure the Final Hidden MSE, defined as the mean squared error in the representations immediately before the vocabulary projection matrix (i.e., the normalized final hidden layer output). As shown in Figure~\ref{fig:error-vs-flip}, which tracks the probability of the choice being flipped on MedQA~\cite{jin2021disease} against this Final Hidden MSE, we observe two distinct operational regimes:

% We begin at the end of the inference pipeline. In mission-critical applications utilizing greedy decoding, 
% \ztnote{there are some gaps here} 
% the model's final decision relies on a simple argmax operation over the output logits: $\text{token} = \arg\max(p(t))$. Therefore, numerical error only manifests as non-reproducibility if it accumulates enough magnitude to bridge the probability margin between the top-1 and top-2 tokens.

% Crucially, this structural reliance on argmax creates a highly non-linear relationship between the accumulated numerical noise and the likelihood of result flipping. The network exhibits a distinct \textit{divergence threshold}. As shown in Figure~\ref{fig:error-vs-flip} \todoinline{clarify final hidden definition}, which tracks the probability of the choice being flipped on MedQA\cite{jin2021disease} against the magnitude of the logit error, we observe two distinct operational regimes:

\squishlist
\item \textbf{The Benign Regime~(Sub-Threshold):} When the accumulated error remains small, it is safely absorbed by the inherent probability margin between the top candidates. The downstream mission-critical logic remains entirely intact, demonstrating the model's natural tolerance to minor numerical fluctuations. 
% \ztnote{I am just a little bit uncertain because I observed cases where there is a 1e-5 difference between two choices, but I would say from what I observed, the semantic meaning is the same.}
\item \textbf{The Catastrophic Regime~(Super-Threshold):} However, once the accumulated error breaches a critical threshold, the probability of a choice flip surges dramatically. When internal errors are heavily amplified, they inject massive absolute deviations that overcome the probability margin. This results in catastrophic divergence, such as misdiagnosing a life-threatening acute condition as a benign symptom due to shifting to a distractor option.
\squishend

% ZT: This illustration is not that clear or correct, fix it in ArXiv version.

Consequently, this non-linear dynamic dictates that strict bitwise parity with an FP32 reference is an unnecessarily strict system requirement. To guarantee downstream functional reproducibility, it is sufficient to maintain the accumulated numerical errors safely within the benign, sub-threshold regime.

% \begin{figure}[!t]
% % \vskip 0.2in
% \begin{center}
% \includegraphics[width=0.85\columnwidth]{figures/neurips/error_vs_flip_rate_errorbars__Qwen-Qwen3-14B.pdf}
% \todoinline{Insert plot here: X-axis = total error, Y-axis = probability of answer flip. \ztnote{Running}}
% \caption{The non-linear relationship between numerical error and reproducibility. Minor errors are safely absorbed by the probability margin (benign regime), but once the error exceeds a critical threshold, the probability of a catastrophic answer flip increases dramatically.}
% \label{fig:error-vs-flip}
% \end{center}
% \end{figure}

\subsection{The Propagation and Accumulation of Error}
\label{sec:propagation-error}

If the network can naturally tolerate minor background noise, how does the numerical error accumulate the magnitude required to breach the catastrophic divergence threshold? To answer this, we abstract the standard LLM inference procedure and track the propagation of error layer-by-layer.

\begin{wrapfigure}{R}{0.61\linewidth}
% \vspace{-1.5em}
% \begin{minipage}{0.6\textwidth}
\begin{algorithm}[H]
\LinesNotNumbered
\caption{Qwen3CausalLM Forward Pass}
\label{alg:vllm_forward_compact}
\footnotesize
\SetAlgoLined
\KwIn{Tokens $\mathbf{T}$, KV Caches $\mathbf{C}$, Weights $\mathbf{W}$, Metadata $\mathcal{M}$}
\KwOut{Next tokens $\mathbf{T}_{next}$}

$\mathbf{H}_0 \leftarrow \text{Embedding}(\mathbf{T})$\;

\For{$i = 0$ \KwTo $L-1$}{
    $\mathrm{q}, \mathrm{k}, \mathbf{V}_i \leftarrow \text{GEMM}\left(\text{Norm}(\mathbf{H_i}), \mathbf{W}_{qkv}^{(i)}\right) $\;
    $\mathbf{Q}_i, \mathbf{K}_i \leftarrow \text{RoPE}\left(\text{QKNorm}(\mathrm{q}, \mathrm{k})\right)$; $\mathbf{C}^{(i)} \leftarrow \mathbf{C}^{(i)} + \left\{ \mathbf{K}_i, \mathbf{V}_i \right\}$\;
    $\mathbf{A}_i \leftarrow \mathbf{H}_i + \text{GEMM}\left(\text{Attention}(\mathbf{Q}_i, \mathbf{C}^{(i)}, \mathcal{M}), \mathbf{W}_o^{(i)}\right)$\;
    $\mathrm{g}, \mathrm{u} \leftarrow \text{GEMM}\left(\text{Norm}(\mathbf{A}_i), \mathbf{W}_{gate\_up}^{(i)}\right)$\;
    $\mathbf{H}_{i+1} \leftarrow \mathbf{A}_i + \text{GEMM}\left(\text{SiLU}(\mathrm{g}) \odot \mathrm{u}, \mathbf{W}_{down}^{(i)}\right)$\;
}

$\mathbf{L} \leftarrow \text{GEMM}(\text{ExtractLastTokens}(\text{Norm}(\mathbf{H}_L), \mathcal{M}), \mathbf{W}_{lm\_head})$\;
\Return{$\text{Sample}(\mathbf{L})$}\;
\end{algorithm}
% \end{minipage}
% \vspace{-1.5em}
\end{wrapfigure}

% \begin{algorithm}
% \caption{Qwen3 CausalLM Forward Pass}
% \label{alg:vllm_forward_compact}
% \SetAlgoLined
% \KwIn{Tokens $\mathbf{T}$, KV Caches $\mathbf{C}$, Weight $\mathbf{W}$, Metadata $\mathcal{M}$}
% \KwOut{Next tokens $\mathbf{T}_{next}$}

% $\mathbf{H}_0 \leftarrow \text{Embedding}(\mathbf{T})$\;

% \For{$i = 0$ \KwTo $L-1$}{
%     $\mathbf{Q}_i, \mathbf{K}_i, \mathbf{V}_i \leftarrow \text{QKNorm}\left(\text{GEMM}(\text{Norm}(\mathbf{H_i}), \mathbf{W}_{qkv}^{(i)})\right)$\;
%     $\mathbf{C}^{(i)} \leftarrow \mathbf{C}^{(i)} + \left\{ \mathbf{K}_i, \mathbf{V}_i \right\}$; $\mathbf{A}_i \leftarrow \mathbf{H}_i + \text{Attention}(\mathbf{Q}_i, \mathbf{C}^{(i)}, \mathcal{M}, \mathbf{W}_o^{(i)})$\;
%     $\mathbf{H}_{i+1} \leftarrow \mathbf{A}_i + \text{MLP}(\text{Norm}(\mathbf{A}_i), \mathbf{W}_{gate\_up}^{(i)}, \mathbf{W}_{down}^{(i)})$\;
% }

% $\mathbf{L} \leftarrow \text{GEMM}(\text{ExtractLastTokens}(\text{Norm}(\mathbf{H}_L), \mathcal{M}), \mathbf{W}_{lm\_head})$\;
% \Return{$\text{Sample}(\mathbf{L})$}\;
% \end{algorithm}

Take the forward pass of Qwen3CausalLM (Algorithm~\ref{alg:vllm_forward_compact}) as an example. For the decoder layer $ i $, there are mainly three steps: first compute $\mathbf{K}_{i}, \mathbf{V}_{i}$ and update the KV cache, then compute $\mathbf{A}_i$, and finally compute $\mathbf{H}_i$ as next layer's input. Suppose we use a lower precision $p$, and let $\mathcal{F}(\mathbf{H}_i, \mathbf{C}^{(i)}, \mathbf{W}_o^{(i)})$ be the mapping from pre-attention tensor $ \mathbf{H}_i $ to post-attention tensor $ \mathbf{A}_i $, and $\mathcal{F}_p$ being its low precision counterpart, and let $\mathcal{Q}_p$ denote the function truncating values to precision $p$. We can then formally decompose the error between the exact $\mathbf{A}_i$ and the computed $\hat{\mathbf{A}}_i$ as:

{\small
% \vspace{-1em}
\begin{align} 
\hat{\mathbf{A}}_i - \mathbf{A}_i =& \mathcal{Q}_p\left(\mathcal{F}_p(\hat{\mathbf{H}}_i, \hat{\mathbf{C}}^{(i)}, \mathbf{W}_o^{(i)})\right) - \mathcal{F}(\mathbf{H}_i,  \mathbf{C}^{(i)}, \mathbf{W}_o^{(i)}) \\[-0.5ex]
=&\ \mathcal{Q}_p\left(\mathcal{F}_p(\hat{\mathbf{H}}_i, \hat{\mathbf{C}}^{(i)}, \mathbf{W}_o^{(i)})\right) - \mathcal{F}(\hat{\mathbf{H}}_i, \hat{\mathbf{C}}^{(i)}, \mathbf{W}_o^{(i)})\ + & \qquad (\epsilon_{\text{new}}) \label{equ:newly-injected-error} \\[-0.5ex]
&\ \mathcal{F}(\hat{\mathbf{H}}_i, \hat{\mathbf{C}}^{(i)}, \mathbf{W}_o^{(i)}) - \mathcal{F}(\hat{\mathbf{H}}_i, \mathbf{C}^{(i)}, \mathbf{W}_o^{(i)})\ + & \qquad (\epsilon_{\text{kv}}) \label{equ:kv-inherited-error} \\[-0.5ex]
&\ \mathcal{F}(\hat{\mathbf{H}}_i, \mathbf{C}^{(i)}, \mathbf{W}_o^{(i)})- \mathcal{F}(\mathbf{H}_i, \mathbf{C}^{(i)}, \mathbf{W}_o^{(i)}). & \qquad (\epsilon_{\text{old}}) \label{equ:residue-inherited-error}
\end{align}
% \vspace{-1em}
}

\begin{wrapfigure}{L}{0.6\linewidth}
%\vspace{-1.5em}
\centering
% \begin{figure}% \vskip 0.2in
% \begin{center}
% \vspace{-2em}
\includegraphics[width=\linewidth]{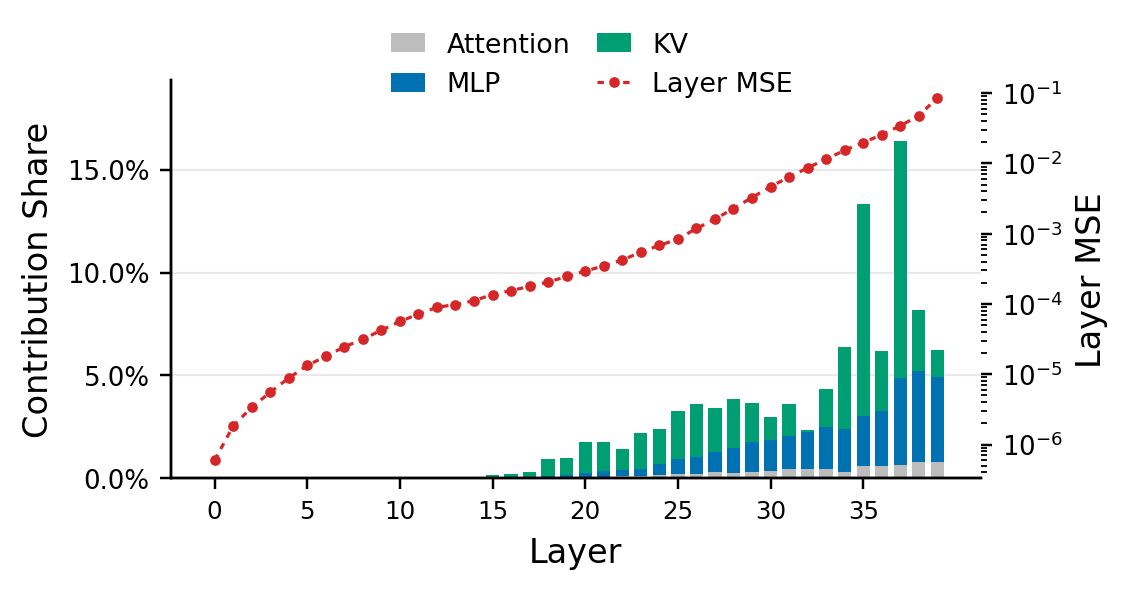}
% \vspace{-2em}
\caption{The contribution of error sources across the network.}
\label{fig:error-contribution}
% \end{center}
% \end{figure}
% \vspace{-1em}
\end{wrapfigure}

There are three distinct structural sources: the newly injected error $\epsilon_{\text{new}}$ contributed by the rounding error of $\mathcal{F}_p$, the horizontal error $\epsilon_{\text{kv}}$ inherited from past tokens via the KV cache, and the vertical error $\epsilon_{\text{old}}$ propagated upward from prior layers via the residual stream. This derivation also applies to $ \mathbf{K_i}, \mathbf{V_i}, \mathbf{H_{i+1}}$ and the final $\mathbf{L}$, in which there is no $\epsilon_{\text{kv}}$. Numerical instability stems from floating-point rounding errors, which is contributed by three mappings within each layer: $\mathcal{F}^{(\text{kv})}: \mathbf{H} \rightarrow \mathbf{K}, \mathbf{V}$, $\mathcal{F}^{(\text{attention})}: \mathbf{H} \rightarrow \mathbf{A}$, $\mathcal{F}^{(\text{mlp})}: \mathbf{A} \rightarrow \mathbf{H}$. Consequently, we isolate the relative contribution of each source, and attribute the final error onto different layers, as shown in Figure~\ref{fig:error-contribution}. Please find our detailed experiments and discussions regarding this process in Appendix~\ref{study:error-contribution}.

This layer-wise tracking reveals two critical operational dynamics of LLM inference: 
i) \emph{Non-Linear Accumulation}: As shown by the red line in Figure~\ref{fig:error-contribution}, the total accumulated Mean Squared Error~(MSE) exhibits exponential growth. The numerical noise magnifies multiplicatively as it propagates deeper into the network, pushing the final log-probabilities toward the catastrophic divergence threshold. 
ii) \emph{Structural Imbalance}: Crucially, the injection of this catastrophic error is not ubiquitous. The bar chart in Figure~\ref{fig:error-contribution} demonstrates that layers have different contributions within the whole model.

% This structural imbalance dictates a straightforward mitigation strategy: we can simply fully upcast the most crucial layers to FP32. However, we observe that even 1\% of the MSE error observed at BF16 is already more significant than that of TF32 .
% \ztnote{Maybe replace with BF16 MSE error is 1000x higher than FP32}. 
% But it is still possible for us to avoid upcasting on some initial layers, where error contributions are really minor.

% This structural imbalance dictates a straightforward mitigation strategy. While the error amplification is concentrated in specific bottlenecks, we observe that even 1\% of the MSE accumulated under native BF16 strictly exceeds the divergence threshold. Consequently, simply upcasting the one or two highest peaks is mathematically insufficient; to guarantee end-to-end reproducibility, we must actively apply our upcasting interventions across the majority of the network's layers. However, we can safely bypass the initial layers entirely, leaving them in lower precisions, as their microscopic error contributions are naturally absorbed without threatening the divergence threshold.

This structural imbalance dictates a straightforward mitigation strategy. While the error amplification is concentrated in specific bottlenecks, we observe that even 1\% of the MSE accumulated under native BF16 strictly exceeds the divergence threshold. Consequently, simply upcasting the one or two highest peaks is mathematically insufficient; to preserve end-to-end reproducibility, we must actively apply our upcasting interventions across the majority of the network's layers. The initial layers, whose error contributions are minor, can optionally remain in lower precision to recover throughput.\footnote{We evaluate this trade-off as the \techname{}-Base+Attn+Skip variant in \S\ref{sec:eval}.}

\subsection{The Microscopic Origin: Boundary Truncation}

Having established that catastrophic divergence is driven by the structural amplification of baseline errors within specific layers, we must isolate the microscopic origin of these errors. What exactly generates the new error injected at every layer?

A common misconception is that standard 16-bit inference implies low-precision arithmetic throughout the entire execution pipeline. If true, the mathematical operations themselves would be the primary source of instability. To test this assumption, we profiled the NVIDIA SASS~(Streaming ASSembly) instructions of the primary inference kernels on an NVIDIA A100, encompassing modern Attention implementations~(FlashAttention~\cite{shah2024flashattention} and TritonAttention~\cite{ringlein2025anatomytritonattentionkernel}).

\begin{table}[h!]
\centering
% \scriptsize
% \footnotesize
% \vspace{-0.5em}
\caption{Breakdown of internal floating-point arithmetic instructions for primary kernels in LLM inference.}
% \vspace{-0.5em}
\resizebox{\linewidth}{!}{
\begin{tabular}{ll}
\hline
Kernel & Instructions \\
\hline
ArgmaxReduce & \textbf{HFMA2.MMA} (100.0\%) \\
FlashFwdSplitKV & \textbf{HMMA.16816.F32.BF16} (43.2\%), FADD.FTZ (11.1\%), FFMA.FTZ (11.1\%), FMNMX.FTZ (11.1\%), FMUL.FTZ (11.5\%) \\
FusedAddRMSNorm & \textbf{HFMA2.BF16\_V2} (16.3\%), FADD (49.3\%), FFMA (9.2\%), FMUL (24.8\%), MUFU.RCP (0.2\%) \\
GEMM & \textbf{HMMA.16816.F32.BF16} (93.0\% - 97.6\%), FADD (1.3\% - 2.3\%), FFMA (2.3\%), FMUL (2.4\% - 5.9\%) \\
GPUIndexKernel & \textbf{HFMA2.MMA} (100.0\%) \\
KernelUnifiedAttention & \textbf{HMMA.16816.F32.BF16} (13.1\% - 15.0\%), FADD (11.5\% - 16.2\%), FFMA (4.9\% - 7.5\%), FMNMX (8.2\% - 11.2\%), FMUL (42.5\% - 56.3\%) \\
PageAttentionV2 & \textbf{HMMA.16816.F32.BF16} (94.1\% - 96.4\%), \textbf{HMNMX2.BF16\_V2} (1.2\%), FFMA (2.4\%), FMNMX (2.9\%), FMUL (2.9\%) \\
RMSNorm & FADD (24.5\% - 51.4\%), FFMA (18.6\% - 25.8\%), FMUL (25.7\% - 49.1\%), MUFU.RCP (0.3\% - 2.9\%), MUFU.RSQ (0.3\% - 1.4\%) \\
ReshapeAndCacheFlash & \textbf{HFMA2.MMA} (14.3\%), MUFU.RCP (85.7\%) \\
RotaryEmbedding & FADD (31.2\%), FMUL (62.5\%), MUFU.RCP (6.2\%) \\
SoftmaxReduce & FADD (21.3\%), FFMA (17.3\%), FMNMX (5.3\%), FMUL (45.3\%), MUFU.EX2 (6.7\%) \\
SplitKReduce & \textbf{HFMA2.MMA} (25.0\% - 28.6\%), FADD (42.9\% - 50.0\%), FFMA (12.5\% - 14.3\%), FMUL (12.5\% - 14.3\%) \\
SwiGLU & FADD (6.9\%), FFMA (55.2\%), FFMA.RM (6.9\%), FFMA.SAT (6.9\%), MUFU.EX2 (6.9\%) \\
TritonFused1 & FADD (55.6\% - 75.0\%), FFMA (12.5\% - 33.3\%), FMUL (11.1\% - 12.5\%) \\
TritonFused2 & \textbf{HFMA2.MMA} (1.2\%), FFMA (29.6\% - 30.0\%), FMUL (59.3\% - 60.0\%), MUFU.RSQ (9.9\% - 10.0\%) \\
TritonFused3 & FADD (18.2\%), FMUL (63.6\%), MUFU.EX2 (9.1\%), MUFU.RCP (9.1\%) \\
TritonFused4 & \textbf{HFMA2.MMA} (2.6\%), FADD (30.8\%), FFMA (12.8\%), FMUL (51.3\%), MUFU.RSQ (2.6\%) \\
TritonFused5 & \textbf{HFMA2.MMA} (1.8\% - 2.8\%), FADD (50.9\% - 61.1\%), FFMA (6.9\% - 9.1\%), FMUL (27.8\% - 36.4\%), MUFU.RSQ (1.4\% - 1.8\%) \\
\hline
\end{tabular}
}
% \vspace{-0.5em}
\label{tab:kernel_instructions}
\end{table}

\begin{wrapfigure}{R}{0.5\linewidth}
% \vspace{-1.5em}
\centering
% \begin{figure}% \vskip 0.2in
% \begin{center}
% \vspace{-2em}
\includegraphics[height=10em]{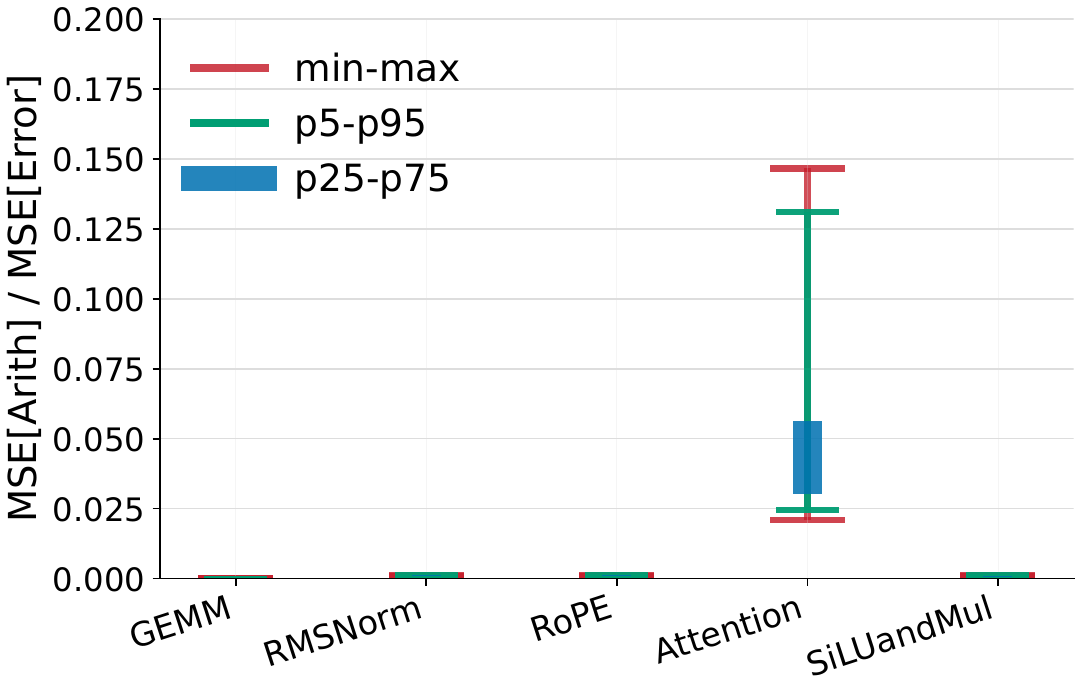}
\caption{The $\mathbb{E}[\epsilon_{\text{arith}}^2] / \mathbb{E}[(\hat{y} - y)^2]$ ratio of kernels in a vLLM forward pass on NVIDIA A100.}
% \vspace{-2em}
\label{fig:quant-vs-arith}
% \end{center}
% \end{figure}
% \vspace{-1em}
\end{wrapfigure}

As detailed in Table~\ref{tab:kernel_instructions}, our analysis reveals a contrasting reality: the vast majority of critical kernels actually perform their internal arithmetic in FP32 to maintain numerical stability. The dominant \codeIn{GEMM} and \codeIn{Attention} kernels rely on Tensor Core instructions~(\eg, \codeIn{HMMA.16816.F32.BF16}). Crucially, while these instructions accept BF16 inputs, they inherently utilize \codeIn{f32} accumulators for the matrix multiplication sums~\cite{nvidia2020a100}. Operations such as \codeIn{RotaryEmbedding}, \codeIn{Softmax}, and \codeIn{RMSNorm} explicitly cast inputs to FP32 registers before executing standard single-precision arithmetic. We verify this stands true across not only NVIDIA GPUs, but also AMD GPUs.

Because the arithmetic logic itself is nearly identical to a fully stable FP32 pipeline, it cannot be the primary source of numerical error. We can formally verify this by modeling the error of an individual operator $\mathcal{F}$ (\eg, \codeIn{RotaryEmbedding}, \codeIn{Softmax}, \codeIn{GEMM}) at a given layer. Given the input $x$ and the exact output $y$, suppose we compute under precision $p$. The overall error is given by: 
% \ztnote{Should we change this to $\epsilon_{\text{trunc}}$?}

{\small
% \vspace{-1em}
\begin{equation}
\hat{y} - y = \mathcal{Q}_p\left(\mathcal{F}_p(x)\right) - \mathcal{F}(x) = \underbrace{(\mathcal{Q}_p\left(\mathcal{F}_p(x)\right) - \mathcal{F}_p(x))}_{\epsilon_{\text{trunc}}} + \underbrace{(\mathcal{F}_p(x) - \mathcal{F}(x))}_{\epsilon_{\text{arith}}}
\end{equation}
% \vspace{-1em}
}

Assuming the values in the tensors have random significands as we observed in our experiments, the quantization error and arithmetic error should be independent. Since modern hardware employs unbiased Round-to-Nearest-Even~(RNE) rounding, we derive: 

{\small
% \vspace{-1em}
\begin{equation}
\mathbb{E}[\epsilon_{\text{trunc}}] = \mathbb{E}[\epsilon_{\text{arith}}] = 0 \Rightarrow \mathbb{E}[(\hat{y} - y)^2] = \mathbb{E}[\epsilon_{\text{trunc}}^2] + \mathbb{E}[\epsilon_{\text{arith}}^2]
\end{equation}
% \vspace{-1em}
}

Based on this, we can measure the contribution of $\epsilon_{\text{arith}}$ via the ratio $\mathbb{E}[\epsilon_{\text{arith}}^2] / \mathbb{E}[(\hat{y} - y)^2]$. As shown in Figure~\ref{fig:quant-vs-arith}, we confirm that $\epsilon_{\text{trunc}} \gg \epsilon_{\text{arith}}$ for almost all critical bottleneck kernels, as the arithmetic error is strictly bounded by the FP32 machine epsilon ($1.19\times10^{-7}$) when computation is natively in FP32. The single notable exception is the \codeIn{Attention} kernel, which exhibits a relatively higher arithmetic error. This anomaly occurs because standard vLLM implementations default to TF32~(TensorFloat-32) within their Triton Attention kernels even under a nominal FP32 pipeline, truncating the machine epsilon $9.77 \times 10^{-4}$.

% Based on this, we can measure the contribution of $\epsilon_{\text{arith}}$ via the ratio $\mathbb{E}[\epsilon_{\text{arith}}^2] / \mathbb{E}[(\hat{y} - y)^2]$. As shown in Figure~\ref{fig:quant-vs-arith} \zt{maybe add a footnote here pointing to appendix}, we confirm that $\epsilon_{\text{trunc}} \gg \epsilon_{\text{arith}}$ for almost all critical bottleneck kernels, as the arithmetic error is strictly bounded by the FP32 machine epsilon ($1.19\times10^{-7}$) when all computation is natively in FP32. \todoinline{Specifically, attention...}

% \begin{figure}[!t]
%   \begin{center}
%     \includegraphics[width=0.8\columnwidth]{figures/neurips_samples/operator_truncation_overview.pdf}
%     \caption{The $\mathbb{E}[\epsilon_{\text{arith}}^2] / \mathbb{E}[(\hat{y} - y)^2]$ ratio for all components confirms truncation dominates. \todoinline{no need to have title, larger font, group by module type}}
%     \label{fig:quant-vs-arith}
%   \end{center}
% \end{figure}

% \begin{figure}[!t]
%   \begin{center}
%     \includegraphics[width=0.8\columnwidth]{figures/neurips_samples/ph_mantissa_flip.jpg}
%     \caption{Answer flip rate decreases as significand bits are preserved. \todoinline{add final figure}}
%     \label{fig:mantissa_flip}
%   \end{center}
% \end{figure}

\begin{wrapfigure}{R}{0.5\linewidth}
% \vspace{-1.5em}
\centering
% \begin{figure}% \vskip 0.2in
% \begin{center}
\includegraphics[height=10em]{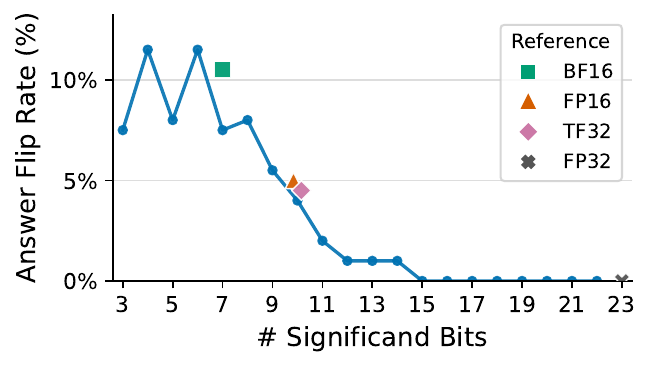}
% \vspace{-1em}
\caption{Answer flip rate decreases as significand bits are preserved.}
\label{fig:mantissa-vs-flip}
% \end{center}
% \end{figure}
% \vspace{-1.5em}
\end{wrapfigure}

The true driver of instability is exclusively $\epsilon_{\text{trunc}}$: the mandatory truncation of high-precision registers down to lower-precision storage at kernel boundaries. To further support our claim, we implemented an entirely FP32 pipeline with manual truncation injected at every kernel boundary. Re-evaluating a subset of MedQA under these controlled conditions, Figure~\ref{fig:mantissa-vs-flip} demonstrates that the answer flip rate can be effectively reduced strictly by increasing the number of preserved significand bits. Crucially, we do not need the full 23 significand bits of native FP32 to achieve near-perfect reproducibility; stability is reached much earlier, proving we only need enough fractional precision to cap the noise below the catastrophic divergence threshold. % Note that the marginal instability remaining in FP32 reflects the model's inherent knowledge boundaries rather than numerical variance, which drives us to develop MCR-Bench.

Therefore, we establish the conclusion: non-reproducibility across heterogeneous hardware is not a computation problem; it is strictly a kernel boundary problem. The microscopic seed of all downstream catastrophic divergence is the truncation of the significand during downcasting. Having identified the exact origin of the error, we now introduce our methodology to neutralize this error and secure end-to-end reproducibility.

\section{\techname: Hybrid Error ALleviation}
\label{sec:solution}

Having established that boundary truncation is the primary driver of numerical divergence, the most straightforward mitigation strategy is a global precision upcast. Specifically, as outlined in the forward pass of Algorithm~\ref{alg:vllm_forward_compact}, we could theoretically promote every kernel and intermediate tensor to FP32, eliminating kernel boundary truncation entirely. While this brute-force approach effectively neutralizes the error, it is hardware-inefficient. Intermediate state tensors like $\mathbf{H}$ and $\mathbf{A}$, alongside kernels like \codeIn{Norm}, can indeed be safely upcast to FP32 with negligible system overhead. However, this global promotion strategy hits a prohibitive performance wall at the network's two most computationally dominant components: the \codeIn{Attention} and \codeIn{GEMM} kernels.

For \codeIn{Attention}, the barrier to upcasting is not computational throughput, but memory capacity. The bottleneck lies entirely within the $\mathbf{Q}$, $\mathbf{K}$, and $\mathbf{V}$ tensors. Elevating the $\mathbf{K}$ and $\mathbf{V}$ tensors to FP32 strictly doubles the memory footprint of the KV cache, while upcasting $\mathbf{Q}$ significantly inflates intermediate memory usage especially when we have large batches. 
% \ztnote{I feel this is not clear for Q} 
Because the KV cache already dictates the memory wall in modern LLM inference, this $2\times$ expansion directly forces a proportional reduction in maximum batch sizes and sequence lengths.

Conversely, for the dense \codeIn{GEMM} kernel, the barrier shifts to computational throughput. Upcasting these matrix multiplications to FP32 incurs a severe performance penalty that directly contradicts the hardware design of modern AI accelerators. Contemporary architectures are fundamentally optimized for 16-bit arithmetic: NVIDIA's H100 Tensor Cores, for instance, deliver a 14.8$\times$ higher throughput for BF16 operations relative to native FP32, while AMD's MI300X Matrix Cores exhibit an 8$\times$ advantage. Consequently, a \naive FP32 promotion would severely degrade inference performance.

% Due to the hardness of promoting these kernels in full, HEAP reimplements them and forces 16-bit input, and use FP32 output to avoid the truncation errors introduced at kernel exit. This reimplementation does not necessarily introduce new computations, as Table \ref{tab:kernel_instructions} has shown that almost all computation in these kernels are already FP32.

To achieve both numerical stability and inference efficiency, we introduce \textbf{H}ybrid \textbf{E}rror \textbf{AL}leviation~(\textbf{\techname}). \techname applies a tailored, hardware-aware algebraic strategy to \codeIn{Attention} and \codeIn{GEMM}. For \codeIn{Attention}, \techname adopts compact INT16 storage and dynamically unpacks it into a high-precision dual-FP16 calculation. For \codeIn{GEMM}, \techname maximizes Tensor Core throughput by executing native FP16 matrix multiplications, rescaling the rare outlier channels to avoid overflow and compensating the residual truncation error through a dual-FP16 decomposition. Together, these techniques surgically neutralize the microscopic truncation errors while maintaining the footprint and latency advantages of a native 16-bit pipeline. In practice, \techname stores K and V with fine-grained group-wise scales to better utilize the INT16 resolution.\footnote{More implementation details can be found in Appendix~\ref{app:impl}.}

\MyPara{Mitigating QKV Truncation Error via INT16 Quantization.} To mitigate truncation error within the memory-bound \codeIn{Attention}, we first observe that standard 16-bit floating-point formats fundamentally underutilize their bit-width when storing $\mathbf{Q}, \mathbf{K}, \mathbf{V}$ tensors. We can formalize this by comparing the expected MSE of the truncation. Given a tensor $h$ with a maximum absolute value $M$ and a standard deviation $\sigma$, an INT16 uniform quantization yields an error uniformly distributed between $[-\Delta/2, \Delta/2]$ where $\Delta=\frac{M}{2^{15}-1}$. The expected MSE is therefore:

{\small
% \vspace{-1em}
\begin{equation}
\mathbb{E}[\text{MSE}_{\text{INT16}}] = \frac{\Delta^2}{12} \approx \frac{M^2 \cdot 2^{-30}}{12}.
\end{equation}
% \vspace{-1em}
}

In contrast, for a standard FP16 tensor, assuming a value $x \in h$ falls within the exponent range $2^{t} < |x| \le 2^{t+1}$, the truncation error is uniformly distributed within $[-2^{t-12}, 2^{t-12}]$. This yields an expected MSE of:

{\small
% \vspace{-1em}
\begin{equation}
\mathbb{E}[\text{MSE}_{\text{FP16}}] = \mathbb{E}\left[\frac{2^{2t-22}}{12}\right] > \frac{\mathbb{E}[x^2] \cdot 2^{-22}}{12} = \frac{\sigma^2 \cdot 2^{-22}}{12}.
\end{equation}
% \vspace{-1em}
}

\begin{wrapfigure}{R}{0.5\linewidth}
% \vspace{-1.5em}
\centering
% \begin{figure}% \vskip 0.2in
% \begin{center}
\includegraphics[height=10em]{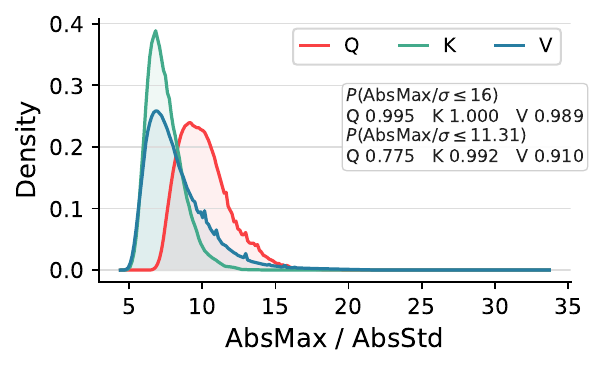}
% \vspace{-1em}
\caption{Q, K, V numerical value distribution}
% no need to update this, new plots are equivalent to this
\label{fig:qkv-distribution}
% \end{center}
% \end{figure}
% \vspace{-1.5em}
\end{wrapfigure}

This observation dictates that, even in the worst-case scenario, the error ratio is strictly bounded by $\mathbb{E}[\text{MSE}_{\text{FP16}}] / \mathbb{E}[\text{MSE}_{\text{INT16}}] > 256 \frac{\sigma^2}{M^2}$. As Figure~\ref{fig:qkv-distribution} demonstrates, this translates to a massive empirical precision gain: for 70\% of Q, K, V tensors, the theoretical MSE is decreased by at least 50\% (and will become 78\% when $x$ is uniformly distributed). Driven by this mathematical reality, \techname discards floating-point storage for the inputs entirely. Instead, any input tensor $\mathbf{X}$ is quantized and stored in INT16: $\mathcal{Q}_{\text{INT16}}(\mathbf{X}) = \text{Round}\left(\frac{(2^{15} - 1)\mathbf{X}}{\max |\mathbf{X}|}\right)$. Crucially, this aggressive scaling does not introduce an overflow risk during downstream matrix multiplications. Because the underlying hardware accelerators natively employ FP32 accumulators for the internal summations of GEMV operations, the intermediate scalar products remain safely bounded within high-precision registers.

To bridge the gap between this optimal storage format and standard hardware\textemdash which lacks native Tensor Core support for high-precision INT16 matrix multiplications\textemdash \techname algebraically decomposes the data prior to computation. During execution, the INT16 tensor is unpacked into two FP16 tensors: $\mathbf{X}_{\text{hi}}$ and $\mathbf{X}_{\text{lo}}$. Because a single FP16 value possesses 10 explicit significand bits, shifting the lower tensor by a factor of $2^{10}$ allows the dual-FP16 pairing ($\mathbf{X}_{\text{hi}} + \mathbf{X}_{\text{lo}} / 2^{10}$) to seamlessly emulate 20 bits of continuous fractional precision. \codeIn{Attention} is thus expanded to:

{\small
% \vspace{-1em}
\begin{equation}
\text{Attention}(\mathbf{Q}, \mathbf{K}, \mathbf{V}) = \text{softmax}\left(\frac{(\mathbf{Q}_{\text{hi}} + \mathbf{Q}_{\text{lo}} / 2^{10})(\mathbf{K}_{\text{hi}} + \mathbf{K}_{\text{lo}} / 2^{10})^T}{\sqrt{d_k}}\right)(\mathbf{V}_{\text{hi}} + \mathbf{V}_{\text{lo}} / 2^{10}).
\end{equation}
% \vspace{-1em}
}

While this algebraic expansion more than triples the FLOPS, it crucially preserves the strict 16-bit memory footprint. Because the \codeIn{Attention} is fundamentally memory-bound, this increased computational penalty is effectively absorbed under the roofline model. 

\MyPara{Mitigating GEMM Truncation via FP16 Promotion and Algebraic Compensation.} To address the remaining input truncation error with \codeIn{GEMM}, we must balance fractional precision against dynamic range. While standard BF16 prevents overflow, its severely truncated significand drives numerical divergence. Conversely, FP16 natively possesses 10 significand bits\textemdash matching the exact fractional precision of TF32. However, this precision comes with an extremely narrow dynamic range, introducing a catastrophic overflow risk: if activation magnitudes exceed the FP16 maximum, the entire inference pass will collapse into \codeIn{NaN}s.

% Fortunately, our profiling reveals a critical structural property of the network's activations: massive values within the intermediate tensors are rare and do not occur randomly. This phenomenon is well-supported by recent literature on large language model quantization \cite{dettmers2022gpt3, lin2024awq}. Specifically, these extreme outliers are highly isolated, existing exclusively within a microscopic fraction (typically 0.1\%-1\%) of fixed channels. Because these dominant channels remain remarkably static across different inputs, prior work has demonstrated that they can be reliably and effectively identified offline using a small calibration set \cite{lin2024awq}\footnote{For detailed profiling methodology, please refer to Appendix~\ref{study:outlier}.} \todoinline{rewrite here}

Fortunately, although there are empirical reports of FP16 overflow~\cite{sun2024massive, vllm_issue_40290}, such cases occur very infrequently. Throughout our testing process, we found that the maximum absolute value consistently remained below 1024, utilizing less than $\frac{1}{64}$ of the representable FP16 range. Even in the rare cases where overflow might happen, these extreme values only appear in a microscopic fraction (typically 0.1\%-1\%) of fixed channels~\cite{dettmers2022gpt3, lin2024awq}. Therefore, \techname can efficiently handle them by applying targeted rescaling (e.g., weight rescaling) to these exact channels. This allows us to eliminate the overflow risk without adding extra overhead.

% Capitalizing on this isolated sparsity, \techname shifts all standard \codeIn{GEMM} to FP16 to harvest its TF32-equivalent precision, while deploying an algebraic decomposition to safely neutralize the overflow risk. For any input tensor $h$, we mathematically split the extreme magnitudes located at the identified outlier channels ($\mathcal{O}$). For an index $i \in \mathcal{O}$, we decompose the value as $h_i = 2^{10} l_i + r_i$, bounding the residual $r_i$ safely within FP16 limits. We then construct a globally safe residual tensor $h'$, where $h'_i = r_i$ if $i \in \mathcal{O}$, and $h'_i = h_i$ otherwise. The original dense computation is thus safely routed into two parallel paths:

% {\small
% \vspace{-1em}
% \begin{equation}
% \text{GEMM}(h, \mathbf{W}^T) = \text{GEMM}(l_{\in \mathcal{O}}, \mathbf{W}_{\in \mathcal{O}}^T) + \text{GEMM}(h', \mathbf{W}^T).
% \end{equation}
% \vspace{-1em}
% }

% By isolating the massive magnitudes into a highly sparse, parallel matrix multiplication, the bulk of the dense workload ($h'$) can safely execute at maximum throughput on standard FP16 Tensor Cores.

% Finally, to push the truncation error floor even lower, \techname optionally performs a secondary decomposition on the residual sum $h'$, splitting it into a base FP16 tensor alongside a scaled INT8 \zt{we changed it to FP16} compensation tensor. Computing \codeIn{GEMM} between this INT8 compensation tensor and an INT8 quantized weight matrix seamlessly injects additional fractional precision into the final output.

Having safely bounded the dynamic range, \techname shifts all standard \codeIn{GEMM} operations to FP16 to harvest its TF32-equivalent precision. Furthermore, when strict task reproducibility demands precision beyond a single FP16 representation, \techname applies the same dual-FP16 pairing strategy introduced in \codeIn{Attention}. We mathematically decompose the high-precision FP32 input tensor $h$ as $(h_{\text{hi}} + h_{\text{lo}} / 2^{10})$. The original dense computation is thus expanded into two parallel, high-throughput FP16 operations:

{\small
% \vspace{-1em}
\begin{equation}
\text{GEMM}(h, \mathbf{W}^T) \approx \text{GEMM}(h_{\text{hi}}, \mathbf{W}^T) + \text{GEMM}(h_{\text{lo}} / 2^{10}, \mathbf{W}^T).
\end{equation}
% \vspace{-1em}
}

Similar to our approach in \codeIn{Attention}, computing the \codeIn{GEMM} with the $h_{\text{lo}}$ tensor seamlessly injects additional fractional bits into the final output. This allows \techname to approximate FP32 precision while keeping all computations fully optimized for standard Tensor Cores.

\section{Evaluation}
\label{sec:eval}

\MyPara{Evaluation Scope and Metrics.} 
For clarity, we distinguish three closely related but conceptually distinct notions: \emph{(i) Cross-hardware reproducibility}, the requirement that a fixed model and input yield functionally identical downstream task outcomes across heterogeneous hardware configurations; \emph{(ii) Agreement with FP32}, the degree to which low-precision execution aligns with a stable H100 FP32 reference anchor; and \emph{(iii) Task accuracy}, whether an output matches the dataset's objective ground-truth label. Crucially, task accuracy reflects a model's intrinsic capability and is orthogonal to reproducibility: \techname does not aim to alter model accuracy, but rather prevents boundary truncation noise from flipping the model's deterministic prediction across devices.

To rigorously evaluate \techname under real-world conditions, we introduce the Mission-Critical Reproducibility Benchmark~(MCR-Bench). MCR-Bench curates 4500 high-stakes reasoning problems from the medical, legal, and financial domains, specifically targeting complex scenarios where numerical divergence translates into catastrophic real-world failures.\footnote{More details about MCR-Bench can be found in Appendix~\ref{app:benchmark}.} To quantify reproducibility, our primary metric is the answer flip rate, which measures the percentage of generated responses that diverge from the FP32 anchor. Alongside this, we report the Final Hidden MSE, which, as established earlier, captures the mean squared error in the normalized representations immediately before the vocabulary projection matrix. Finally, to evaluate system efficiency, we utilize the ShareGPT dataset~\cite{sharegpt_v3} to measure both the Time Per Output Token~(TPOT) and the base memory footprint (accounting strictly for model weights and computational workspace). These performance metrics demonstrate that \techname successfully circumvents the severe computational slowdowns associated with full-precision pipelines, achieving high reproducibility while maintaining minimal performance overhead.

\MyPara{Setup \& Baselines.} We evaluate our system across three diverse hardware types: NVIDIA A100, NVIDIA H100, and AMD MI300X. We compare against four baseline precisions: BF16, FP16, TF32, and FP32 with LayerCast~\cite{yuan2025understanding} applied for 32-bit baselines. Alongside our complete system (\techname), we evaluate three cumulative ablation steps (\techname-Base, \techname{}-Base+Attn, and \techname{}-Base+Attn+Skip) to isolate the contribution of each component: i) \techname-Base: maintains the residual stream in FP32 while promoting standard \codeIn{GEMM} layers to FP16; ii) \techname{}-Base+Attn: integrates INT16 quantization and dual-FP16 unpacking for the \codeIn{Attention} QKV tensors; iii) \techname{}-Base+Attn+Skip: selectively bypasses upcasting for early layers whose error contributions safely remain below the result flipping threshold; iv) \techname: the complete recommended pipeline, incorporating the dual-FP16 pairing compensation for \codeIn{GEMM} to maximize fractional precision.\footnote{More details about the experiment setup and results can be found in Appendix~\ref{app:setup}.}

\MyPara{Reproducibility.}
We first evaluate end-to-end reproducibility on MCR-Bench. As we observe that the accumulated numerical error remains remarkably consistent across hardware platforms under a fixed precision configuration, we report the metrics averaged across all three GPUs in Table~\ref{tab:mse_flip_rate}. Note that \techname does not systematically change task accuracy.\footnote{More details about this can be found in Appendix~\ref{app:quality}.}

\begin{table*}[ht]
\centering
\caption{Final Hidden MSE and answer flip rate across five models on MCR-Bench.}
\label{tab:mse_flip_rate}
\resizebox{\textwidth}{!}{
\begin{tabular}{l|cc|cc|cc|cc|cc}
\toprule
\multirow{2}{*}{Approach} & \multicolumn{2}{c|}{Qwen3-8B} & \multicolumn{2}{c|}{Qwen3-14B} & \multicolumn{2}{c|}{Qwen3-32B} & \multicolumn{2}{c|}{Llama-8B} & \multicolumn{2}{c}{DeepSeek-14B} \\
 & MSE & Flip Rate & MSE & Flip Rate & MSE & Flip Rate & MSE & Flip Rate & MSE & Flip Rate \\
\midrule
BF16                & 1.6e-3  & 9.70\% & 1.3e-3  & 8.03\% & 4.6e-3  & 9.18\% & 1.9e-3  & 16.53\% & 1.4e-3  & 13.90\% \\
FP16                & 6.9e-5  & 4.90\% & 2.6e-5  & 3.84\% & 8.5e-5  & 4.50\% & 3.9e-5  & 3.82\%  & 3.3e-5  & 6.03\% \\
TF32 (+LayerCast)   & 2.2e-5  & 2.87\% & 1.2e-5  & 2.11\% & 4.9e-5  & 2.47\% & 2.8e-5  & 1.97\%  & 1.6e-5  & 3.77\% \\
FP32 (+LayerCast)   & 1.9e-10 & 0.01\% & 1.6e-10 & 0.01\% & 6.9e-10 & 0.00\% & 3.2e-10 & 0.01\%  & 1.4e-10 & 0.01\% \\
\midrule
\techname{}-Base           & 9.9e-6  & 2.16\% & 6.9e-6  & 1.36\% & 1.4e-5  & 1.73\% & 1.1e-5  & 1.22\%  & 5.0e-6  & 2.39\% \\
\techname{}-Base+Attn               & 3.4e-6  & 1.25\% & 1.4e-6  & 0.92\% & 4.3e-6  & 1.02\% & 2.8e-6  & 0.66\%  & 1.6e-6  & 1.25\% \\
\techname{}-Base+Attn+Skip              & 4.3e-6  & 1.48\% & 5.5e-6  & 1.01\% & 5.4e-6  & 1.09\% & 7.3e-6  & 0.84\%  & 2.9e-6  & 1.34\% \\
\techname{}                & 6.5e-8  & 0.06\% & 1.3e-8  & 0.07\% & 5.4e-8  & 0.07\% & 3.5e-8  & 0.10\%  & 1.0e-8  & 0.07\% \\
\bottomrule
\end{tabular}
}
\begin{flushleft}
\footnotesize $^*$ \textit{Llama-8B} refers to Llama-3.1-8B-Instruct and \textit{DeepSeek-14B} refers to DeepSeek-Distill-Qwen-14B.
\end{flushleft}
\end{table*}

As demonstrated in Table~\ref{tab:mse_flip_rate}, native low-precision executions~(BF16 and FP16) exhibit high token divergence due to boundary truncation. \techname-Base successfully pulls the accumulated MSE back into the benign regime. Ultimately, \techname neutralizes microscopic errors, closely aligned with the flip rate and MSE of the FP32 baseline, justifying that reproducibility does not require a global high-precision pipeline.

\MyPara{Performance Overhead.}
To benchmark practical deployment overhead, we evaluate TPOT and memory overhead by processing real-world inference workloads from the ShareGPT \cite{sharegpt_v3} dataset. For brevity and representativeness, we only present the results on Qwen3-14B in Table~\ref{tab:sharegpt_perf_14b}. The results for the remaining models are provided in Appendix~\ref{app:performance}.

\begin{table}[ht]
\centering
\caption{ShareGPT TPOT and Memory Usage for the \textbf{Qwen3-14B} model.}
\label{tab:sharegpt_perf_14b}
\resizebox{\textwidth}{!}{
\begin{tabular}{l|c|c|c|c}
\toprule
Configuration & A100 TPOT (ms) & H100 TPOT (ms) & MI300X TPOT (ms) & Memory Usage (GiB) \\
\midrule
BF16                 & $108.2\pm0.3$ & $41.7\pm0.1$ & $55.4\pm0.2$ & $28.1$ \\
TF32 (+LayerCast)    & $259.3\pm1.9$ & $114.8\pm0.1$ & $128.1\pm0.2$ & $29.6$ \\
FP32 (+LayerCast)    & $1332.3\pm5.9$ & $546.5\pm1.2$ & $243.2\pm0.4$ & $29.6$ \\
\midrule
\techname{}-Base            & $120.5\pm0.1$ & $47.5\pm0.0$ & $59.6\pm0.1$ & $28.3$ \\
\techname{}-Base+Attn                & $130.3\pm0.8$ & $53.7\pm0.1$ & $78.0\pm0.2$ & $28.3$ \\
\techname{}-Base+Attn+Skip               & $126.4\pm0.7$ & $51.6\pm0.2$ & $71.9\pm0.2$ & $28.3$ \\
\techname{}                 & $235.5\pm1.4$ & $91.7\pm0.5$ & $116.9\pm0.2$ & $28.5$ \\
\bottomrule
\end{tabular}
}
\end{table}

% While the FP32 baseline severely degrades throughput by bypassing Tensor Cores, \techname and \techname{}+ maintain a strict 16-bit memory footprint for the KV cache. \techname{}++ further recaptures throughput by skipping redundant upcasts. Specifically for the Qwen3-14B model across all platforms, \techname incurs an average 116.2\% overhead relative to the most efficient BF16 baseline, achieving reproducibility without compromising inference efficiency. \todoinline{Check this narrative.}

While the FP32 baseline severely degrades throughput by bypassing Tensor Cores, our progressive variants (\techname-Base and \techname{}-Base+Attn) preserve a strict 16-bit KV-cache footprint with modest latency costs, and \techname{}-Base+Attn+Skip further recaptures throughput by skipping redundant early-layer upcasts. Across all platforms on Qwen3-14B, the complete \techname pipeline operates within the native 16-bit memory footprint while incurring an average TPOT overhead of 116.2\% relative to BF16, avoiding the prohibitive slowdowns of full FP32 execution to deliver practical, mission-critical reproducibility.\footnote{More details about this can be found in Appendix~\ref{app:perf}.}
% \vspace{-.5em}
\section{Related Work}
\label{sec:related}

\MyPara{Reproducibility in LLM inference.} Achieving bitwise-identical inference is challenging due to numerical non-associativity and system-level variance~(\eg, parallel reductions)~\cite{yuan2025understanding}. Existing mitigation strategies typically attempt to enforce strict bit-level consistency by constraining execution paths, such as employing framework-level deterministic modes, batch-invariant kernels~\cite{he2025defeatingnondeterminism, lmsys2025sglang}, or tensor-parallel alignment~\cite{zhang2025deterministicinferencetensorparallel}. While these techniques mitigate run-to-run variance within a fixed software-hardware stack, they fundamentally fail across heterogeneous GPUs with disparate instruction sets and reduction implementations. Other approaches, such as LLM-42~\cite{gond2026llm} with verified speculation and ReproBLAS~\cite{demmel2016efficient} with order-independent specialized accumulators at the operator level, would require a deterministic pipeline, which is exceedingly difficult to guarantee across heterogeneous hardware, and often incur substantial performance overhead. In contrast, instead of enforcing bitwise determinism, \techname suppresses kernel-boundary truncation noise well below the argmax flipping threshold, achieving cross-hardware reproducibility while directly reusing native hardware kernels.

\MyPara{Mixed precision in LLM inference.} Mixed-precision strategies improve throughput by performing sensitive operations (\eg, attention, accumulation) in FP32 while storing data in BF16/FP16 \cite{nvidia2020a100, kwon2023vllm, shah2024flashattention}. Recent work also advocates for native FP16 execution~\cite{qi2025defeating} to eliminate training-inference mismatches, but lack deep investigation into fundamental instability mechanisms (\eg, kernel-boundary truncation) and offer no systematic mitigation against catastrophic overflow. Moreover, reproducibility still fails due to precision loss when writing back to lower precision and hardware-specific IEEE-754 behaviors \cite{nvidiacudafp}, which induce consistent but device-dependent perturbations.

% \MyPara{Quantization and numerical sensitivity.} Work on post-training quantization highlights how outliers dominate error, requiring specific smoothing or scaling strategies to preserve quality \cite{dettmers2022gpt3, lin2024awq, xiao2023smoothquant}. While prior research focuses on the accuracy-throughput trade-off, these findings align with our analysis of how rounding effects amplify drift and degrade cross-hardware reproducibility.

\MyPara{Quantization and numerical sensitivity.} Existing work on post-training quantization highlights how outliers dominate error, requiring specific smoothing or scaling strategies to preserve quality~\cite{dettmers2022gpt3}. While this line of work targets the accuracy-throughput trade-off rather than reproducibility, its findings align with our analysis of how rounding effects amplify drift. Specifically, weight-activation~\cite{dettmers2022gpt3, xiao2023smoothquant} and low-bit KV-cache quantization~\cite{hooper2024kvquant, liu2024kivi} inherently downcast intermediate states at kernel boundaries, introducing additional truncation that further degrades reproducibility. In contrast, weight-only quantization~\cite{frantar2023gptq, lin2024awq} compresses static parameters without introducing additional runtime truncation, making it orthogonal to and composable with \techname (Appendix \ref{study:quant}). \techname itself builds upon established numerical primitives, such as multi-component expansion~\cite{dekker1971floating, ootomo2022recovering} and channel-wise scaling~\cite{lin2024awq, xiao2023smoothquant}; \techname's contribution lies in the diagnosis of kernel-boundary truncation and its application to cross-hardware reproducibility.

% Recent efforts have sought to mitigate these performance penalties by exploring alternative execution paradigms. For instance, LLM-42~\cite{gond2026llm} attempts to mask the latency of deterministic execution through verified speculation, while other works advocate for native FP16 execution to align training and inference without the massive memory footprint of FP32~\cite{qi2025defeating}. However, these compromises still fall short for practical deployments: verified speculation remains structurally bottlenecked by the expensive deterministic verifier, and pure FP16 pipelines introduce catastrophic overflow risks due to their critically narrow dynamic range, leaving the fundamental tradeoff between hardware efficiency and numerical stability unresolved

% \vspace{-.5em}
\section{Conclusion}
\label{sec:conclusion}
% \vspace{-.5em}
This work demonstrates that cross-hardware non-reproducibility in LLM inference stems not from internal kernel arithmetic, but from significand truncation during downcasting at kernel boundaries, which accumulates across layers to trigger answer flips. To avoid the prohibitive cost of global FP32 execution, we propose \techname, a hybrid framework that bounds boundary truncation errors below the result-flipping threshold via compact INT16 Q/K/V quantization and dual-FP16 algebraic GEMM compensation on native 16-bit Tensor Cores. Evaluated on MCR-Bench across diverse models and accelerators, \techname achieves FP32-level functional reproducibility within the native 16-bit memory footprint while reducing the TPOT overhead of the FP32 pipeline by up to 6.0$\times$.

% \vspace{-.5em}
\section*{Acknowledgments}
% \vspace{-.5em}
We thank the anonymous reviewers for their comments.
This work was supported by the National Science Foundation (CNS-2301343, CNS-2330831, CNS-2403254, IIS-2546642, IIS-2551122), as well as the Alibaba Research Intern Program. Junyi Shu is the corresponding author.
\clearpage

\bibliography{paper}

@inproceedings{xu2024llm4workflow,
  title={Llm4workflow: An llm-based automated workflow model generation tool},
  author={Xu, Jia and Du, Weilin and Liu, Xiao and Li, Xuejun},
  booktitle={Proceedings of the 39th IEEE/ACM International Conference on Automated Software Engineering},
  pages={2394--2398},
  year={2024}
}

@misc{ringlein2025anatomytritonattentionkernel,
      title={{The Anatomy of a Triton Attention Kernel}}, 
      author={Burkhard Ringlein and Jan van Lunteren and Radu Stoica and Thomas Parnell},
      year={2025},
      eprint={2511.11581},
      archivePrefix={arXiv},
      primaryClass={cs.LG},
      url={https://arxiv.org/abs/2511.11581}, 
}

@misc{song2026evaluatingimplicitregulatorycompliance,
      title={{Evaluating Implicit Regulatory Compliance in LLM Tool Invocation via Logic-Guided Synthesis}}, 
      author={Da Song and Yuheng Huang and Boqi Chen and Tianshuo Cong and Randy Goebel and Lei Ma and Foutse Khomh},
      year={2026},
      eprint={2601.08196},
      archivePrefix={arXiv},
      primaryClass={cs.CL},
      url={https://arxiv.org/abs/2601.08196}, 
}

@misc{khatchadourian2025llmoutputdriftcrossprovider,
      title={{LLM Output Drift: Cross-Provider Validation \& Mitigation for Financial Workflows}}, 
      author={Raffi Khatchadourian and Rolando Franco},
      year={2025},
      eprint={2511.07585},
      archivePrefix={arXiv},
      primaryClass={cs.LG},
      url={https://arxiv.org/abs/2511.07585}, 
}

@article{ullah2024challenges,
author = {Ullah, Ehsan and Parwani, Anil and Baig, Mirza Mansoor and Singh, Rajendra},
address = {[London] :},
issn = {1746-1596},
journal = {Diagnostic pathology},
lccn = {2006243249},
number = {1},
publisher = {BioMed Central,},
title = {{Challenges and barriers of using large language models (LLM) such as ChatGPT for diagnostic medicine with a focus on digital pathology – a recent scoping review}},
volume = {19},
year = {2024},
}

@inproceedings{yue2024lawllm,
author = {Yue, Shengbin and Liu, Shujun and Zhou, Yuxuan and Shen, Chenchen and Wang, Siyuan and Xiao, Yao and Li, Bingxuan and Song, Yun and Shen, Xiaoyu and Chen, Wei and Huang, Xuanjing and Wei, Zhongyu},
booktitle = {{Lecture Notes in Computer Science}},
isbn = {978-981-97-5568-4},
issn = {0302-9743},
journal = {Lecture notes in computer science.},
lccn = {sf 92020465},
publisher = {Springer-Verlag},
series = {Database Systems for Advanced Applications},
title = {{LawLLM: Intelligent Legal System with Legal Reasoning and Verifiable Retrieval}},
volume = {14854},
year = {2024},
}

@misc{ingonyama2024llmreproducibility,
  author = {{Ingonyama}},
  title = {{Solving Reproducibility Challenges in Deep Learning and LLMs: Our Journey}},
  year = {2024},
  month = {September},
  day = {22},
  howpublished = {\url{https://www.ingonyama.com/post/solving-reproducibility-challenges-in-deep-learning-and-llms-our-journey}},
}

@misc{he2025defeatingnondeterminism,
  author = {He, Horace and Thinking Machines Lab},
  title = {{Defeating Nondeterminism in LLM Inference}},
  year = {2025},
  month = {September},
  day = {10},
  howpublished = {\url{https://thinkingmachines.ai/blog/defeating-nondeterminism-in-llm-inference/}},
  publisher = {Thinking Machines Lab},
  note = {Accessed: 2026-01-19}
}

@inproceedings{yuan2025understanding,
  title={{Understanding and mitigating numerical sources of nondeterminism in llm inference}},
  author={Yuan, Jiayi and Li, Hao and Ding, Xinheng and Xie, Wenya and Li, Yu-Jhe and Zhao, Wentian and Wan, Kun and Shi, Jing and Hu, Xia and Liu, Zirui},
  booktitle={The Thirty-ninth Annual Conference on Neural Information Processing Systems},
  year={2025}
}

@misc{zhang2025deterministicinferencetensorparallel,
  title={{Deterministic Inference across Tensor Parallel Sizes That Eliminates Training-Inference Mismatch}}, 
  author={Ziyang Zhang and Xinheng Ding and Jiayi Yuan and Rixin Liu and Huizi Mao and Jiarong Xing and Zirui Liu},
  year={2025},
  eprint={2511.17826},
  archivePrefix={arXiv},
  primaryClass={cs.LG},
  url={https://arxiv.org/abs/2511.17826}, 
}

@inproceedings{shah2024flashattention,
  title={{FlashAttention-3: Fast and Accurate Attention with Asynchrony and Low-precision}},
  author={Shah, Jay and Bikshandi, Ganesh and Zhang, Ying and Thakkar, Vijay and Ramani, Pradeep and Dao, Tri},
  booktitle={Proceedings of the 38th International Conference on Neural Information Processing Systems},
  pages={68658--68685},
  year={2024}
}

@inproceedings{xiao2023smoothquant,
  title={{Smoothquant: Accurate and Efficient Post-Training Quantization for Large Language Models}},
  author={Xiao, Guangxuan and Lin, Ji and Seznec, Mickael and Wu, Hao and Demouth, Julien and Han, Song},
  booktitle={International conference on machine learning},
  pages={38087--38099},
  year={2023},
  organization={PMLR}
}

@inproceedings{kwon2023vllm,
  title={{Efficient Memory Management for Large Language Model Serving with PagedAttention}},
  author={Kwon, Woosuk and Li, Zhuohan and Zhuang, Siyuan and Sheng, Ying and Zheng, Lianmin and Yu, Cody Hao and Gonzalez, Joseph and Zhang, Hao and Stoica, Ion},
  booktitle={Proceedings of the 29th symposium on operating systems principles},
  pages={611--626},
  year={2023}
}

@techreport{nvidia2020a100,
  title       = {{NVIDIA A100 Tensor Core GPU Architecture}},
  author      = {NVIDIA Corporation},
  year        = {2020},
  institution = {NVIDIA},
  type        = {Whitepaper},
  url         = {https://images.nvidia.com/aem-dam/en-zz/Solutions/data-center/nvidia-ampere-architecture-whitepaper.pdf}
}

@manual{nvidiacudafp,
  title        = {{Floating Point and IEEE 754 Compliance for NVIDIA GPUs}},
  author       = {NVIDIA Corporation},
  url          = {https://docs.nvidia.com/cuda/floating-point/index.html},
  note         = {CUDA Toolkit Documentation},
  key          = {NVIDIA}
}

@misc{lmsys2025sglang,
  author = {The SGLang Team},
  title = {Towards Deterministic Inference in SGLang and Reproducible RL Training},
  year = {2025},
  howpublished = {\url{https://lmsys.org/blog/2025-09-22-sglang-deterministic/}},
}

@misc{zhou2023ifeval,
      title={{Instruction-Following Evaluation for Large Language Models}}, 
      author={Jeffrey Zhou and Tianjian Lu and Swaroop Mishra and Siddhartha Brahma and Sujoy Basu and Yi Luan and Denny Zhou and Le Hou},
      year={2023},
      eprint={2311.07911},
      archivePrefix={arXiv},
      primaryClass={cs.CL},
      url={https://arxiv.org/abs/2311.07911}, 
}

@misc{merity2016wikitext2,
      title={{Pointer Sentinel Mixture Models}},
      author={Stephen Merity and Caiming Xiong and James Bradbury and Richard Socher},
      year={2016},
      eprint={1609.07843},
      archivePrefix={arXiv},
      primaryClass={cs.CL}
}

@misc{hipBLASLt,
  author = {Advanced Micro Devices Inc.},
  title = {{hipBLASLt documentation}},
  howpublished = {\url{https://rocm.docs.amd.com/projects/hipBLASLt/en/latest/}},
  year         = {2026}
}

@misc{cublaslt,
  author = {NVIDIA Corporation},
  title = {{Using the cuBLASLt API}},
  howpublished = {\url{https://docs.nvidia.com/cuda/cublas/#using-the-cublaslt-api}},
  year         = {2026}
}

@article{yang2025qwen3,
  title={{Qwen3 Technical Report}},
  author={Yang, An and Li, Anfeng and Yang, Baosong and Zhang, Beichen and Hui, Binyuan and Zheng, Bo and Yu, Bowen and Gao, Chang and Huang, Chengen and Lv, Chenxu and others},
  journal={arXiv preprint arXiv:2505.09388},
  year={2025}
}

@article{guo2025deepseek,
  title={{DeepSeek-R1: Incentivizing Reasoning Capability in LLMs via Reinforcement Learning}},
  author={Guo, Daya and Yang, Dejian and Zhang, Haowei and Song, Junxiao and Zhang, Ruoyu and Xu, Runxin and Zhu, Qihao and Ma, Shirong and Wang, Peiyi and Bi, Xiao and others},
  journal={arXiv preprint arXiv:2501.12948},
  year={2025}
}

@article{grattafiori2024llama,
  title={{The Llama 3 Herd of Models}},
  author={Grattafiori, Aaron and Dubey, Abhimanyu and Jauhri, Abhinav and Pandey, Abhinav and Kadian, Abhishek and Al-Dahle, Ahmad and Letman, Aiesha and Mathur, Akhil and Schelten, Alan and Vaughan, Alex and others},
  journal={arXiv preprint arXiv:2407.21783},
  year={2024}
}

@article{dettmers2022gpt3,
  title={{LLM.int8(): 8-bit Matrix Multiplication for Transformers at Scale}},
  author={Dettmers, Tim and Lewis, Mike and Belkada, Younes and Zettlemoyer, Luke},
  journal={Advances in neural information processing systems},
  volume={35},
  pages={30318--30332},
  year={2022}
}

@misc{ibm-redbooks,
  author = {Qais Noorshams and Chinmaya Mishra and Harald Seipp and Sailendu Patra and Dietmar Fischer and Kedar Karmarkar and Mathias Defiebre},
  title = {{Next-generation Enterprise AI solutions with IBM watsonx.ai and IBM Storage Scale}},
  year = {2026},
  howpublished = {\url{https://www.redbooks.ibm.com/redpieces/pdfs/redp5765.pdf}},
}

@misc{star,
  author = {Aleksandr Gornostal and Lori Mazor and Rachel Curtis},
  title = {{6 ways generative AI is transforming financial services}},
  howpublished = {\url{https://star.global/posts/generative-ai-in-finance/}},
}

@misc{blairstanek2025llmsprovideunstableanswers,
      title={{LLMs Provide Unstable Answers to Legal Questions}}, 
      author={Andrew Blair-Stanek and Benjamin Van Durme},
      year={2025},
      eprint={2502.05196},
      archivePrefix={arXiv},
      primaryClass={cs.CL},
      url={https://arxiv.org/abs/2502.05196}, 
}

@misc{coveo,
  author = {J\'er\'emie Sanfa\c{c}on},
  title = {{When temperature=0 isn’t zero: A Bedrock Determinism Bug in LangChain AWS}},
  year = {2026},
  howpublished = {\url{https://www.coveo.com/blog/bedrock-determinism-bug-in-langchain-aws/}},
}

@article{cardamone2025classifying,
  title={{Classifying Unstructured Text in Electronic Health Records for Mental Health Prediction Models: Large Language Model Evaluation Study}},
  author={Cardamone, Nicholas C and Olfson, Mark and Schmutte, Timothy and Ungar, Lyle and Liu, Tony and Cullen, Sara W and Williams, Nathaniel J and Marcus, Steven C},
  journal={JMIR Medical Informatics},
  volume={13},
  number={1},
  pages={e65454},
  year={2025},
  publisher={JMIR Publications Inc., Toronto, Canada}
}

@article{gim2026llm,
  title={{An LLM-assisted framework for accelerated and verifiable clinical hypothesis testing from electronic health records}},
  author={Gim, Nayoon and Gim, In and Jiang, Yu and Kihara, Yuka and Blazes, Marian and Wu, Yue and Lee, Cecilia S and Lee, Aaron Y},
  journal={medRxiv},
  year={2026},
  publisher={Cold Spring Harbor Laboratory Press}
}

@misc{icd,
  author = {Centers for Medicare \& Medicaid Services},
  title = {{ICD Code Lists}},
  year = {2026},
  howpublished = {\url{https://www.cms.gov/medicare/coordination-benefits-recovery/overview/icd-code-lists}},
}

@misc{aws-a100,
  author = {Charlotte Trueman},
  title = {{AWS has “never retired” an Nvidia A100 server, CEO Matt Garman claims}},
  year = {2026},
  howpublished = {\url{https://www.datacenterdynamics.com/en/news/aws-has-never-retired-an-nvidia-a100-server-ceo-matt-garman-claims/}},
}

@misc{nvidia-a100,
  author = {Rounak Jain},
  title = {{Nvidia Says Its 6-Year Old A100 Chips Are Running At Full Tilt, Countering Michael Burry’s Depreciation Warning}},
  howpublished = {\url{https://stocktwits.com/news-articles/markets/equity/nvidia-a100-gpu-shipped-6-years-ago-full-utilization-michael-burry-depreciation/cLPALD9RE9w}},
}

@misc{anthropic-google,
  author = {Anthropic},
  title = {{Anthropic expands partnership with Google and Broadcom for multiple gigawatts of next-generation compute}},
  year = {2026},
  howpublished = {\url{https://www.anthropic.com/news/google-broadcom-partnership-compute}},
}

@misc{anthropic-aws,
  author = {Anthropic},
  title = {{Powering the next generation of AI development with AWS}},
  year = {2024},
  howpublished = {\url{https://www.anthropic.com/news/anthropic-amazon-trainium}},
}

@misc{openai-aws,
  author = {OpenAI},
  title = {{OpenAI and Amazon announce strategic partnership}},
  year = {2026},
  howpublished = {\url{https://openai.com/index/amazon-partnership/}},
}

@misc{openai-amd,
  author = {OpenAI},
  title = {{AMD and OpenAI announce strategic partnership to deploy 6 gigawatts of AMD GPUs}},
  year = {2025},
  howpublished = {\url{https://openai.com/index/openai-amd-strategic-partnership/}},
}

@article{hpai2025careqa,
    author = {Anna Arias-Duart and Pablo Agustin Martin-Torres and Daniel Hinjos and Pablo Bernabeu-Perez and Lucia Urcelay Ganzabal and Marta Gonzalez Mallo and Ashwin Kumar Gururajan and Enrique Lopez-Cuena and Sergio Alvarez-Napagao and Dario Garcia-Gasulla},
    title = {Automatic Evaluation of Healthcare LLMs Beyond Question-Answering},
    year = {2025},
    eprint={2502.06666},
    archivePrefix={arXiv},
    primaryClass={cs.CL},
    url={https://arxiv.org/pdf/2502.06666},
}

@inproceedings{jin2019pubmedqa,
  title={PubMedQA: A Dataset for Biomedical Research Question Answering},
  author={Jin, Qiao and Dhingra, Bhuwan and Liu, Zhengping and Cohen, William and Lu, Xinghua},
  booktitle={Proceedings of the 2019 Conference on Empirical Methods in Natural Language Processing and the 9th International Joint Conference on Natural Language Processing (EMNLP-IJCNLP)},
  pages={2567--2577},
  year={2019}
}

@misc{guha2023legalbench,
      title={LegalBench: A Collaboratively Built Benchmark for Measuring Legal Reasoning in Large Language Models}, 
      author={Neel Guha and Julian Nyarko and Daniel E. Ho and Christopher Ré and Adam Chilton and Aditya Narayana and Alex Chohlas-Wood and Austin Peters and Brandon Waldon and Daniel N. Rockmore and Diego Zambrano and Dmitry Talisman and Enam Hoque and Faiz Surani and Frank Fagan and Galit Sarfaty and Gregory M. Dickinson and Haggai Porat and Jason Hegland and Jessica Wu and Joe Nudell and Joel Niklaus and John Nay and Jonathan H. Choi and Kevin Tobia and Margaret Hagan and Megan Ma and Michael Livermore and Nikon Rasumov-Rahe and Nils Holzenberger and Noam Kolt and Peter Henderson and Sean Rehaag and Sharad Goel and Shang Gao and Spencer Williams and Sunny Gandhi and Tom Zur and Varun Iyer and Zehua Li},
      year={2023},
      eprint={2308.11462},
      archivePrefix={arXiv},
      primaryClass={cs.CL}
}

@inproceedings{zheng2021does,
  title={When does pretraining help? assessing self-supervised learning for law and the casehold dataset of 53,000+ legal holdings},
  author={Zheng, Lucia and Guha, Neel and Anderson, Brandon R and Henderson, Peter and Ho, Daniel E},
  booktitle={Proceedings of the eighteenth international conference on artificial intelligence and law},
  pages={159--168},
  year={2021}
}

@inproceedings{zheng2025reasoning,
  title={A reasoning-focused legal retrieval benchmark},
  author={Zheng, Lucia and Guha, Neel and Arifov, Javokhir and Zhang, Sarah and Skreta, Michal and Manning, Christopher D and Henderson, Peter and Ho, Daniel E},
  booktitle={Proceedings of the 2025 Symposium on Computer Science and Law},
  pages={169--193},
  year={2025}
}

@inproceedings{chalkidis-etal-2022-lexglue,
    title = "{L}ex{GLUE}: A Benchmark Dataset for Legal Language Understanding in {E}nglish",
    author = "Chalkidis, Ilias  and
      Jana, Abhik  and
      Hartung, Dirk  and
      Bommarito, Michael  and
      Androutsopoulos, Ion  and
      Katz, Daniel  and
      Aletras, Nikolaos",
    booktitle = "Proceedings of the 60th Annual Meeting of the Association for Computational Linguistics (Volume 1: Long Papers)",
    month = may,
    year = "2022",
    address = "Dublin, Ireland",
    publisher = "Association for Computational Linguistics",
    url = "https://aclanthology.org/2022.acl-long.297",
    pages = "4310--4330",
}

@article{chen2021finqa,
  title={FinQA: A Dataset of Numerical Reasoning over Financial Data},
  author={Chen, Zhiyu and Chen, Wenhu and Smiley, Charese and Shah, Sameena and Borova, Iana and Langdon, Dylan and Moussa, Reema and Beane, Matt and Huang, Ting-Hao and Routledge, Bryan and Wang, William Yang},
  journal={Proceedings of EMNLP 2021},
  year={2021}
}

@misc{FinClaimRadar2025sksayan01,
  author    = {sksayan01},
  title     = {FinClaimRadar: A Dataset for Evidentiary Reasoning and Classification of Financial Claims},
  year      = {2025},
  publisher = {Hugging Face},
  version   = {1.0.0},
  url       = {https://huggingface.co/datasets/sksayan01/FinClaimRadar-Financial-Reasoning}
}

@article{sun2024massive,
  title={Massive activations in large language models},
  author={Sun, Mingjie and Chen, Xinlei and Kolter, J Zico and Liu, Zhuang},
  journal={arXiv preprint arXiv:2402.17762},
  year={2024}
}

@misc{vllm_issue_40290,
  author = {{wenqiangire-commits}},
  title = {[Bug]: {Gemma} 4 (31B/26B-A4B) vision outputs only <pad> under fp16 --- vision\_tower standardize overflows},
  howpublished = {GitHub Issue \#40290},
  year = {2026},
  url = {https://github.com/vllm-project/vllm/issues/40290}
}

@article{qi2025defeating,
  title={Defeating the training-inference mismatch via fp16},
  author={Qi, Penghui and Liu, Zichen and Zhou, Xiangxin and Pang, Tianyu and Du, Chao and Lee, Wee Sun and Lin, Min},
  journal={arXiv preprint arXiv:2510.26788},
  year={2025}
}

@article{gond2026llm,
  title={LLM-42: Enabling Determinism in LLM Inference with Verified Speculation},
  author={Gond, Raja and Kamath, Aditya K and Ramjee, Ramachandran and Panwar, Ashish},
  journal={arXiv preprint arXiv:2601.17768},
  year={2026}
}

@article{williams1989learning,
  title={A learning algorithm for continually running fully recurrent neural networks},
  author={Williams, Ronald J and Zipser, David},
  journal={Neural computation},
  volume={1},
  number={2},
  pages={270--280},
  year={1989},
  publisher={MIT Press}
}

@article{jin2021disease,
  title={What disease does this patient have? a large-scale open domain question answering dataset from medical exams},
  author={Jin, Di and Pan, Eileen and Oufattole, Nassim and Weng, Wei-Hung and Fang, Hanyi and Szolovits, Peter},
  journal={Applied Sciences},
  volume={11},
  number={14},
  pages={6421},
  year={2021},
  publisher={MDPI}
}

@article{lin2024awq,
  title={Awq: Activation-aware weight quantization for on-device llm compression and acceleration},
  author={Lin, Ji and Tang, Jiaming and Tang, Haotian and Yang, Shang and Chen, Wei-Ming and Wang, Wei-Chen and Xiao, Guangxuan and Dang, Xingyu and Gan, Chuang and Han, Song},
  journal={Proceedings of machine learning and systems},
  volume={6},
  pages={87--100},
  year={2024}
}

@misc{sharegpt_v3,
  author={{anon8231489123}},
  title={{ShareGPT} Vicuna Unfiltered Dataset},
  year={2023},
  howpublished={\url{https://huggingface.co/datasets/anon8231489123/ShareGPT_Vicuna_unfiltered}}
}

@article{2019t5,
  author = {Colin Raffel and Noam Shazeer and Adam Roberts and Katherine Lee and Sharan Narang and Michael Matena and Yanqi Zhou and Wei Li and Peter J. Liu},
  title = {Exploring the Limits of Transfer Learning with a Unified Text-to-Text Transformer},
  journal = {arXiv e-prints},
  year = {2019},
  archivePrefix = {arXiv},
  eprint = {1910.10683},
}

@article{paszke2019pytorch,
  title={Pytorch: An imperative style, high-performance deep learning library},
  author={Paszke, Adam and Gross, Sam and Massa, Francisco and Lerer, Adam and Bradbury, James and Chanan, Gregory and Killeen, Trevor and Lin, Zeming and Gimelshein, Natalia and Antiga, Luca and others},
  journal={Advances in neural information processing systems},
  volume={32},
  year={2019}
}

@article{sen1968estimates,
  title={Estimates of the regression coefficient based on Kendall's tau},
  author={Sen, Pranab Kumar},
  journal={Journal of the American statistical association},
  volume={63},
  number={324},
  pages={1379--1389},
  year={1968},
  publisher={Taylor \& Francis}
}

@article{demmel2016efficient,
  title={Efficient reproducible floating point summation and BLAS},
  author={Demmel, James and Ahrens, Peter and Nguyen, Hong Diep},
  journal={EECS Department, University of California, Berkeley, Tech. Rep. UCB/EECS-2016-121},
  year={2016}
}

@misc{pytorch-reproducibility,
  title={Reproducibility --- {PyTorch} Documentation},
  author={{PyTorch Contributors}},
  howpublished={\url{https://docs.pytorch.org/docs/stable/notes/randomness.html}},
}

@inproceedings{liu2024kivi,
  title={{KIVI}: A Tuning-Free Asymmetric 2bit Quantization for {KV} Cache},
  author={Liu, Zirui and Yuan, Jiayi and Jin, Hongye and Zhong, Shaochen and Xu, Zhaozhuo and Braverman, Vladimir and Chen, Beidi and Hu, Xia},
  booktitle={Proceedings of the 41st International Conference on Machine Learning (ICML)},
  series={PMLR},
  volume={235},
  pages={32332--32344},
  year={2024}
}

@inproceedings{hooper2024kvquant,
  title={{KVQuant}: Towards 10 Million Context Length {LLM} Inference with {KV} Cache Quantization},
  author={Hooper, Coleman and Kim, Sehoon and Mohammadzadeh, Hiva and Mahoney, Michael W. and Shao, Yakun Sophia and Keutzer, Kurt and Gholami, Amir},
  booktitle={Advances in Neural Information Processing Systems (NeurIPS)},
  volume={37},
  year={2024}
}

@inproceedings{frantar2023gptq,
  title={{GPTQ}: Accurate Post-Training Quantization for Generative Pre-trained Transformers},
  author={Frantar, Elias and Ashkboos, Saleh and Hoefler, Torsten and Alistarh, Dan},
  booktitle={International Conference on Learning Representations (ICLR)},
  year={2023}
}

@article{dekker1971floating,
  title={A floating-point technique for extending the available precision},
  author={Dekker, Theodorus J.},
  journal={Numerische Mathematik},
  volume={18},
  number={3},
  pages={224--242},
  year={1971},
  doi={10.1007/BF01397083}
}

@article{ootomo2022recovering,
  title={Recovering single precision accuracy from {Tensor Cores} while surpassing the {FP32} theoretical peak performance},
  author={Ootomo, Hiroyuki and Yokota, Rio},
  journal={The International Journal of High Performance Computing Applications},
  volume={36},
  number={4},
  pages={475--491},
  year={2022},
  doi={10.1177/10943420221090256}
}

@inproceedings{tillet2019triton,
  title={Triton: An Intermediate Language and Compiler for Tiled Neural Network Computations},
  author={Tillet, Philippe and Kung, H. T. and Cox, David},
  booktitle={Proceedings of the 3rd ACM SIGPLAN International Workshop on Machine Learning and Programming Languages},
  pages={10--19},
  year={2019},
  doi={10.1145/3315508.3329973}
}

@inproceedings{gema2025mmluredux,
  title={Are We Done with {MMLU}?},
  author={Gema, Aryo Pradipta and Leang, Joshua Ong Jun and Hong, Giwon and Devoto, Alessio and Mancino, Alberto Carlo Maria and Saxena, Rohit and He, Xuanli and Zhao, Yu and Du, Xiaotang and Ghasemi Madani, Mohammad Reza and Barale, Claire and McHardy, Robert and Harris, Joshua and Kaddour, Jean and van Krieken, Emile and Minervini, Pasquale},
  booktitle={Proceedings of the 2025 Conference of the Nations of the Americas Chapter of the Association for Computational Linguistics: Human Language Technologies (Volume 1: Long Papers)},
  year={2025}
}

@inproceedings{lightman2024verify,
  title={Let's Verify Step by Step},
  author={Lightman, Hunter and Kosaraju, Vineet and Burda, Yura and Edwards, Harri and Baker, Bowen and Lee, Teddy and Leike, Jan and Schulman, John and Sutskever, Ilya and Cobbe, Karl},
  booktitle={International Conference on Learning Representations (ICLR)},
  year={2024}
}

@inproceedings{jain2025livecodebench,
  title={LiveCodeBench: Holistic and Contamination Free Evaluation of Large Language Models for Code},
  author={Jain, Naman and Han, King and Gu, Alex and Li, Wen-Ding and Yan, Fanjia and Zhang, Tianjun and Wang, Sida and Solar-Lezama, Armando and Sen, Koushik and Stoica, Ion},
  booktitle={International Conference on Learning Representations (ICLR)},
  year={2025}
}

@misc{openai2024mmmlu,
  title={Multilingual {MMLU} ({MMMLU}) Dataset},
  author={{OpenAI}},
  year={2024},
  howpublished={\url{https://huggingface.co/datasets/openai/MMMLU}}
}

@inproceedings{hsieh2024ruler,
  title={{RULER}: What's the Real Context Size of Your Long-Context Language Models?},
  author={Hsieh, Cheng-Ping and Sun, Simeng and Kriman, Samuel and Acharya, Shantanu and Rekesh, Dima and Jia, Fei and Zhang, Yang and Ginsburg, Boris},
  booktitle={First Conference on Language Modeling (COLM)},
  year={2024}
}

@article{malo2014good,
  title={{Good Debt or Bad Debt: Detecting Semantic Orientations in Economic Texts}},
  author={Malo, Pekka and Sinha, Ankur and Korhonen, Pekka and Wallenius, Jyrki and Takala, Pyry},
  journal={Journal of the Association for Information Science and Technology},
  volume={65},
  number={4},
  pages={782--796},
  year={2014},
  doi={10.1002/asi.23062}
}

@inproceedings{maia2018www,
  title={{WWW'18 Open Challenge: Financial Opinion Mining and Question Answering}},
  author={Maia, Macedo and Handschuh, Siegfried and Freitas, Andr{\'e} and Davis, Brian and McDermott, Ross and Zarrouk, Manel and Balahur, Alexandra},
  booktitle={Companion Proceedings of The Web Conference 2018},
  pages={1941--1942},
  year={2018},
  doi={10.1145/3184558.3192301}
}

@inproceedings{hendrycks2021measuring,
  title={{Measuring Massive Multitask Language Understanding}},
  author={Hendrycks, Dan and Burns, Collin and Basart, Steven and Zou, Andy and Mazeika, Mantas and Song, Dawn and Steinhardt, Jacob},
  booktitle={International Conference on Learning Representations (ICLR)},
  year={2021}
}

@inproceedings{yang2018hotpotqa,
  title={{HotpotQA: A Dataset for Diverse, Explainable Multi-hop Question Answering}},
  author={Yang, Zhilin and Qi, Peng and Zhang, Saizheng and Bengio, Yoshua and Cohen, William and Salakhutdinov, Ruslan and Manning, Christopher D.},
  booktitle={Proceedings of the 2018 Conference on Empirical Methods in Natural Language Processing},
  pages={2369--2380},
  year={2018},
  doi={10.18653/v1/D18-1259}
}

@misc{gao2024lmeval,
  title={{The Language Model Evaluation Harness}},
  author={Gao, Leo and Tow, Jonathan and Abbasi, Baber and Biderman, Stella and Black, Sid and DiPofi, Anthony and Foster, Charles and Golding, Laurence and Hsu, Jeffrey and Le Noac'h, Alain and Li, Haonan and McDonell, Kyle and Muennighoff, Niklas and Ociepa, Chris and Phang, Jason and Reynolds, Laria and Schoelkopf, Hailey and Skowron, Aviya and Sutawika, Lintang and Tang, Eric and Thite, Anish and Wang, Ben and Wang, Kevin and Zou, Andy},
  year={2024},
  publisher={Zenodo},
  doi={10.5281/zenodo.12608602},
  howpublished={\url{https://zenodo.org/records/12608602}}
}

@misc{habib2023lighteval,
  title={{LightEval: A Lightweight Framework for LLM Evaluation}},
  author={Habib, Nathan and Fourrier, Cl{\'e}mentine and Kydl{\'\i}{\v{c}}ek, Hynek and Wolf, Thomas and Tunstall, Lewis},
  year={2023},
  howpublished={\url{https://github.com/huggingface/lighteval}}
}
\bibliographystyle{plainnat}

%%%%%%%%%%%%%%%%%%%%%%%%%%%%%%%%%%%%%%%%%%%%%%%%%%%%%%%%%%%%

\clearpage

\appendix

\section{Detailed Experimental Setup}
\label{app:setup}

\MyPara{Setup.} We conducted our experiments on top of vLLM~\cite{kwon2023vllm}, a commonly used inference engine. To ensure reproducibility and align with prior work~\cite{yuan2025understanding}, we strictly controlled the software environment. Specifically, the test environment was initialized using an official Ubuntu 22.04 Docker base image with g++-11, Python 3.12.13, and vLLM v0.19.0 installed. We use CUDA 12.8 with the standard Release 580 series drivers for NVIDIA GPUs, and ROCm 7.2.1 for AMD GPUs.

To ensure our findings represent fundamental systemic behaviors rather than artifacts of a specific architecture, our diagnostic analysis spans five widely deployed open-source models: Qwen3-8B, 14B, and 32B~\cite{yang2025qwen3}, Llama-3.1-8B-Instruct~\cite{grattafiori2024llama}, and DeepSeek-R1-Distill-Qwen-14B~\cite{guo2025deepseek}. We used one A100 80GB SXM4 GPU, one H100 80GB SXM5 GPU, and one AMD MI300X GPU to conduct the experiments. There is no GPU sharing when serving the models, and the temperature is fixed to zero unless otherwise specified.

\MyPara{Baselines.} Throughout our experiments, we evaluate four baselines with varied precision formats: BF16, FP16, TF32, and FP32. Specifically, these formats feature varying numbers of significand bits for precision: 7 for BF16, 10 for both FP16 and TF32, and 23 for FP32. For the 32-bit baselines (TF32 and FP32), we execute them with LayerCast~\cite{yuan2025understanding} enabled. While native FP32 and FP32 with LayerCast are theoretically equivalent, exhibiting only minor variations in their compiled CUDA graphs, we opt for LayerCast due to its reduced memory overhead. This design choice is practically necessary, as it enables us to serve Qwen3-32B entirely in 32-bit precision on a single 80GB GPU.

To isolate the impact of numerical representation and ensure a fair comparison across all precisions, we implement all baselines using TritonAttention, since standard FlashAttention currently lacks support for 32-bit formats. Crucially, to ensure rigorous 32-bit execution for the FP32 baseline, we explicitly set the \codeIn{input\_precision} argument to \codeIn{"ieee"} for all \codeIn{tl.dot} operations within the TritonAttention kernel. Without this explicit enforcement, the core matrix multiplications inside the TritonAttention silently fall back to TF32, introducing unintended numerical noise that confounds exact reproducibility analysis.

\MyPara{Experiments.} Throughout our evaluations, we use the FP32 LayerCast execution on the NVIDIA H100 as our reference anchor. Unless otherwise specified, we use vLLM's default parameters. Note that CUDA Graphs are disabled during our numerical and reproducibility analyses, and are exclusively enabled when measuring end-to-end inference performance.

% \MyPara{Experiments.} There are mainly two running modes of vLLM in our experiment. The first running mode is the native vLLM path, where models read prompts and auto-regressively generate the responses. In this running mode, CUDA Graph is always enabled to ensure the best performance. \todoinline{list what uses this running mode.}

% The other one enforces strict path alignment via Teacher Forcing \todoinline{cite}, utilizing FP32 LayerCast results on H100 as the ground truth. This Teacher Forcing mechanism allows us to track the layer-by-layer numerical divergence of low-precision executions against the anchor\zt{, avoiding divergence caused by different token histories}. \todoinline{list what uses this running mode.}

% Despite the aforementioned configurations, we use vLLM's default parameters.

\MyPara{Datasets.} For the motivating experiments in \S\ref{sec:anatomy} and \S\ref{sec:solution}, we randomly sample 200 questions from the 1273 questions in the MedQA benchmark \cite{jin2021disease}. To elicit detailed reasoning while preventing unbounded auto-regressive outputs, we enable the models' Chain-of-Thought (CoT) capabilities and enforce a conservative generation cap of 4096 tokens. We append a strict formatting instruction to the default prompt template to facilitate automated parsing:
\begin{promptbox}
"You are solving a US medical licensing multiple-choice question. Think step by step, then end with exactly one final line formatted as `Final Answer: X' where X is A, B, C, or D."
\end{promptbox}

Table~\ref{tab:medqa-subset-ground-truth} summarizes the execution statistics of this MedQA subset, utilizing the H100 FP32 LayerCast execution as our reference. The ``Truncated Due to Length'' metric reflects the rare instances where a model hit the aforementioned token cap, while ``Successfully Parsed'' indicates outputs that strictly adhered to our requested final answer format.

\begin{table}[H]
\centering
\small
\caption{H100 FP32 LayerCast Reference Statistics for the MedQA-200 Subset}
\label{tab:medqa-subset-ground-truth}
\begin{tabularx}{\textwidth}{l *{5}{>{\centering\arraybackslash}X}}
\toprule
\textbf{Metric} & \textbf{Qwen3-8B} & \textbf{Qwen3-14B} & \textbf{Qwen3-32B} & \textbf{Llama-8B} & \textbf{DS-14B} \\ \midrule
\# Questions & 200 & 200 & 200 & 200 & 200 \\
Median Response Length & 1022.5 & 751.5 & 823.5 & 353 & 1421 \\
Truncated Due to Length & 9/200 & 2/200 & 5/200 & 4/200 & 12/200 \\
Successfully Parsed & 191/200 & 198/200 & 193/200 & 195/200 & 188/200 \\
\# Generated Tokens & 295616 & 213541 & 222327 & 86354 & 337479 \\
Accuracy & 149/200 & 161/200 & 175/200 & 137/200 & 144/200 \\ \bottomrule
\end{tabularx}
\begin{flushleft}
\footnotesize $^*$ \textit{Llama-8B} refers to Llama-3.1-8B-Instruct and \textit{DS-14B} refers to DeepSeek-R1-Distill-Qwen-14B.
\end{flushleft}
\end{table}

For the MSE error and Flip Rate Analysis in \S\ref{sec:eval}, we transition to \textbf{MCR-Bench}, an evaluation protocol curated from existing datasets and specifically designed to assess LLM reproducibility on downstream tasks. MCR-Bench encompasses diverse workloads and manages its own dynamic generation constraints (see Appendix~\ref{app:benchmark} for further details).

For the performance evaluation in \S\ref{sec:eval}, we use the ShareGPT V3 dataset~\cite{sharegpt_v3} under the following vLLM serving setup:

\begin{promptbox}
\begin{verbatim}
vllm serve <model> \
  --max-model-len 8192 \
  --max-num-batched-tokens 8192 \
  --max-num-seqs 256 \
  --gpu-memory-utilization 0.90

vllm bench serve \
  --dataset-name sharegpt \
  --num-prompts 1024 \
  --request-rate inf \
  --seed 0 \
  --temperature 0
\end{verbatim}
\end{promptbox}

The detailed execution time of experiments is summarized in Table \ref{tab:fig1-fig4-execution-time}. In total, the reported experiments take approximately 440 A100 hours, 275 H100 hours, and 150 MI300X hours.

\begin{table}[H]
\centering
\small
\caption{Execution Time of Each Experiment (GPU Hours)}
\label{tab:fig1-fig4-execution-time}
\begin{tabularx}{\textwidth}{l c *{5}{>{\centering\arraybackslash}X}}
\toprule
\textbf{Task} & \textbf{GPU} & \textbf{Qwen3-8B} & \textbf{Qwen3-14B} & \textbf{Qwen3-32B} & \textbf{Llama-8B} & \textbf{DS-14B} \\ \midrule
Reference Anchor Generation
& H100   & 0.2 & 0.2 & 1.9 & 0.1 & 0.4 \\ \midrule
Error Contribution (Figure \ref{fig:error-contribution}, \ref{app:fig:error-contribution})
& H100   & 0.4 & 0.6 & 1.1 & 0.2 & 0.6 \\ 
\midrule
& A100   & 9.1 & 8.6 & 15.7 & 3.1 & 12.9 \\
Final Hidden MSE Analysis (Figure \ref{fig:error-vs-flip}, \ref{app:fig:error-vs-flip})
& H100   & 4.5 & 4.2 & 8.3 & 1.6 & 6.4 \\
& MI300X & 3.2 & 3.3 & 6.3 & 1.3 & 5.0 \\ \midrule
Flip Rate Analysis (Figure \ref{fig:error-vs-flip}, \ref{fig:mantissa-vs-flip}, \ref{app:fig:error-vs-flip}, \ref{app:fig:mantissa-vs-flip})
& MI300X & 1.8 & 2.2 & 6.2 & 1.2 & 2.8 \\ \midrule
Arithmetic vs Truncation (Figure \ref{fig:quant-vs-arith}, \ref{app:fig:quant-vs-arith})
& A100   & 0.1 & 0.2 & 1.9 & 0.1 & 0.2 \\ \midrule
QKV Distribution (Figure \ref{fig:qkv-distribution}, \ref{app:fig:qkv-distribution})
& H100   & 0.1 & 0.1 & 0.6 & 0.1 & 0.1 \\ \midrule
Weight-only Quantization (Table \ref{tab:quant})
& A100   & 7.6 & 7.9 & 25.2 & 3.0 & 12.0 \\ \midrule
& A100   & 26.9 & 45.7 & 167.9 & 18.2 & 57.0 \\
MCR-Bench (Table \ref{tab:mse_flip_rate})
& H100   & 26.3 & 39.2 & 113.0 & 17.2 & 41.0 \\
& MI300X & 15.6 & 21.2 & 46.1 & 8.8 & 22.1 \\ \midrule
& A100   & 2.0 & 2.9 & 8.8 & 1.9 & 3.0 \\
ShareGPT V3 (Table \ref{tab:sharegpt_perf_14b}, \ref{tab:sharegpt_perf_qwen3_8b}, \ref{tab:sharegpt_perf_qwen3_32b}, \ref{tab:sharegpt_perf_llama_8b}, \ref{tab:sharegpt_perf_deepseek_14b})
& H100   & 0.5 & 0.8 & 3.4 & 0.5 & 0.9 \\
& MI300X & 0.8 & 1.1 & 2.1 & 0.7 & 1.1 \\
\bottomrule
\end{tabularx}
\begin{flushleft}
\footnotesize $^*$ \textit{Llama-8B} refers to Llama-3.1-8B-Instruct and \textit{DS-14B} refers to DeepSeek-R1-Distill-Qwen-14B.
\end{flushleft}
\end{table}

\MyPara{Evaluation Metrics and Error Bars.} For all experiments where error bars are reported, we display the mean alongside the standard error of the mean (SEM). Formally, this interval is represented as:
\begin{equation}
\mu \pm \frac{\sigma}{\sqrt{N}} 
\end{equation}
where $\mu$ is the sample mean, $\sigma$ is the standard deviation, and $N$ represents the sample size (e.g., the total number of evaluated tokens for MSE evaluations, or the number of independent runs for generation experiments). For answer flip rates, the dominant source of variability is the finite question sample. With 4500 questions per GPU, the binomial standard error $\sqrt{p(1-p)/N}$ is at most 0.6\% for the baselines and about 0.05\% for \techname.

\section{Additional Study Results}
\label{app:study}

This section presents the results of our study on the set of supplementary models, including Qwen3-8B, Llama3.1-8B-Instruct (referred as \textbf{Llama-8B}), DeepSeek-R1-Distill-Qwen-14B (referred as \textbf{DeepSeek-14B}, \textbf{DS-14B}), and Qwen3-32B.

\subsection{Error and Flip Rate (Figure \ref{fig:error-vs-flip} and Figure \ref{fig:mantissa-vs-flip})}
\label{study:fig1} \label{study:fig4}

To investigate the relationship between numerical precision and model behavior, we developed a two-stage evaluation pipeline. We first implement an FP32 execution path with manual significand truncation injected at every kernel boundary. This pipeline is controlled by a parameter $k$, representing the number of preserved significand bits. By varying $k$ from 3 to 22 (when $k=23$ it becomes FP32), we generate 20 distinct precision modes. Combined with our 4 baselines, this results in 24 distinct configurations to evaluate. Our diagnostic analysis focuses on two primary metrics: Final Hidden Layer MSE and Answer Flip Rate.

\MyPara{Final Hidden Layer MSE Analysis.} To isolate numerical divergence from the compounding effects of different token histories, we utilize Teacher Forcing (replaying the reference anchor token trajectory). We use the FP32 LayerCast results on the NVIDIA H100 as the anchor. At each step, to quantify the accumulated noise just before the final decision, we measure the Final Hidden MSE, defined as the mean squared error in the representations immediately before the vocabulary projection matrix (i.e., the normalized final hidden layer output). We deliberately evaluate on this hidden dimension rather than the pre-softmax logits because its significantly smaller size allows us to substantially scale up our evaluation data volume.

Moreover, the pure replay mechanism avoids auto-regressive generation, which significantly accelerates the data collection process. Consequently, we are able to evaluate the MSE across all three hardware platforms: NVIDIA A100, NVIDIA H100, and AMD MI300X.

\MyPara{Answer Flip Rate Analysis.} To evaluate how internal numerical errors manifest as actual token deviations, we calculate the Answer Flip Rate, which requires a full auto-regressive decoding process. Based on our prompt format, we parse the model's final response into one of five discrete categories: A, B, C, D, or F, where F denotes a failure to successfully parse a formatted answer. An ``Answer Flip'' is explicitly defined as any instance where this categorical outcome diverges from our anchor execution (H100 FP32 LayerCast).

\begin{figure}
\begin{center}
\includegraphics[width=0.48\columnwidth]{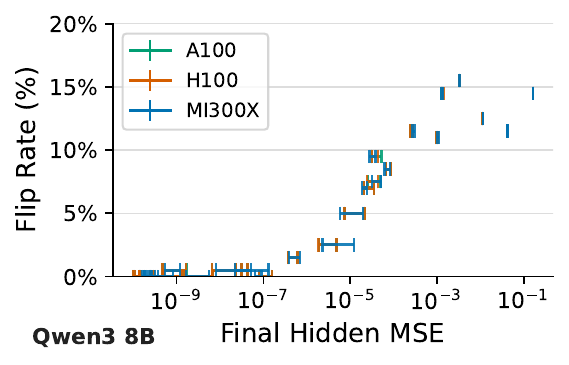}
\includegraphics[width=0.48\columnwidth]{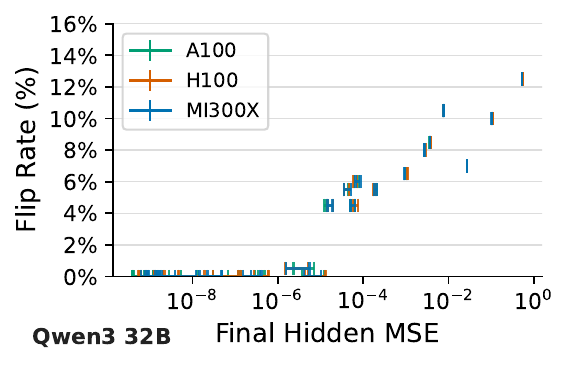}
\includegraphics[width=0.48\columnwidth]{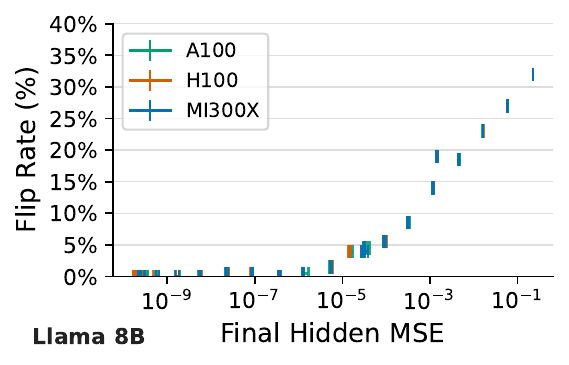}
\includegraphics[width=0.48\columnwidth]{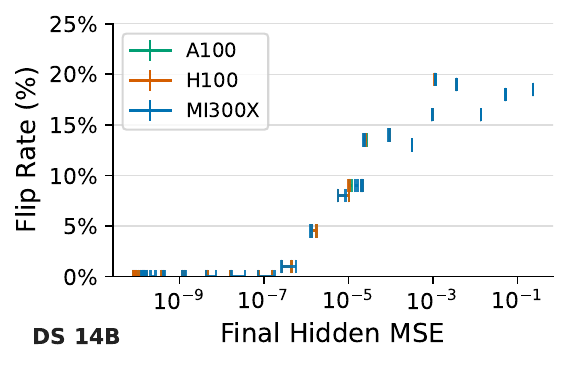}
\caption{Answer flip rate vs. total numerical error, complement of Figure \ref{fig:error-vs-flip}.}
\label{app:fig:error-vs-flip}
\end{center}
\end{figure}

Executing the 20 precision-controlled modes requires falling back to the FP32 execution path to apply manual truncation, which imposes a heavy demand on the hardware's FP32 throughput. Furthermore, auto-regressive generation inherently suffers from poor concurrency. To achieve acceptable hardware utilization and generation efficiency, a large batch size is strictly required. However, the 80GB memory capacity of the A100 and H100 GPUs severely limits the maximum batch size, making efficient generation impossible and resulting in prohibitively high execution times under these constraints. Consequently, we conduct the Flip Rate analysis exclusively on the AMD MI300X. Its massive 192GB memory capacity comfortably accommodates the large batch sizes needed to overcome the low concurrency of generation, while its superior FP32 throughput (163.4 TFLOPS, compared to 67 TFLOPS on H100 and 19.5 TFLOPS on A100) effectively absorbs the heavy overhead of manual truncation.

\begin{figure}
\begin{center}
\includegraphics[width=0.48\columnwidth]{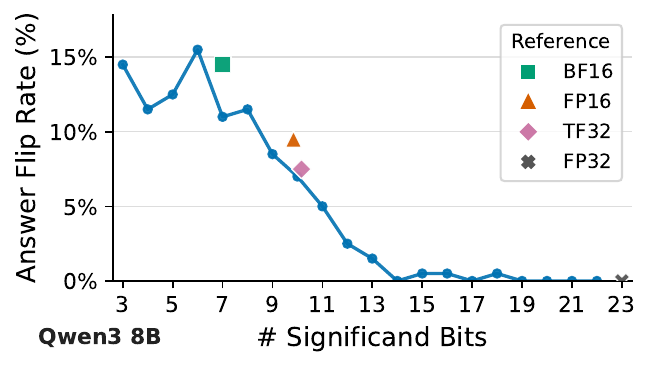}
\includegraphics[width=0.48\columnwidth]{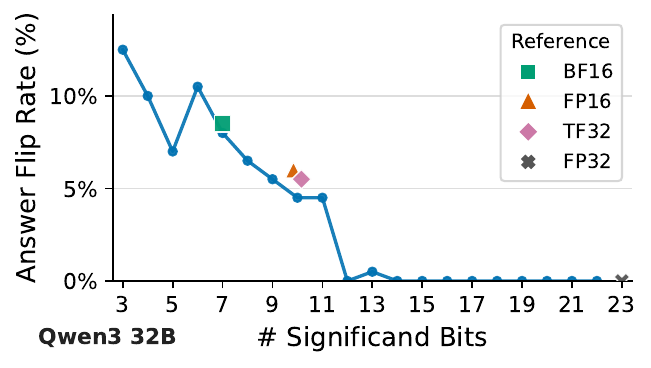}
\includegraphics[width=0.48\columnwidth]{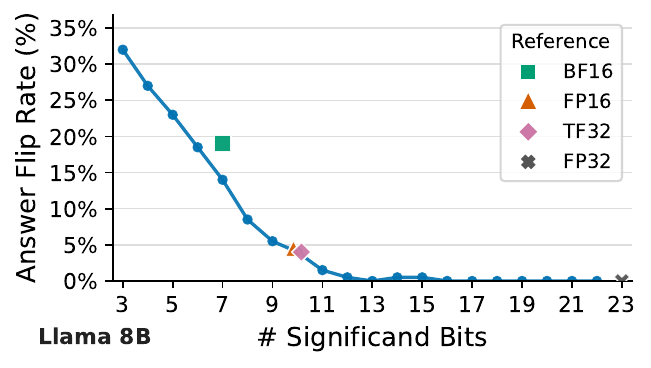}
\includegraphics[width=0.48\columnwidth]{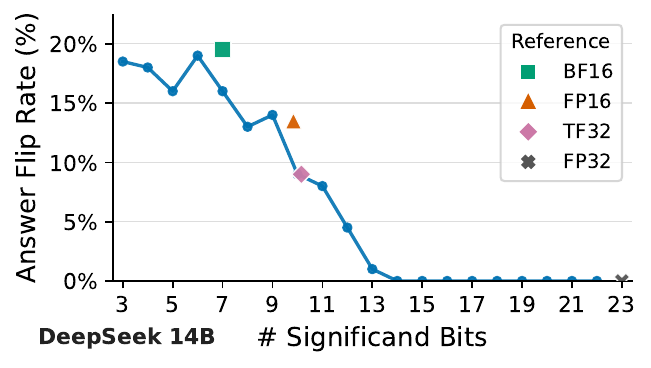}
\caption{Answer flip rate vs. number of significand bits, complement of Figure \ref{fig:mantissa-vs-flip}.}
\label{app:fig:mantissa-vs-flip}
\end{center}
\end{figure}

% \todoinline{This transition directly moves into the results and figures}
Figure \ref{fig:error-vs-flip} and Figure \ref{app:fig:error-vs-flip} correlate these two metrics to visualize how incremental precision loss at the kernel level eventually manifests as discrete token divergence. Notably, the empirical results indicate that under a fixed precision configuration, the final hidden MSE remains hardware-agnostic. Figure \ref{fig:mantissa-vs-flip} and Figure \ref{app:fig:mantissa-vs-flip} further isolate the impact of significand bit width on the stability of the model output. Table \ref{tab:fig1-fig4-execution-time} summarizes the execution time required for both analysis pipelines.

\MyPara{Observations on High-Precision Flips.} Through manual inspection, we observed that a substantial portion of the answer flips, especially those occurring under relatively high-precision configurations, generally originate from highly unstable questions. For these specific questions, the model typically exhibits severe uncertainty in its reasoning trace, often debating between multiple choices or doubting the validity of all options. This extreme hesitation can easily derail the auto-regressive generation, sometimes causing the model to degenerate into a self-repeating loop or simply fail to reach a definitive conclusion, ultimately leading to a parse failure (F). Under such inherently unstable conditions, even microscopic numerical perturbations can easily alter the decoding trajectory. Consequently, while these instances contribute to the overall flip rate, they primarily expose the model's limited capability on these challenging questions rather than systemic precision issues. This observation suggests that accurately measuring numerical reproducibility requires selecting more carefully curated datasets where the model is confident, rather than evaluating on questions that push its capability boundaries.

\subsection{Error Contribution Analysis (Figure~\ref{fig:error-contribution})}
\label{study:error-contribution} \label{study:fig2}
\MyPara{Error Decomposition.}
We first revisit the structural decomposition introduced in \S\ref{sec:propagation-error}. For decoder layer $i$, let $\mathbf{H}_i$ denote the pre-Attention residual stream, $\mathbf{A}_i$ denote the post-Attention residual stream, and $\mathbf{H}_{i+1}$ denote the next-layer input after the MLP block. Hatted tensors denote the corresponding candidate low-precision execution, while unhatted tensors denote the FP32 LayerCast reference. Each layer comprises three primary mappings susceptible to numerical-error injection: the KV-write mapping $\mathbf{H}_i \rightarrow (\mathbf{K}_i, \mathbf{V}_i)$, the Attention mapping $(\mathbf{H}_i, \mathbf{C}^{(i)}) \rightarrow \mathbf{A}_i$ (where $\mathbf{C}^{(i)}$ is the KV cache consumed by Attention), and the MLP mapping $\mathbf{A}_i \rightarrow \mathbf{H}_{i+1}$. We can then decompose the error as:

\begin{equation}
\begin{array}{c @{\qquad} c @{\qquad} l}
    \hat{\mathbf{A}}_i - \mathbf{A}_i
    & = &
    \epsilon^{\text{A}}_{\text{new}}
    +
    \epsilon^{\text{A}}_{\text{kv}}
    +
    \epsilon^{\text{A}}_{\text{old}}, \\[0.6ex]
    \hat{\mathbf{H}}_{i+1} - \mathbf{H}_{i+1}
    & = &
    \epsilon^{\text{M}}_{\text{new}}
    +
    \epsilon^{\text{M}}_{\text{old}}, \\[0.6ex]
    (\hat{\mathbf{K}}_i,\hat{\mathbf{V}}_i) - (\mathbf{K}_i,\mathbf{V}_i)
    & = &
    \epsilon^{\text{K}}_{\text{new}}
    +
    \epsilon^{\text{K}}_{\text{old}} .
\end{array}
\end{equation}
Here the superscripts $\text{A}$, $\text{M}$, and $\text{K}$ denote Attention, MLP, and KV-write mappings, respectively $\epsilon_{\text{new}}$ denotes the newly injected local numerical error, $\epsilon_{\text{old}}$ denotes the error inherited from the mapping input, and $\epsilon_{\text{kv}}$ denotes the horizontal error inherited through the KV cache.

\MyPara{Error Measurement.} We measure these error propagation routes using Teacher Forcing against the H100 FP32 LayerCast reference trajectory, comparing the newly generated candidate tensors against the reference tensors at identical layer and token positions. Using the notation above, for each layer $i$ and token position $t$, we record the following scalar MSE measurements:
\begin{align}
    E_{\text{in}}(i,t)
    &=
    \text{MSE}(\widehat{\mathbf{H}}_{i,t}, \mathbf{H}_{i,t}), \\
    E_{\text{post}}(i,t)
    &=
    \text{MSE}(\widehat{\mathbf{A}}_{i,t}, \mathbf{A}_{i,t}), \\
    E_{\text{next}}(i,t)
    &=
    \text{MSE}(\widehat{\mathbf{H}}_{i+1,t}, \mathbf{H}_{i+1,t}), \\
    E_{\text{K}}(i,t)
    &=
    \frac{1}{2}
    \left[
        \text{MSE}(\widehat{\mathbf{K}}_{i,t}, \mathbf{K}_{i,t})
        +
        \text{MSE}(\widehat{\mathbf{V}}_{i,t}, \mathbf{V}_{i,t})
    \right], \\
    E_{\text{pK}}(i,t)
    &=
    \frac{1}{t+1}
    \sum_{s=0}^{t}
    E_{\text{K}}(i,s).
\end{align}
Here $E_{\text{in}}$, $E_{\text{post}}$, and $E_{\text{next}}$ measure the accumulated residual-stream error before the Attention, after the Attention, and after the MLP, respectively. $E_{\text{K}}$ measures the current-token KV-cache error, while $E_{\text{pK}}$ measures the prefix-averaged KV error seen by Attention over $\mathbf{C}^{(i)}_{\le t}$. For the final decoder layer, $E_{\text{next}}$ is derived from the output MSE of that layer. We additionally track two local error growth terms:
\begin{align}
    J_{\text{A}}(i,t)
    &=
    \text{MSE}
    \left(
        \widehat{\mathbf{A}}_{i,t}-\widehat{\mathbf{H}}_{i,t},
        \mathbf{A}_{i,t}-\mathbf{H}_{i,t}
    \right), \\
    J_{\text{M}}(i,t)
    &=
    \text{MSE}
    \left(
        \widehat{\mathbf{H}}_{i+1,t}-\widehat{\mathbf{A}}_{i,t},
        \mathbf{H}_{i+1,t}-\mathbf{A}_{i,t}
    \right).
\end{align}
These growth terms, $J_{\text{A}}$ and $J_{\text{M}}$, isolate the error increments introduced natively by the Attention and MLP updates. Although these updates operate on tensors already harboring inherited numerical errors, such inherited perturbations are typically orders of magnitude smaller than the original hidden activations. Consequently, the error behavior of the current update is predominantly governed by the original activation values rather than by coherent interactions with errors from preceding layers. This characteristic justifies approximating the accumulated residual-flow MSE by summing the layer-wise local growth MSEs. In practice, utilizing $J_{\text{A}}$ and $J_{\text{M}}$ provides a more stable estimation of each layer's intrinsic error contribution for our analysis.

\MyPara{Error Propagation Model.} For each layer, we fit the replayed token rows with the following first-order approximation:
\begin{align}
    E_{\text{K}}(i,t)
    &\approx
    \lambda_{\text{K}}(i) E_{\text{in}}(i,t)
    +
    \beta_{\text{K}}(i), \\
    J_{\text{A}}(i,t)
    &\approx
    \lambda_{\text{in}}(i) E_{\text{in}}(i,t)
    +
    \lambda_{\text{pK}}(i) E_{\text{pK}}(i,t)
    +
    \beta_{\text{A}}(i), \\
    J_{\text{M}}(i,t)
    &\approx
    \lambda_{\text{M}}(i) E_{\text{post}}(i,t)
    +
    \beta_{\text{M}}(i).
\end{align}
The replay rows contain token-level stochastic variation, so the fitted model estimates the average trend of each propagation relation rather than enforcing a point-wise equality. The slope terms $\lambda$ describe how inherited residual or prefix-KV error is transported to the next stage, while the intercept terms $\beta$ estimate layer-local source magnitudes.

\MyPara{Robust Fit.} The replay rows occasionally contain extreme MSE outliers, which can make standard least-squares estimators unstable. We therefore estimate the propagation slopes with a Theil--Sen robust fit~\cite{sen1968estimates}. For a single-feature relation, the slope is estimated from the median pairwise trend:
\begin{equation}
    \tilde{\lambda}
    =
    \text{median}_{r<s,\,\tilde{x}_r\ne\tilde{x}_s}
    \frac{\tilde{y}_r-\tilde{y}_s}{\tilde{x}_r-\tilde{x}_s}.
\end{equation}
For the two-feature Attention relation, we estimate the two slopes by alternating Theil--Sen updates until the slopes stabilize. At each iteration, we first subtract the current prefix-KV term and update the residual-stream slope from the remaining residual, then subtract the updated residual-stream term and update the prefix-KV slope from the new residual. This alternating procedure captures the marginal effect of each feature after accounting for the other feature. After the slopes have stabilized, the remaining average offset is used as the fitted intercept \(\beta\), which represents the local source magnitude not explained by inherited residual or prefix-KV error.

\begin{figure}[t!]
\begin{center}
\includegraphics[width=0.48\columnwidth]{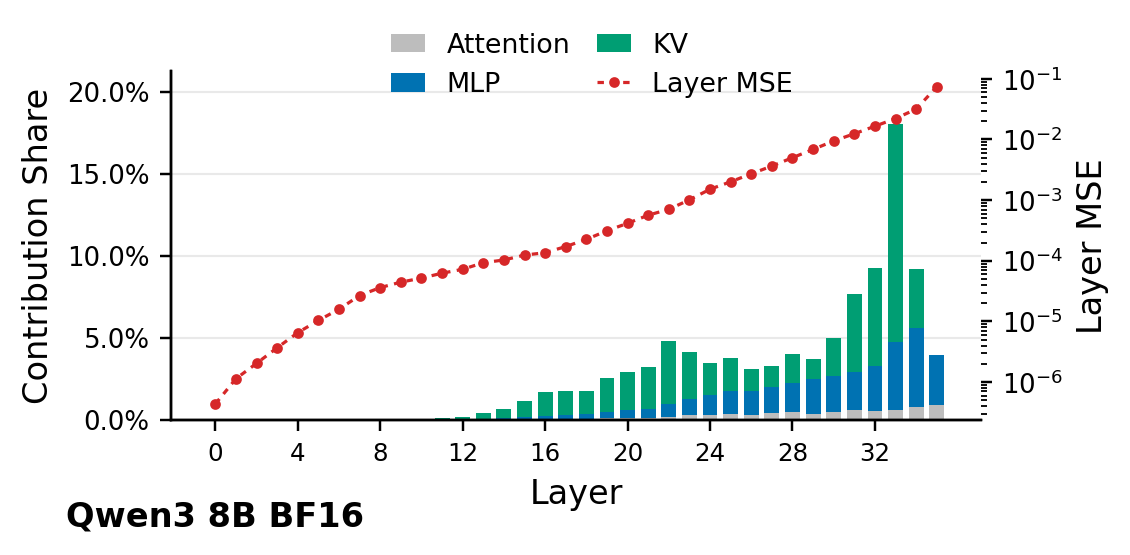}
\includegraphics[width=0.48\columnwidth]{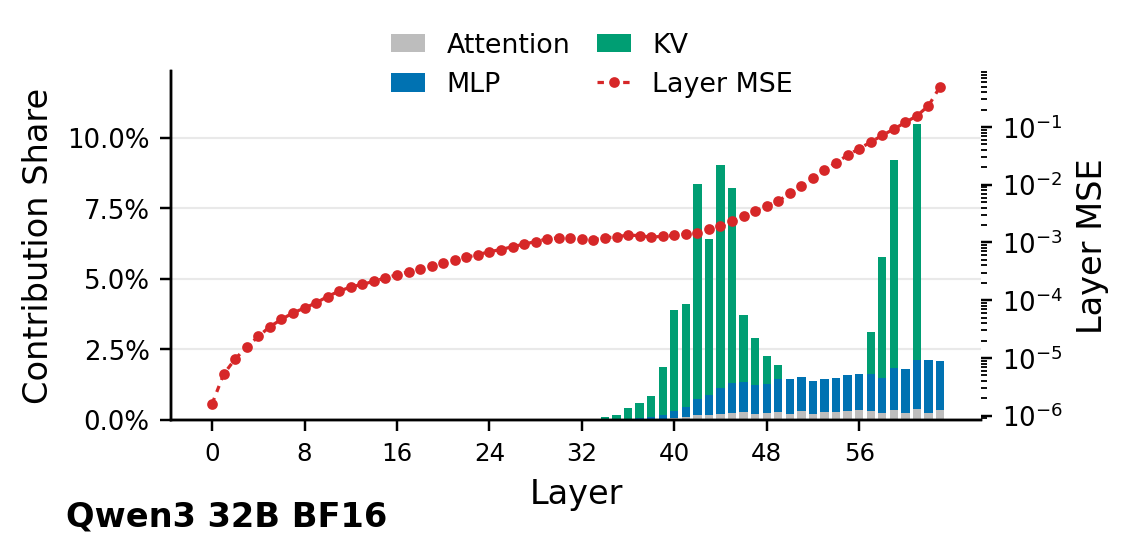}
\includegraphics[width=0.48\columnwidth]{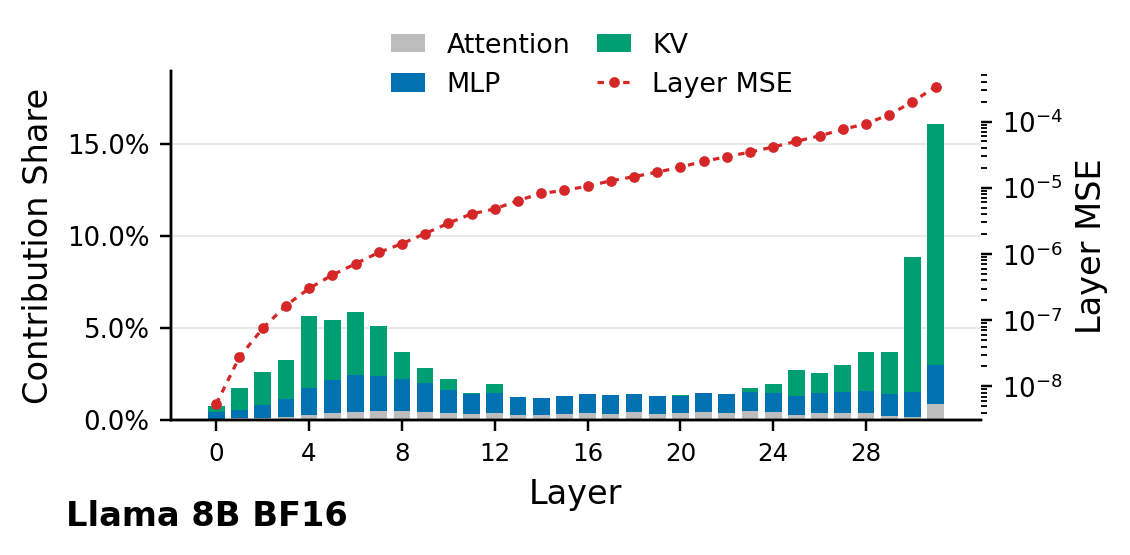}
\includegraphics[width=0.48\columnwidth]{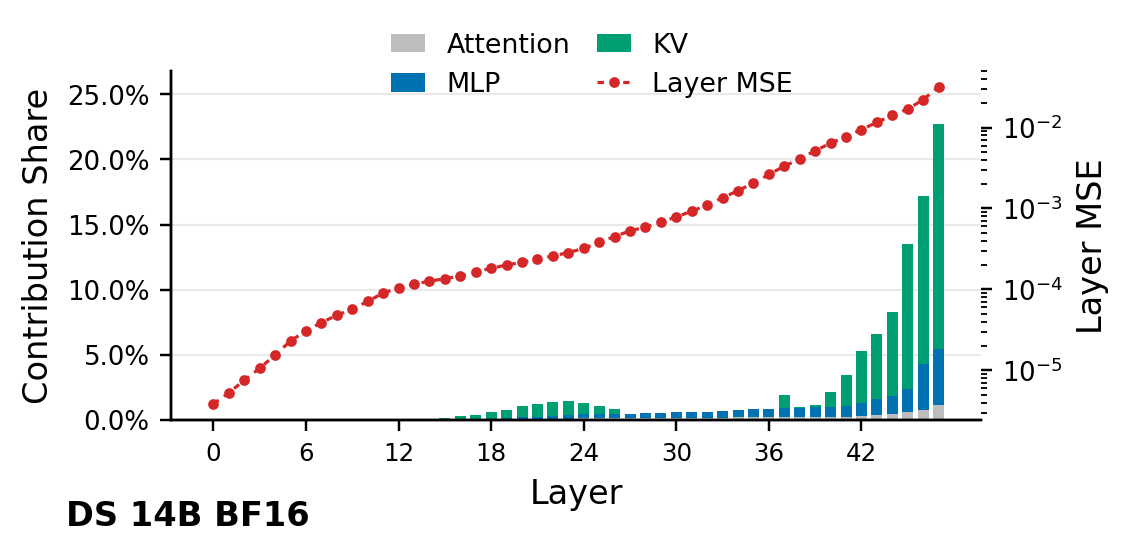}
\caption{Contribution of error across network depth for the supplementary models, complementing Figure~\ref{fig:error-contribution}.}
\label{app:fig:error-contribution}
\end{center}
\end{figure}

\MyPara{Error Share Attribution.} We convert the fitted local source magnitudes into local shares by normalizing them with the expected tensor MSE at the corresponding stage over replay rows:
\begin{equation}
    s_{\text{A}}(i)
    =
    \frac{\beta_{\text{A}}(i)}
    {\mathbb{E}_{t}\!\left[E_{\text{post}}(i,t)\right]},
    \qquad
    s_{\text{K}}(i)
    =
    \frac{\lambda_{\text{pK}}(i)\beta_{\text{K}}(i)}
    {\mathbb{E}_{t}\!\left[E_{\text{post}}(i,t)\right]},
    \qquad
    s_{\text{M}}(i)
    =
    \frac{\beta_{\text{M}}(i)}
    {\mathbb{E}_{t}\!\left[E_{\text{next}}(i,t)\right]}.
\end{equation}
Here $s_{\text{A}}$, $s_{\text{K}}$, and $s_{\text{M}}$ denote the local Attention-compute, KV-cache, and MLP-compute shares, respectively. The KV source is multiplied by $\lambda_{\text{pK}}$ because newly written KV error affects the residual stream only after it is consumed by Attention.

We then recursively allocate the final hidden-state error share from the last layer to the first. Let $\rho$ be the currently unexplained share arriving from deeper layers, initialized as $\rho=1$. At layer $i$, the MLP contribution is assigned first:
\begin{equation}
    c_{\text{M}}(i)
    =
    \rho s_{\text{M}}(i),
    \qquad
    \rho_{\text{A}}(i)
    =
    \rho\left(1-s_{\text{M}}(i)\right).
\end{equation}
The remaining share then enters the Attention side, where Attention computes, and KV-cache contributions are assigned as
\begin{equation}
    c_{\text{A}}(i)
    =
    \rho_{\text{A}}(i)s_{\text{A}}(i),
    \qquad
    c_{\text{K}}(i)
    =
    \rho_{\text{A}}(i)s_{\text{K}}(i).
\end{equation}
The unassigned share is propagated to the previous layer:
\begin{equation}
    \rho
    \leftarrow
    \rho_{\text{A}}(i)
    \left(
        1-s_{\text{A}}(i)-s_{\text{K}}(i)
    \right).
\end{equation}
This process assigns the remaining error mass to the local sources encountered along the backward residual flow, and produces the layer-wise attribution scores \(c_{\text{A}}(i)\), \(c_{\text{M}}(i)\), and \(c_{\text{K}}(i)\).

\MyPara{Figure Construction.} Figure~\ref{fig:error-contribution} reports the recursive attribution for Qwen3-14B under native BF16 execution, and Figure~\ref{app:fig:error-contribution} reports the corresponding results for the supplementary models. The plotted layer-wise shares are \(c_{\text{A}}(i)\), \(c_{\text{M}}(i)\), and \(c_{\text{K}}(i)\), shown as gray, blue, and green bars for Attention compute, MLP compute, and KV-cache contribution, respectively. The red curve on the secondary axis reports the layer-ending MSE \(\mathbb{E}_{t}\!\left[E_{\text{next}}(i,t)\right]\). We compute these figures on a 32-example MedQA calibration split, with at most 1024 generated tokens replayed per example.

\subsection{Truncation vs Arithmetic (Figure \ref{fig:quant-vs-arith})}
\label{study:fig3}

\begin{figure}
\begin{center}
\includegraphics[width=0.48\columnwidth]{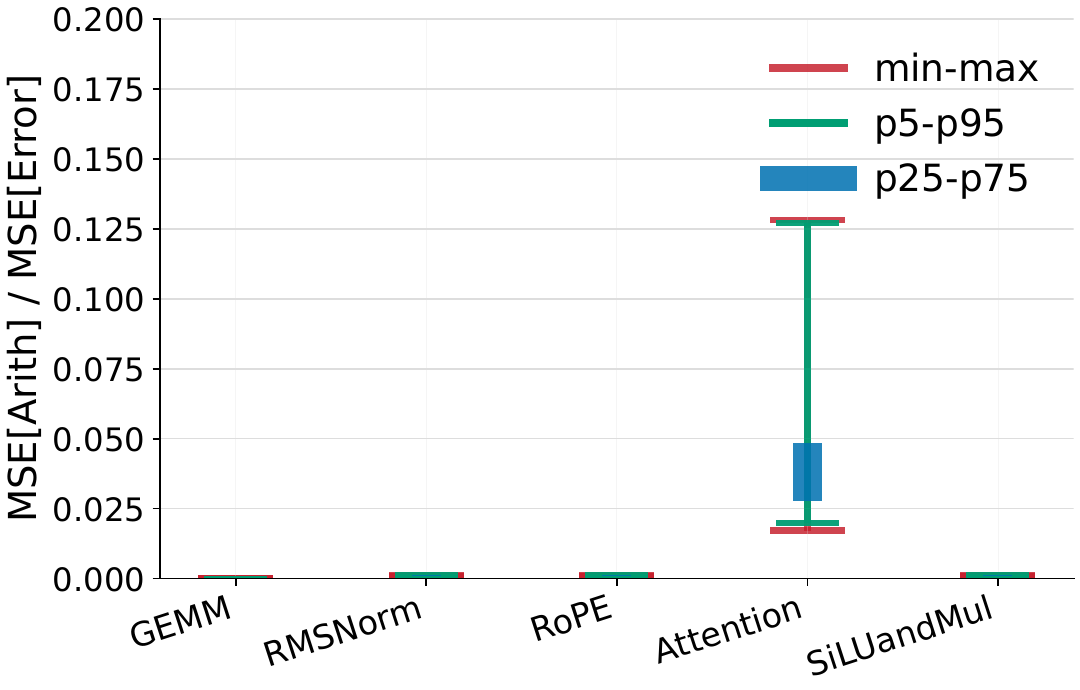}
\includegraphics[width=0.48\columnwidth]{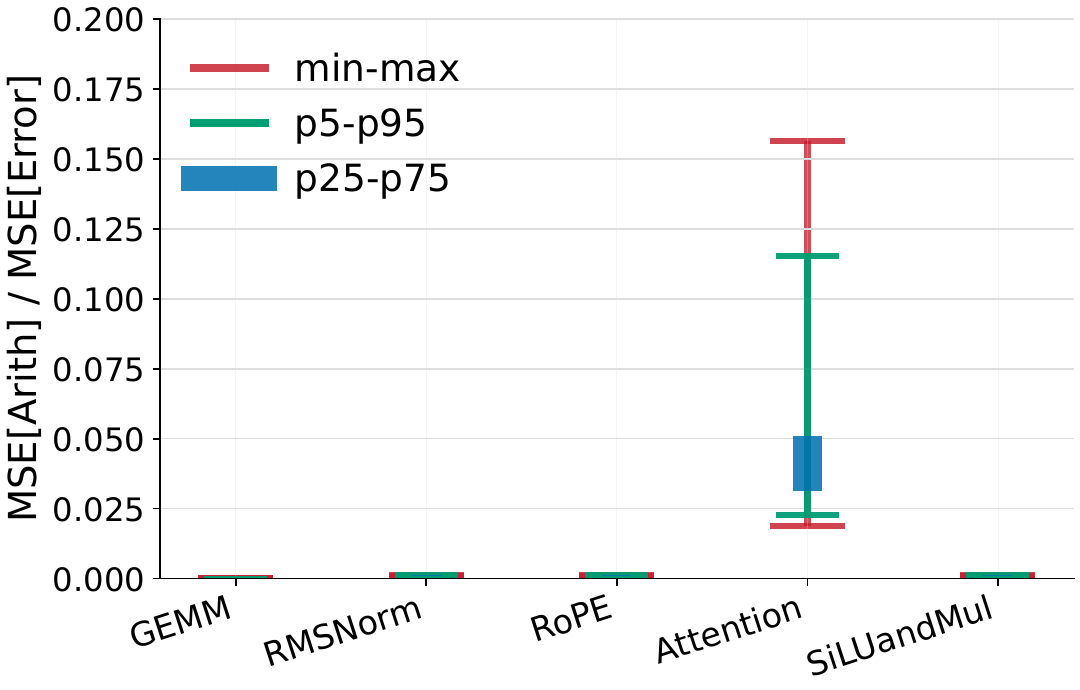}
\includegraphics[width=0.48\columnwidth]{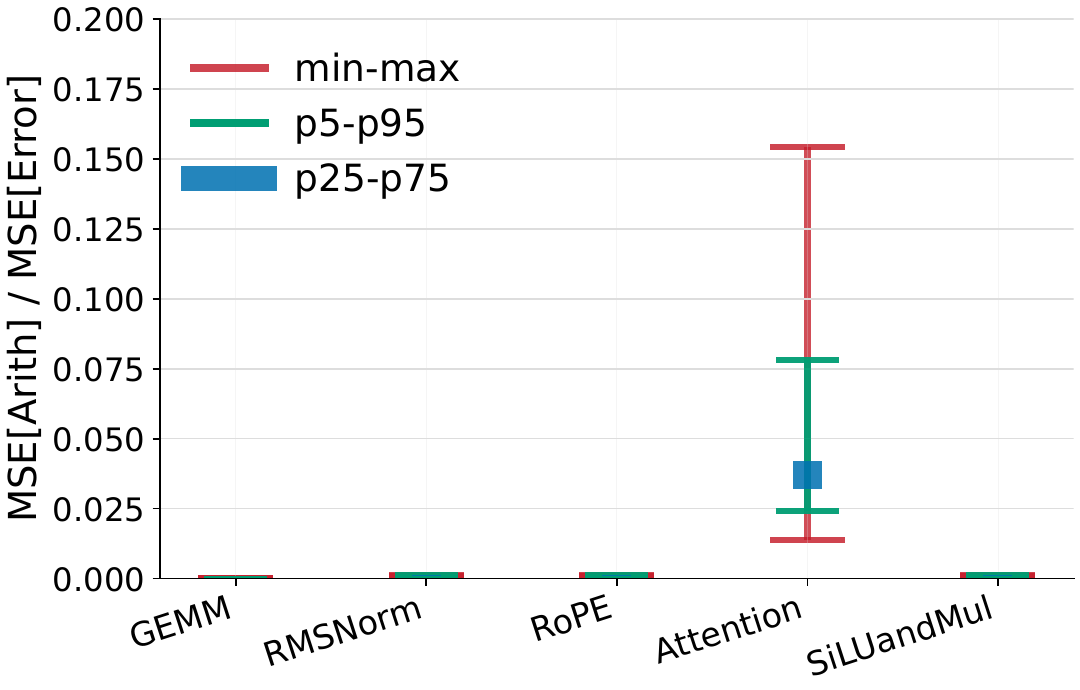}
\includegraphics[width=0.48\columnwidth]{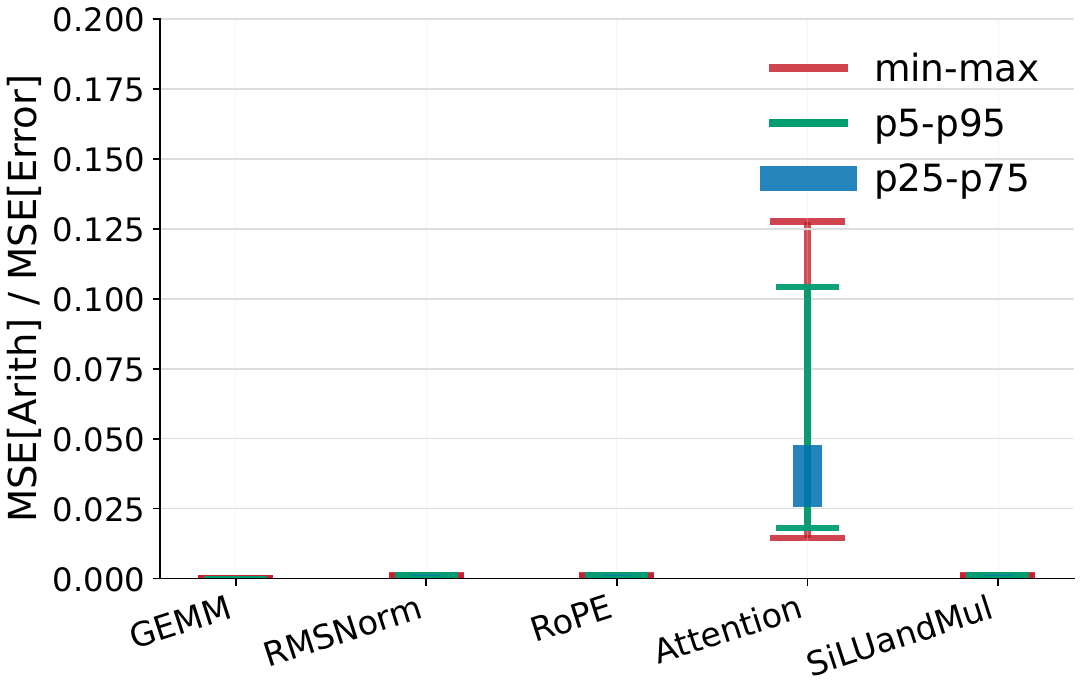}
\caption{The $\mathbb{E}[\epsilon_{\text{arith}}^2] / \mathbb{E}[(\hat{y} - y)^2]$ ratio for all components, complement of Figure \ref{fig:quant-vs-arith}.}
\label{app:fig:quant-vs-arith}
\end{center}
\end{figure}

To quantify the exact source of numerical noise, we empirically isolate the truncation error ($\epsilon_{\text{trunc}}$) from the arithmetic error ($\epsilon_{\text{arith}}$). Specifically, for any given operator $ \mathcal{F} $ and its exact FP32 input $x$, we evaluate three distinct execution paths:
\begin{itemize}
    \item \textbf{Reference FP32:} $y = \mathcal{F}_{\text{FP32}}(x)$.
    \item \textbf{Native Low-Precision (Total Error):} $\hat{y} = \mathcal{Q}_{\text{BF16}}(\mathcal{F}_{\text{BF16}}(\mathcal{Q}_{\text{BF16}}(x)))$, representing the standard execution utilizing the native vLLM BF16 path.
    \item \textbf{Truncation-Only (Control):} $\tilde{y} = \mathcal{Q}_{\text{BF16}}(\mathcal{F}_{\text{FP32}}(\mathcal{Q}_{\text{BF16}}(x)))$, a strictly controlled implementation where inputs and outputs are quantized, but all internal kernel arithmetic is securely performed in full FP32.
\end{itemize}

Based on these execution paths, we measure three corresponding Mean Squared Error (MSE) terms: the total error $E = \mathbb{E}[(\hat{y} - y)^2]$, the isolated truncation error $T = \mathbb{E}[(\tilde{y} - y)^2]$, and the direct arithmetic discrepancy $L = \mathbb{E}[(\hat{y} - \tilde{y})^2]$.

Rather than assuming strict additive independence between quantization and arithmetic operations, we employ a two-factor Shapley value decomposition to fairly attribute the final error. In this formulation, we assess the marginal contributions of the arithmetic operations under two different cooperative conditions:
\begin{itemize}
    \item \textbf{Residual Contribution:} The marginal error introduced by the arithmetic operations given that truncation has already occurred, calculated as $(E - T) / E$.
    \item \textbf{Direct Contribution:} The marginal error of the arithmetic operations evaluated directly against the truncation-only baseline, calculated as $L / E$.
\end{itemize}

The Shapley arithmetic contribution is the exact average of these two marginal effects. Consequently, the normalized contribution ratio plotted in our analysis (i.e., the y-axis in Figure \ref{fig:quant-vs-arith} and \ref{app:fig:quant-vs-arith}) is formally derived as:
\begin{equation}
\frac{\text{Shapley}[\text{Arith}]}{\text{MSE}[\text{Error}]} = \frac{1}{2} \left( \frac{E - T}{E} + \frac{L}{E} \right) = \frac{E - T + L}{2E}    
\end{equation}

To construct the profiling distribution, we systematically evaluate every operator instance across all decoder layers. Specifically, we randomly sample 16 questions from our MedQA 200 subset and trace the forward passes for the first 64 generated tokens. During the reference FP32 execution, we record the exact input tensors for each operator. We then independently replay these recorded inputs, compute the expected MSE values across all 1,024 token positions ($16 \times 64$) for all operators, and calculate their corresponding Shapley ratios. The boxplots in Figure \ref{app:fig:quant-vs-arith} illustrate the distribution of these computed ratios across all instances belonging to the same operator family.

As clearly shown in the distribution (Figure \ref{fig:quant-vs-arith} and Figure \ref{app:fig:quant-vs-arith} for the additional models), the arithmetic error is negligible for almost all fundamental kernels, with Attention being the sole exception. We break down the root cause for each kernel family below, perfectly corroborating our instruction-level profiling in Table \ref{tab:kernel_instructions}:

\MyPara{GEMM.} For matrix multiplications, while the input and output tensors are stored in BF16, the internal dot-product accumulation within the Tensor Cores is natively executed in FP32. Consequently, the minor numerical variance introduced by the non-associativity of floating-point accumulation order is completely absorbed by the massive precision loss that occurs when the final 32-bit registers are cast back down to memory storage at the kernel boundary.

\MyPara{RMSNorm, RoPE, and SiLUAndMul.} A superficial inspection of the vLLM implementation might suggest that these operations are executed natively in BF16. However, whether it is the reduction and accumulation steps required for variance computation in RMSNorm, or the fundamental arithmetic required for computing activations (SiLUAndMul) and rotational embeddings (RoPE), the actual ALU execution is universally upcast to FP32. Tracing down to the PyTorch~\cite{paszke2019pytorch} ATen bindings (the underlying framework vLLM relies on), we find that applying basic arithmetic to BF16 tensors triggers an automatic upcast to Float32. Specifically, examining the underlying \texttt{c10::BFloat16} implementation confirms that the overloaded addition, subtraction, and multiplication operators explicitly cast operands to single-precision:

\begin{lstlisting}[
  language=C++, 
  basicstyle=\ttfamily\small, 
  morekeywords={BFloat16, C10_HOST_DEVICE}, 
  frame=single, 
  rulecolor=\color{black!30},
  aboveskip=1em,
  belowskip=1em
]
inline C10_HOST_DEVICE BFloat16
operator+(const BFloat16& a, const BFloat16& b) {
  return static_cast<float>(a) + static_cast<float>(b);
}

inline C10_HOST_DEVICE BFloat16
operator-(const BFloat16& a, const BFloat16& b) {
  return static_cast<float>(a) - static_cast<float>(b);
}

inline C10_HOST_DEVICE BFloat16
operator*(const BFloat16& a, const BFloat16& b) {
  return static_cast<float>(a) * static_cast<float>(b);
}
\end{lstlisting}

This software-level upcasting behavior exactly mirrors our FP32 reference and is consistent across both NVIDIA A100 and H100 architectures. It directly maps to the \texttt{FADD} and \texttt{FMUL} 32-bit PTX instructions observed in our hardware profiling (Table \ref{tab:kernel_instructions}), leaving boundary truncation as the only practical source of numerical divergence.

\MyPara{Attention.} The Attention kernel is the distinct exception, exhibiting a non-negligible arithmetic error fraction. This anomaly is driven by a two-fold mechanism. Primarily, unlike standard kernels that strictly defer quantization to the final memory write, the complex multi-stage nature of fused Attention introduces intermediate intra-kernel truncations that do not occur at the kernel boundary. Secondarily, this instability is exacerbated by a compiler-induced behavior where Triton's \texttt{tl.dot} operation silently falls back to the TF32 format on modern GPUs. This aggressive fallback restricts the arithmetic precision to only 10 bits and transforms what should be a strict boundary issue into a compounding in-kernel arithmetic error.

In summary, this analysis definitively proves that with the exception of the Attention kernel's intra-kernel truncations and TF32 fallback, end-to-end non-reproducibility is not caused by the internal arithmetic sequence, but is strictly a kernel boundary truncation problem.

\subsection{QKV Distribution (Figure \ref{fig:qkv-distribution})}

\begin{figure}
\begin{center}
\includegraphics[width=0.48\columnwidth]{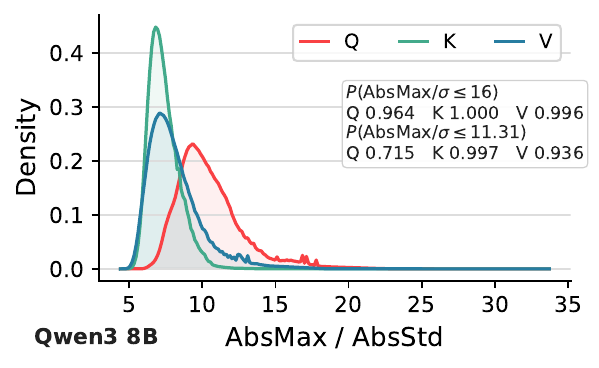}
\includegraphics[width=0.48\columnwidth]{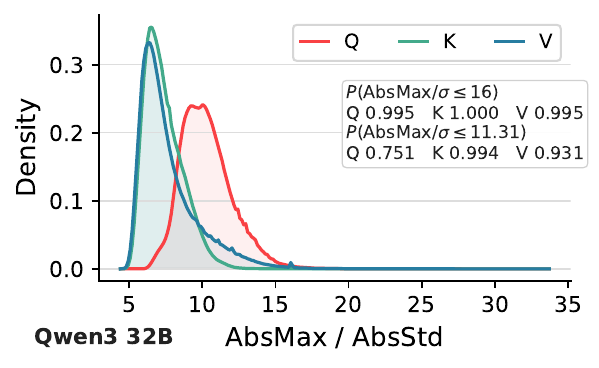}
\includegraphics[width=0.48\columnwidth]{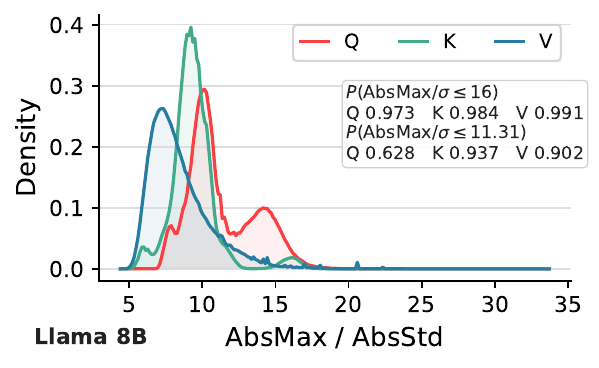}
\includegraphics[width=0.48\columnwidth]{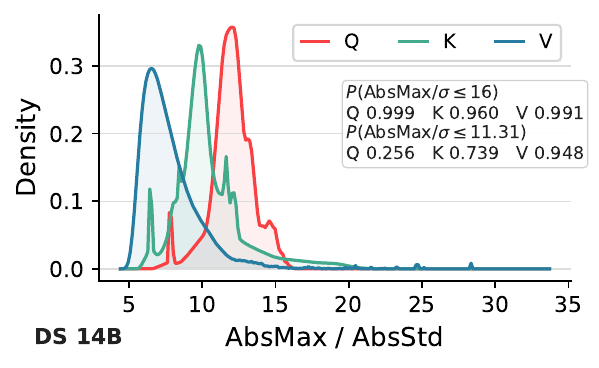}
\caption{QKV distribution, complement of Figure \ref{fig:qkv-distribution}.}
\label{app:fig:qkv-distribution}
\end{center}
\end{figure}

We evaluate the pre-Attention Query (Q), Key (K), and Value (V) tensors under native BF16 execution by computing the ratio of the absolute maximum value to the standard deviation ($\text{AbsMax}/\sigma$) for each tensor. The corresponding probability density distributions are plotted across all tokens within the complete MedQA 200 subset. Figure \ref{fig:qkv-distribution} and Figure \ref{app:fig:qkv-distribution} illustrate these distributions.

\subsection{Comparison with Quantization Method}
\label{study:quant}

Unlike weight-activation and KV-cache quantization, weight-only quantization (\eg, GPTQ~\cite{frantar2023gptq} and AWQ~\cite{lin2024awq}) compresses only static parameters and can be directly stacked with \techname. To evaluate this composability, we apply \techname to W4A16 checkpoints (with group size 128) across all five models on the full MedQA test set~\cite{jin2021disease} (1273 questions) on A100. \techname runs on the dequantized FP16 weights, which are bit-identical to those multiplied by the native W4A16 kernel, so both paths compute the exact same function. To avoid conflating quantization accuracy loss with reproducibility failure, we anchor each checkpoint against its own weights executed in FP32 with LayerCast~\cite{yuan2025understanding}.

\begin{table*}[ht]
\centering
\caption{Answer flip rate and task accuracy of weight-only quantized models (W4A16) on the MedQA test set (1{,}273 questions, A100). Flip rates are measured against each checkpoint's own FP32 execution.}
\label{tab:quant}
\resizebox{\textwidth}{!}{
\begin{tabular}{l|cc|cc|cc|cc|cc}
\toprule
\multirow{2}{*}{Configuration} & \multicolumn{2}{c|}{Qwen3-8B} & \multicolumn{2}{c|}{Qwen3-14B} & \multicolumn{2}{c|}{Qwen3-32B} & \multicolumn{2}{c|}{Llama-8B} & \multicolumn{2}{c}{DeepSeek-14B} \\
 & Flip Rate & Accuracy & Flip Rate & Accuracy & Flip Rate & Accuracy & Flip Rate & Accuracy & Flip Rate & Accuracy \\
\midrule
AWQ (FP32 reference) & --- & 76.36\% & --- & 80.99\% & --- & 85.00\% & --- & 67.32\% & --- & 72.43\% \\
AWQ (W4A16)          & 9.43\% & 75.81\% & 5.81\% & 81.78\% & 6.60\% & 85.23\% & 5.66\% & 68.66\% & 14.53\% & 72.43\% \\
AWQ + \techname      & 0.31\% & 76.59\% & 0.08\% & 80.99\% & 0.00\% & 85.00\% & 0.16\% & 67.32\% & 0.31\% & 72.43\% \\
\midrule
GPTQ (FP32 reference) & --- & 75.57\% & --- & 80.91\% & --- & 86.33\% & --- & 65.91\% & --- & 72.03\% \\
GPTQ (W4A16)          & 10.68\% & 75.33\% & 5.97\% & 81.38\% & 7.54\% & 86.96\% & 4.40\% & 66.22\% & 13.83\% & 72.35\% \\
GPTQ + \techname      & 0.31\% & 75.49\% & 0.16\% & 80.75\% & 0.39\% & 86.49\% & 0.08\% & 65.91\% & 0.16\% & 72.03\% \\
\bottomrule
\end{tabular}
}
\end{table*}

As shown in Table~\ref{tab:quant}, native W4A16 execution still suffers from 4.4--14.5\% answer flip rates. Stacking \techname on the same checkpoints reduces the flip rate to 0--0.39\% and lowers the Final Hidden MSE on MedQA-200 from $1.6\times10^{-5}$--$2.1\times10^{-4}$ to $1.2\times10^{-8}$--$9.9\times10^{-8}$, while keeping task accuracy within 0.25 percentage points of the FP32 reference.

\subsection{Model Quality Preservation}
\label{app:quality}

To verify that \techname does not introduce systematic quality degradation, we compare \techname against the native BF16 baseline on Qwen3-14B across perplexity, general knowledge, instruction following, mathematical reasoning, coding, multilingual understanding, and long-context tasks, alongside reference numbers from the Qwen3 Technical Report~\cite{yang2025qwen3}. Following the Technical Report (Table~15), downstream benchmarks are evaluated in thinking mode using the official sampling settings, whereas WikiText-2~\cite{merity2016wikitext2} is evaluated via raw likelihood scoring with thinking disabled. As shown in Table~\ref{tab:quality}, \techname matches the native BF16 baseline within sampling variance across all benchmarks; the largest deviation (on LiveCodeBench-v5) remains within 1.5 standard errors across its 167 problems and slightly favors \techname.

\begin{table}[ht]
\centering
\small
\caption{Model quality of Qwen3-14B under native BF16 and \techname.}
\label{tab:quality}
\resizebox{\textwidth}{!}{
\begin{tabular}{l|l|r|r|r}
\toprule
Benchmark / metric & Task & Tech Report & BF16 & \techname \\
\midrule
WikiText-2 PPL $\downarrow$ & Language modeling, full test split & --- & 11.1106 & 11.1026 \\
MMLU-Redux Acc.\ $\uparrow$~\cite{gema2025mmluredux} & General knowledge, 5330 examples & 88.6\% & 89.72\% & 89.81\% \\
IFEval Strict Acc.\ $\uparrow$~\cite{zhou2023ifeval} & Instruction following & 85.4\% & 87.06\% & 86.32\% \\
MATH-500 Pass@1 $\uparrow$~\cite{lightman2024verify} & Mathematical reasoning & 96.8\% & 95.6\% & 95.6\% \\
LiveCodeBench-v5 Pass@1 $\uparrow$~\cite{jain2025livecodebench} & Code generation & 63.5\% & 59.28\% & 64.67\% \\
MMMLU-14 Acc.\ $\uparrow$~\cite{openai2024mmmlu} & Multilingual knowledge, fixed 10\% subset & 77.9\% & 77.89\% & 77.84\% \\
\bottomrule
\end{tabular}
}
\end{table}

To assess long-context behavior, we evaluate a compact RULER~\cite{hsieh2024ruler} subset covering four task categories: needle retrieval (\texttt{niah\_single\_2}), variable tracking (\texttt{ruler\_vt}), frequent-word extraction (\texttt{ruler\_fwe}), and multi-hop QA (\texttt{ruler\_qa\_hotpot}). We evaluate two instances per task across six context lengths (48 instances in total) using greedy decoding with thinking disabled. As shown in Table~\ref{tab:ruler}, \techname achieves identical accuracy to native BF16 at every context length.

\begin{table}[ht]
\centering
\small
\caption{Long-context accuracy (\%) of Qwen3-14B on the four-task RULER subset.}
\label{tab:ruler}
\begin{tabular}{l|cccccc|c}
\toprule
 & 4K & 8K & 16K & 32K & 64K & 128K & Avg. \\
\midrule
BF16 & 87.50 & 87.50 & 83.33 & 83.33 & 79.17 & 79.17 & 83.33 \\
\techname & 87.50 & 87.50 & 83.33 & 83.33 & 79.17 & 79.17 & 83.33 \\
\bottomrule
\end{tabular}
\end{table}

% \subsection{Limited Structural Outliers}
% \label{study:outlier}

% In our experiments, we used a 50-size subset of the C4 \cite{2019t5} dataset and profiled the channels with maximal average squared value. Then, at GEMM, we complement the top 1\% channels with an additional FP16 to preserve the precision. Since as shown by \cite{sun2024massive} \cite{vllm_issue_40290}, the FP16 overflow is relatively rare, and we did not encounter it in our testing process.
\section{\techname{} Implementation}
\label{app:impl}

\techname{} achieves low overhead by improving precision at truncation boundaries while keeping heavy computation on efficient 16-bit Tensor Core paths and avoiding a substantial increase in memory footprint. The implementation is organized as a layered set of precision-protection mechanisms. \techname{}-Base first protects the main residual-flow path: hidden states, residual tensors, normalization inputs/outputs, and operator outputs are kept in FP32, while GEMMs still use FP16-path Tensor Core inputs and Attention remains on the native FP16 Attention path. \techname{}-Base+Attn builds on this FP32 residual-flow path and adds INT16-based Q/K/V protection for Attention, storing Q/K/V in compact INT16 form and unpacking them into dual-FP16 dot products during Attention computation. \techname{}-Base+Attn+Skip is the policy-optimized version of \techname{}-Base+Attn: under the error-attribution policy, it skips the extra INT16 Q/K/V protection on a selected number of initial layers, so those layers fall back to the \techname{}-Base Attention path rather than native BF16/FP16 execution. Finally, \techname{} applies FP16 promotion and algebraic compensation to GEMM inputs on top of \techname{}-Base+Attn, stores the KV cache with a grouped INT16 codec, and protects all layers.

\MyPara{\techname{}-Base.} \techname{}-Base keeps the main hidden states, residual stream, normalization inputs/outputs, and operator outputs in FP32, while leaving Attention on the native FP16 path. Its low overhead comes from two factors. First, the dominant GEMM operations still use hardware-optimized Tensor Core kernels: cuBLASLt~\cite{cublaslt} and hipBLASLt~\cite{hipBLASLt} expose GEMMs with 16-bit inputs and FP32 accumulation/output, so \techname{}-Base can keep GEMM outputs in FP32 while still relying on optimized vendor-library implementations. Second, the additional FP32 residual-flow operations account for a relatively small fraction of the end-to-end forward pass, so keeping these tensors in FP32 does not dominate serving latency. We also apply a per-weight-matrix power-of-two scaling before GEMM and invert the scale after the GEMM output, which places the weights in a better FP16 representable range and keeps the weight-representation error small. In ShareGPT serving evaluation, \techname{}-Base stays within \(-1.3\%\) to \(17.1\%\) TPOT overhead relative to native BF16 across models and hardware platforms. The slightly negative overhead occurs for Qwen3-32B on A100, which we attribute to normal system-level measurement fluctuation and report directly.

\MyPara{\techname{}-Base+Attn.} \techname{}-Base+Attn builds on \techname{}-Base and protects the Attention path. We re-implement a Triton version of Attention in which Q/K/V tensors are stored in compact INT16 form with scaling, unpacked into dual-FP16 pieces for the Attention dot products, and written to an FP32 output buffer. The GEMM, normalization, and residual paths remain the same as in \techname{}-Base. This design mainly reduces Q/K/V storage truncation error without substantially increasing the memory footprint. Although the dual-FP16 expansion adds extra dot-product work, the overhead is acceptable because Attention is largely memory-bound. In ShareGPT serving evaluation, \techname{}-Base+Attn stays within \(5.9\%\) to \(50.0\%\) TPOT overhead relative to native BF16 across models and hardware platforms.

\MyPara{\techname{}-Base+Attn+Skip.} \techname{}-Base+Attn+Skip is the policy-optimized version of \techname{}-Base+Attn. It keeps the same FP32 residual-flow path as \techname{}-Base and applies the same INT16-based Q/K/V protection as \techname{}-Base+Attn on most layers. Following the error-attribution policy, \techname{}-Base+Attn+Skip skips the extra Q/K/V protection on a selected number of initial layers, which fall back to the \techname{}-Base Attention path. In ShareGPT serving evaluation, \techname{}-Base+Attn+Skip stays within \(2.2\%\) to \(41.5\%\) TPOT overhead relative to native BF16 across models and hardware platforms.

\MyPara{\techname.} \techname builds on \techname{}-Base+Attn and further applies the FP16 algebraic compensation for GEMM as described in \S\ref{sec:solution}. This compensation represents each high-precision activation with two FP16 components and combines two FP16 Tensor Core GEMMs into an FP32 output, thereby recovering the residual information that would be lost by a single FP16 cast. We also implemented an INT8 compensation path, but its quantization overhead offsets the compute benefit of INT8 execution and introduces additional GPU memory cost. Therefore, the final design uses FP16 compensation. In ShareGPT serving evaluation, \techname ranges from \(70.1\%\) to \(119.9\%\) TPOT overhead relative to native BF16 across models and hardware platforms.

\MyPara{Grouped INT16 KV Codec.} Because a single tensor-wide scale wastes INT16 resolution when a few channels dominate in magnitude, \techname uses a lightweight grouped encoding for the KV cache. Before quantization, K and V are normalized using fixed per-channel statistics calibrated offline on a small ShareGPT subset (shared across all hardware platforms). Each 128-element K or V vector then stores a shared BF16 base scale alongside compact power-of-two sub-group shifts in a 4-byte metadata header, incurring only a 1.6\% storage overhead (520 bytes per KV head per token) over native 16-bit caches.

\MyPara{Error-Attribution Policy.} For \techname{}-Base+Attn+Skip, we use a small calibration set of 8 MedQA examples (replaying up to 1024 generated tokens per example) to determine which layers can safely skip INT16 Q/K/V protection. Using the median Final Hidden MSE as our validation metric to reduce sensitivity to extreme-error outliers, we require the accepted policy to stay within a \(1.2\times\) median-MSE budget relative to \techname{}-Base+Attn. For each model, we evaluate two candidate skipping orders: a \emph{contribution order} that skips the \(k\) lowest-contribution layers based on our error-attribution analysis, and a \emph{prefix order} that skips the initial \(k\) layers. We select the largest \(k\) that satisfies the budget and pick the order with the lower median MSE. In our calibration, prefix policies are consistently selected (reported in Table~\ref{tab:heal-skip-layers}), indicating that while error attribution reliably identifies which mechanism to optimize, coarse layer-level skipping works best on the earliest layers.

\begin{table}[ht]
\centering
\small
\caption{Number of skipped layers in \techname{}-Base+Attn+Skip.}
\label{tab:heal-skip-layers}
\resizebox{0.8 \textwidth}{!}{
\begin{tabular}{l|ccccc}
\toprule
Configuration & \textbf{Qwen3-8B} & \textbf{Qwen3-14B} & \textbf{Qwen3-32B} & \textbf{Llama-8B} & \textbf{DeepSeek-14B} \\
\midrule
\techname{}-Base+Attn+Skip & 11 & 14 & 13 & 5 & 15 \\
\bottomrule
\end{tabular}
}
\end{table}

\MyPara{Fine-Grained Channel Policy.} When extending this skipping policy to the complete \techname pipeline, we find that because \techname drives the overall residual error close to the FP32 noise floor, layer-level error attribution becomes too noisy to reliably guide whole-layer GEMM skipping without degrading precision. Instead of dropping entire layers, \techname supports a finer-grained per-layer policy that restricts dual-FP16 GEMM compensation in selected MLP projections to the top input channels ranked by residual energy: on Qwen3-14B, compensating an average of 42\% of input channels across 39 of the 80 MLP projections saves 12.7\% of GEMM FLOPs and 7.6\% of H100 TPOT, while adding only \(\sim\)0.2 percentage points to the answer flip rate on MCR-Bench. For maximum precision, our default \techname configuration applies full compensation across all layers.

\begin{figure}[t!]
\begin{center}
\includegraphics[width=0.48\columnwidth]{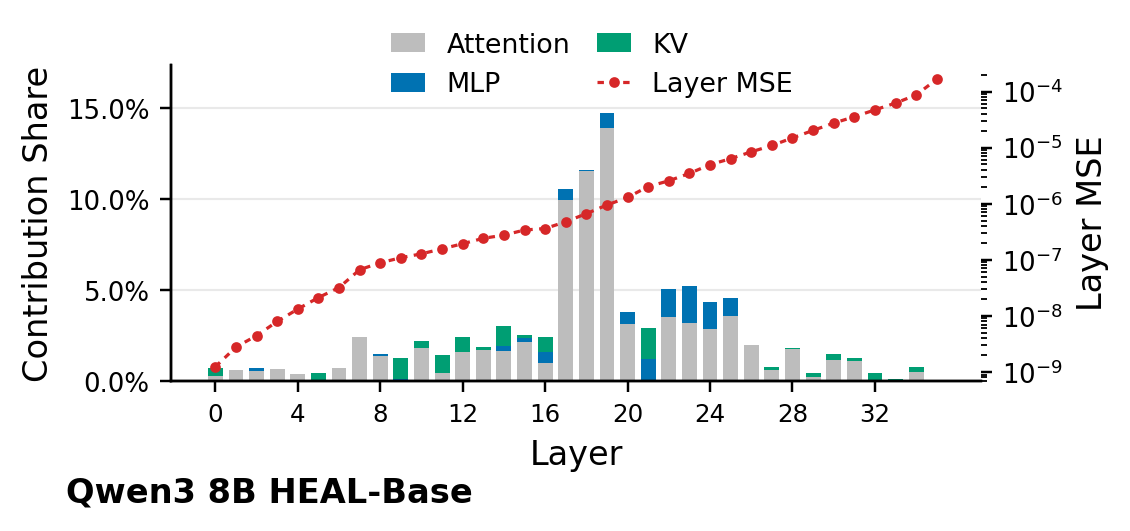}
\includegraphics[width=0.48\columnwidth]{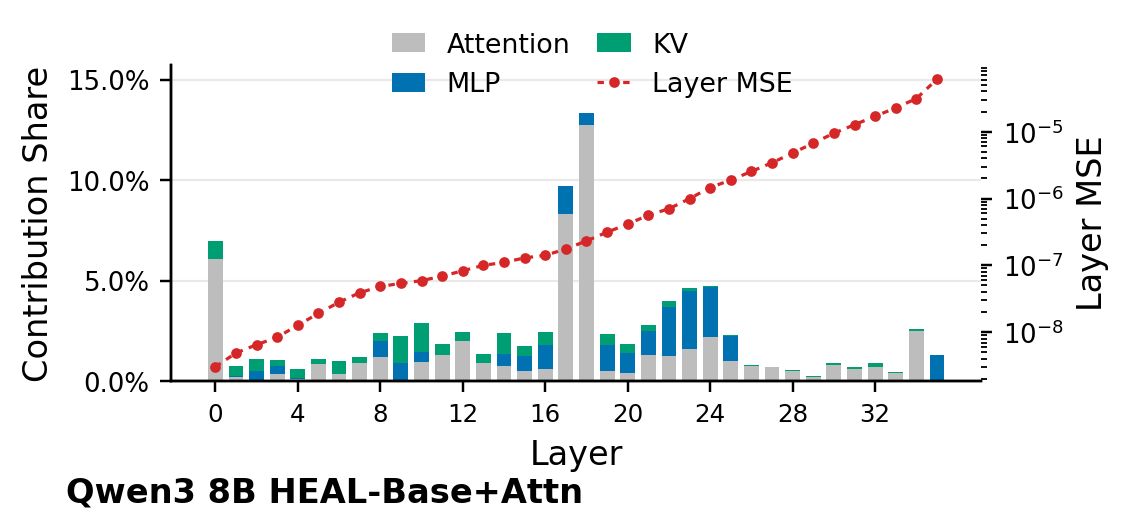}
\includegraphics[width=0.48\columnwidth]{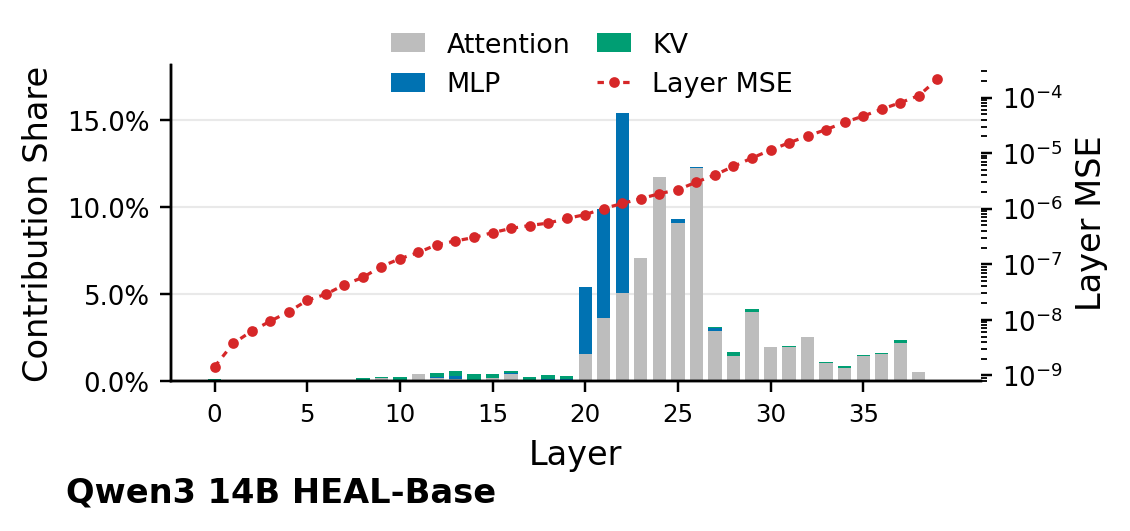}
\includegraphics[width=0.48\columnwidth]{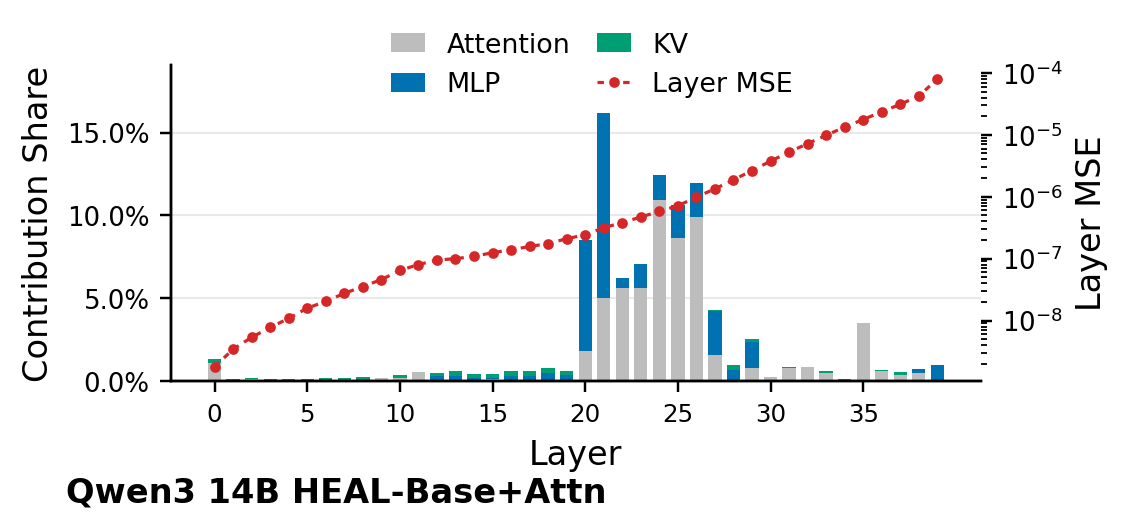}
\includegraphics[width=0.48\columnwidth]{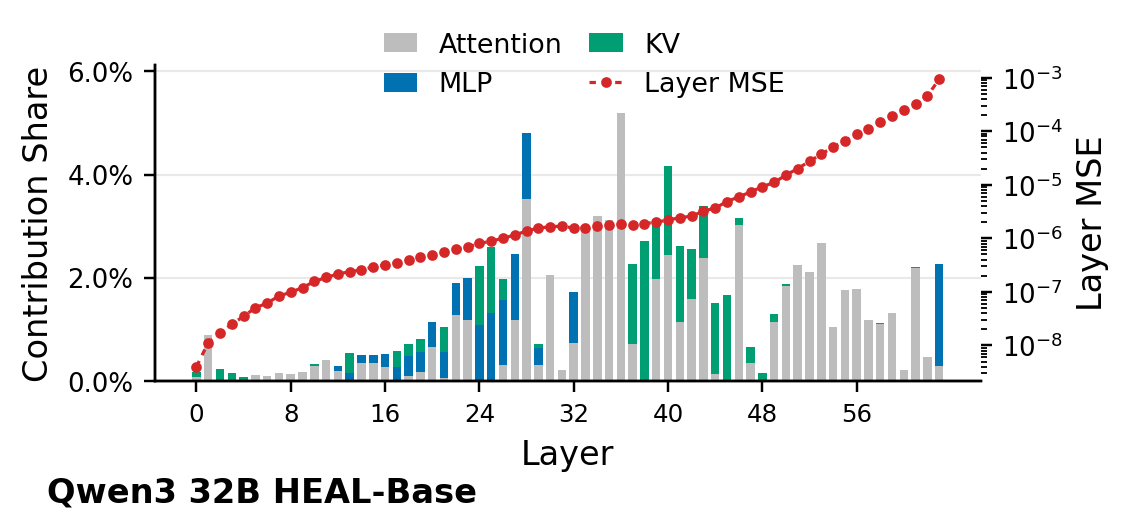}
\includegraphics[width=0.48\columnwidth]{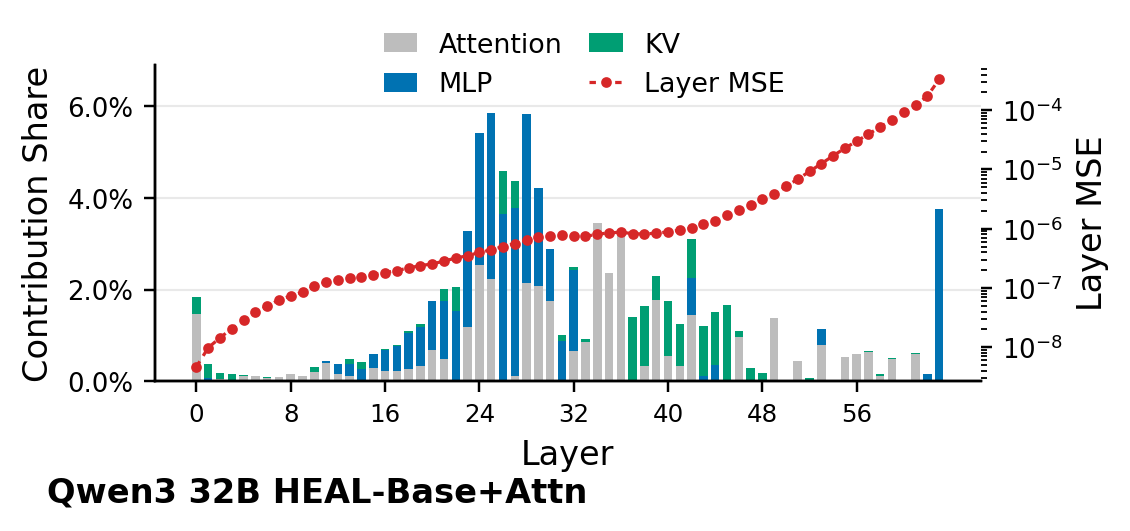}
\includegraphics[width=0.48\columnwidth]{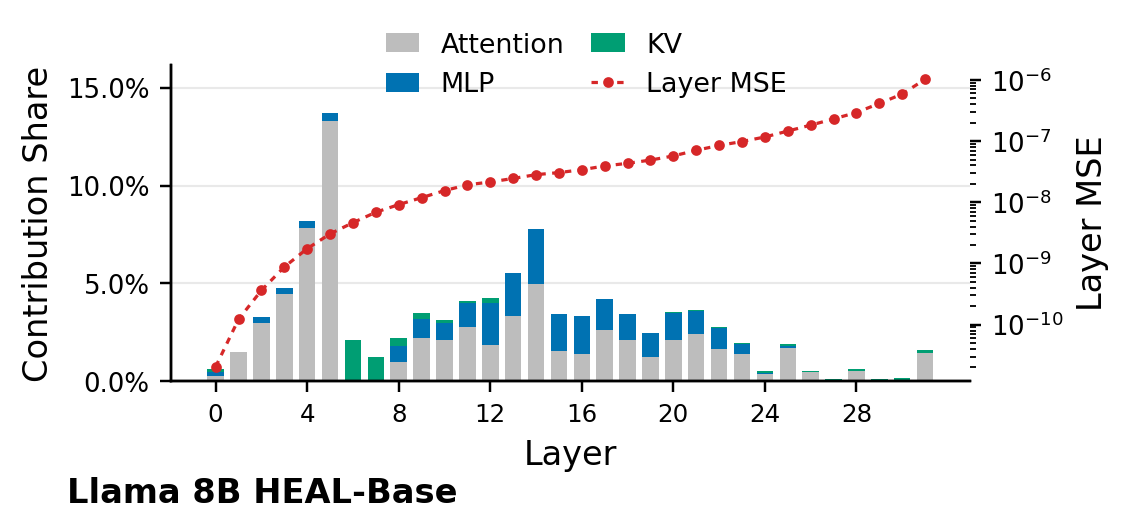}
\includegraphics[width=0.48\columnwidth]{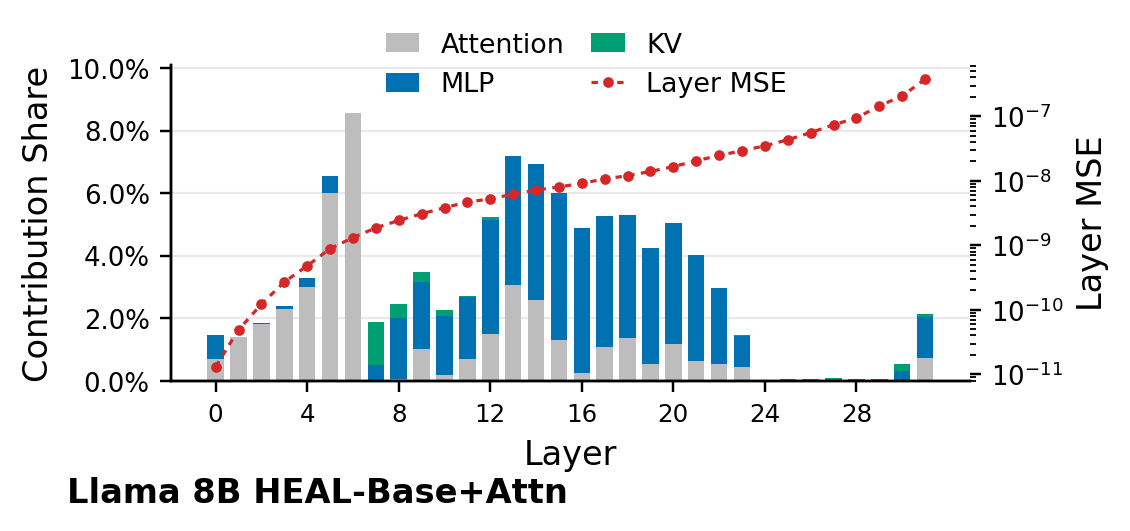}
\includegraphics[width=0.48\columnwidth]{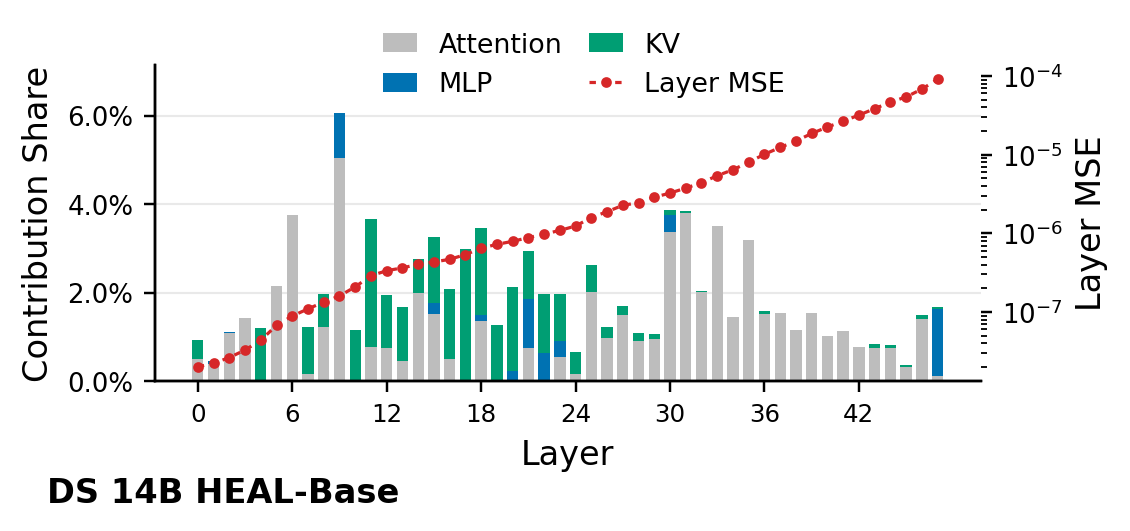}
\includegraphics[width=0.48\columnwidth]{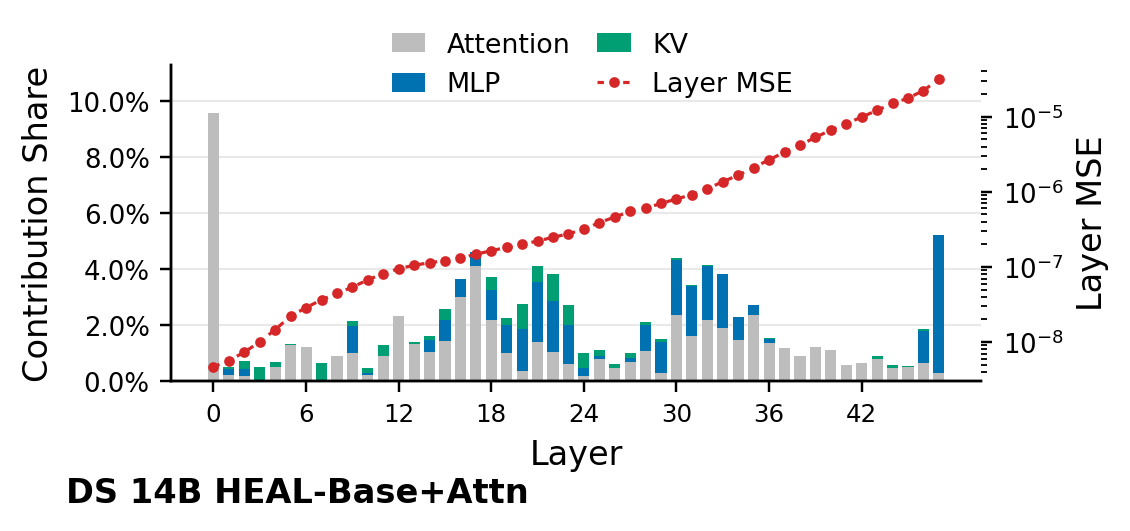}
\caption{Error-contribution maps for \techname{}-Base and \techname{}-Base+Attn across all evaluated models.}
\label{app:fig:heal-variant-contribution}
\end{center}
\end{figure}

\MyPara{Contribution Diagnostics under \techname{} Variants.} We also report the error-contribution maps for \techname{}-Base and \techname{}-Base+Attn in Figure~\ref{app:fig:heal-variant-contribution}. These plots illustrate the staged optimization path of \techname{}: after \techname{}-Base broadly reduces residual-boundary truncation, the remaining error becomes concentrated in attention and KV-cache propagation, motivating the Q/K/V protection in \techname{}-Base+Attn; after \techname{}-Base+Attn protects the attention path, the relative MLP/GEMM contribution becomes more visible, motivating the GEMM compensation in \techname.

\section{Additional Performance Results}
\label{app:performance} \label{app:perf}

\MyPara{Per-Model Breakdown.} In addition to the serving setup described in Appendix~\ref{app:setup}, each GPU--model pair is evaluated over three independent runs. Tables~\ref{tab:sharegpt_perf_qwen3_8b}, \ref{tab:sharegpt_perf_qwen3_32b}, \ref{tab:sharegpt_perf_llama_8b}, and~\ref{tab:sharegpt_perf_deepseek_14b} report the mean TPOT alongside the standard error of the mean for the remaining models.

\begin{table}[ht]
\centering
\caption{ShareGPT TPOT and Memory Usage for the \textbf{Qwen3-8B} model.}
\label{tab:sharegpt_perf_qwen3_8b}
\resizebox{\textwidth}{!}{
\begin{tabular}{l|c|c|c|c}
\toprule
Configuration & A100 TPOT (ms) & H100 TPOT (ms) & MI300X TPOT (ms) & Memory Usage (GiB) \\
\midrule
BF16                 & $62.4\pm0.6$ & $25.2\pm0.1$ & $33.6\pm0.1$ & $15.8$ \\
TF32 (+LayerCast)    & $147.9\pm0.5$ & $67.3\pm0.0$ & $74.9\pm0.2$ & $16.6$ \\
FP32 (+LayerCast)    & $747.5\pm4.5$ & $304.2\pm1.4$ & $139.7\pm0.2$ & $16.6$ \\
\midrule
\techname{}-Base           & $70.9\pm0.2$ & $29.5\pm0.1$ & $36.4\pm0.0$ & $16.0$ \\
\techname{}-Base+Attn      & $78.4\pm0.2$ & $34.3\pm0.1$ & $50.4\pm0.1$ & $16.0$ \\
\techname{}-Base+Attn+Skip & $76.2\pm0.1$ & $32.9\pm0.0$ & $46.3\pm0.1$ & $16.0$ \\
\techname                  & $136.1\pm0.1$ & $54.0\pm0.4$ & $69.4\pm0.1$ & $16.2$ \\
\bottomrule
\end{tabular}
}
\end{table}

\begin{table}[ht]
\centering
\caption{ShareGPT TPOT and Memory Usage for the \textbf{Qwen3-32B} model.}
\label{tab:sharegpt_perf_qwen3_32b}
\resizebox{\textwidth}{!}{
\begin{tabular}{l|c|c|c|c}
\toprule
Configuration & A100 TPOT (ms) & H100 TPOT (ms) & MI300X TPOT (ms) & Memory Usage (GiB) \\
\midrule
BF16                 & $119.3\pm0.6$ & $49.1\pm0.2$ & $118.0\pm0.2$ & $61.9$ \\
TF32 (+LayerCast)    & $268.1\pm3.5$ & $149.2\pm0.4$ & $289.7\pm0.8$ & $64.0$ \\
FP32 (+LayerCast)    & $617.0\pm3.5$ & $338.2\pm0.5$ & $555.4\pm1.0$ & $64.0$ \\
\midrule
\techname{}-Base           & $117.7\pm0.4$ & $51.7\pm0.2$ & $126.0\pm0.2$ & $62.3$ \\
\techname{}-Base+Attn      & $126.3\pm0.8$ & $56.9\pm0.2$ & $162.2\pm0.7$ & $62.4$ \\
\techname{}-Base+Attn+Skip & $121.9\pm1.5$ & $55.7\pm0.4$ & $154.8\pm0.2$ & $62.3$ \\
\techname                  & $205.8\pm1.4$ & $83.5\pm1.7$ & $255.8\pm0.5$ & $62.7$ \\
\bottomrule
\end{tabular}
}
\end{table}

\begin{table}[ht]
\centering
\caption{ShareGPT TPOT and Memory Usage for the \textbf{Llama-8B} model.}
\label{tab:sharegpt_perf_llama_8b}
\resizebox{\textwidth}{!}{
\begin{tabular}{l|c|c|c|c}
\toprule
Configuration & A100 TPOT (ms) & H100 TPOT (ms) & MI300X TPOT (ms) & Memory Usage (GiB) \\
\midrule
BF16                 & $63.8\pm0.3$ & $25.0\pm0.0$ & $32.8\pm0.1$ & $15.5$ \\
TF32 (+LayerCast)    & $149.1\pm0.5$ & $66.4\pm0.0$ & $74.0\pm0.1$ & $16.4$ \\
FP32 (+LayerCast)    & $752.9\pm1.4$ & $311.2\pm0.5$ & $137.6\pm0.3$ & $16.4$ \\
\midrule
\techname{}-Base           & $69.7\pm0.4$ & $28.6\pm0.1$ & $35.5\pm0.1$ & $15.6$ \\
\techname{}-Base+Attn      & $77.2\pm0.3$ & $32.7\pm0.1$ & $48.4\pm0.2$ & $15.6$ \\
\techname{}-Base+Attn+Skip & $75.5\pm0.3$ & $32.3\pm0.4$ & $46.4\pm0.1$ & $15.6$ \\
\techname                  & $136.3\pm0.7$ & $52.6\pm0.2$ & $68.6\pm0.2$ & $15.8$ \\
\bottomrule
\end{tabular}
}
\end{table}

\begin{table}[ht]
\centering
\caption{ShareGPT TPOT and Memory Usage for the \textbf{DS-14B} model.}
\label{tab:sharegpt_perf_deepseek_14b}
\resizebox{\textwidth}{!}{
\begin{tabular}{l|c|c|c|c}
\toprule
Configuration & A100 TPOT (ms) & H100 TPOT (ms) & MI300X TPOT (ms) & Memory Usage (GiB) \\
\midrule
BF16                 & $115.1\pm0.7$ & $44.6\pm0.1$ & $59.6\pm0.1$ & $28.2$ \\
TF32 (+LayerCast)    & $270.0\pm1.1$ & $120.4\pm0.0$ & $137.1\pm0.4$ & $29.4$ \\
FP32 (+LayerCast)    & $1405.1\pm4.2$ & $580.6\pm1.6$ & $258.5\pm0.2$ & $29.4$ \\
\midrule
\techname{}-Base           & $124.3\pm0.2$ & $50.2\pm0.2$ & $64.1\pm0.2$ & $28.4$ \\
\techname{}-Base+Attn      & $137.1\pm0.7$ & $57.2\pm0.3$ & $86.2\pm0.2$ & $28.4$ \\
\techname{}-Base+Attn+Skip & $132.9\pm0.7$ & $54.9\pm0.0$ & $79.4\pm0.2$ & $28.4$ \\
\techname                  & $246.2\pm0.4$ & $96.9\pm0.4$ & $124.8\pm0.4$ & $28.7$ \\
\bottomrule
\end{tabular}
}
\end{table}

\MyPara{Cost--Reproducibility Trade-off.} Table~\ref{tab:tradeoff} summarizes the overall cost--reproducibility trade-off across all five models and three GPUs. The answer flip rate is averaged across the five models (from Table~\ref{tab:mse_flip_rate}), while TPOT and base memory footprint are normalized to the native BF16 baseline on the same model and GPU (summarizing Table~\ref{tab:sharegpt_perf_14b} and Tables~\ref{tab:sharegpt_perf_qwen3_8b}--\ref{tab:sharegpt_perf_deepseek_14b}). All \techname configurations remain close to the native BF16 memory footprint: model weights and runtime workspace grow by only 0.6--1.3\% for the ablation variants and 1.3--2.5\% for \techname (compared with 3.4--5.8\% for the 32-bit LayerCast baselines), while the grouped INT16 KV codec adds only 1.6\% storage overhead over native 16-bit caches (Appendix~\ref{app:impl}).

\begin{table}[ht]
\centering
\small
\caption{Cost--reproducibility trade-off across five models and three GPUs. Brackets indicate the range across models (for flip rate and memory) or across all 15 model--GPU pairs (for TPOT).}
\label{tab:tradeoff}
\resizebox{\textwidth}{!}{
\begin{tabular}{l|c|c|c|l}
\toprule
Configuration & Flip Rate & TPOT vs.\ BF16 & Memory vs.\ BF16 & Role \\
\midrule
BF16 & 11.47\% [8.03\%, 16.53\%] & 1.00$\times$ & --- & Native baseline \\
TF32 (+LayerCast) & 2.64\% [1.97\%, 3.77\%] & 2.47$\times$ [2.23$\times$, 3.04$\times$] & +3.4--5.8\% & 32-bit baseline \\
FP32 (+LayerCast) & 0.03\% [0.00\%, 0.11\%] & 8.85$\times$ [4.16$\times$, 13.11$\times$] & +3.4--5.8\% & High-precision reference \\
\midrule
\techname{}-Base & 1.77\% [1.22\%, 2.39\%] & 1.10$\times$ [0.99$\times$, 1.17$\times$] & +0.6--1.3\% & Ablation \\
\techname{}-Base+Attn & 1.02\% [0.66\%, 1.25\%] & 1.30$\times$ [1.06$\times$, 1.50$\times$] & +0.6--1.3\% & Ablation \\
\techname{}-Base+Attn+Skip & 1.15\% [0.84\%, 1.48\%] & 1.25$\times$ [1.02$\times$, 1.42$\times$] & +0.6--1.3\% & Ablation \\
\textbf{\techname} & \textbf{0.07\% [0.06\%, 0.10\%]} & \textbf{2.08$\times$ [1.70$\times$, 2.20$\times$]} & \textbf{+1.3--2.5\%} & \textbf{Recommended} \\
\bottomrule
\end{tabular}
}
\end{table}

\MyPara{Deployment Guidance.} For mission-critical deployments requiring strict cross-hardware reproducibility, we recommend the complete \techname pipeline, which achieves FP32-level flip rates (0.07\%) at 1.70--2.20$\times$ the BF16 TPOT (versus 4.16--13.11$\times$ for full FP32). For latency-constrained deployments that can tolerate a $\sim$1\% residual flip rate, \techname{}-Base+Attn and \techname{}-Base+Attn+Skip reduce the BF16 flip rate by $\sim$10$\times$ at $\le$1.50$\times$ the BF16 TPOT, while \techname{}-Base alone reduces flips by $>6\times$ at $\le$1.17$\times$ TPOT. The fine-grained channel policy in Appendix~\ref{app:impl} further provides intermediate operating points between these lightweight variants and full \techname.

\section{Detailed Description of MCR-Bench}
\label{app:benchmark}

% MCR-Bench is constructed from high-quality datasets across three critical domains, curated to emphasize multi-step reasoning over simple recall: i) Medicine: We curate the medical subset from MedQA-USMLE-4-options \cite{jin2021disease}, CareQA \cite{hpai2023careqa}, and PubMedQA \cite{jin2019pubmedqa}. We combine these to cover complementary forms of clinically meaningful reasoning, filtering for questions that require reasoning over symptoms, lab findings, and imaging. The subset covers \textbf{emergency diagnosis, sepsis identification, and medical ethics}, where model errors could lead to unsafe recommendations. ii) Law: The legal subset includes CaseHOLD \cite{zhengguha2021}, LegalBench \cite{guha2023legalbench}, BarExamQA \cite{reglab2020barexam}, and LexGLUE \cite{chalkidis-etal-2022-lexglue}. Use cases include \textbf{contract risk review, civil rights analysis, and statutory interpretation}. These are mission-critical because reasoning errors can materially alter conclusions regarding rights, liabilities, or access to legal remedies. iii) Finance: Constructed from Financial Sentiment Analysis \cite{malo2014good}, FinQA \cite{chen2021finqa}, and FinClaim \cite{FinClaimRadar2025sksayan01}. We capture use cases such as \textbf{claim credibility, market sentiment monitoring, and numerical reasoning over tables}. These target settings where outputs influence investment analysis and compliance review, where errors often arise from misreading disclosures or misclassifying claim credibility under uncertainty.

% --------------------

Within MCR-Bench, we curate 4500 test cases drawn from 10 established datasets: 1500 each from the medical, legal, and financial domains, covering more than 20 use cases. We select tasks that are consequential, where an inconsistent answer carries real cost, and objectively evaluable, with a discrete, parseable answer so that a change in output is a well-defined event. Questions are sampled with a fixed seed from the evaluation pools of the constituent datasets, without model-specific filtering. To evaluate answer flip rates, each configuration is executed once on each GPU with greedy decoding and up to 4096 generated tokens. An answer flip is counted whenever the parsed final answer differs from that of the reference FP32 execution, with parse failures and length truncations treated as a distinct answer. Specifically, the Final Hidden MSE is measured on a 100-question subset of MCR-Bench, where the token-averaged error already stabilizes.

\MyPara{Medical Domain.}
We construct the 1500-question medical subset from MedQA-USMLE-4-options \cite{jin2021disease}, CareQA \cite{hpai2025careqa}, and PubMedQA \cite{jin2019pubmedqa} to cover complementary forms of clinically meaningful reasoning. We used questions that require multi-step reasoning over symptoms, laboratory findings, imaging, treatment options, or study conclusions, and excluded items primarily based on isolated factual recall. The resulting subset spans 9 mission-critical use cases: \textbf{emergency diagnosis, critical care management, sepsis and severe infection identification, cardiovascular emergencies, pediatric emergency recognition, toxicology, adverse drug reactions, procedural decision-making, and medical ethics in emergencies}. These settings are particularly important for evaluating robustness because model errors may lead not only to incorrect answers, but also to unsafe recommendations that threaten patient safety.

\MyPara{Legal Domain.}
The 1500-question legal subset is drawn from CaseHOLD \cite{zheng2021does}, LegalBench (insurance policy interpretation) \cite{guha2023legalbench}, BarExamQA \cite{zheng2025reasoning}, and LexGLUE (UNFAIR-ToS) \cite{chalkidis-etal-2022-lexglue}, covering 6 practical legal reasoning use cases: \textbf{contract risk review, civil rights analysis, criminal procedure, information protection, statutory interpretation, and insurance coverage analysis}. These use cases are mission-critical because small reasoning perturbations can materially alter conclusions regarding rights, obligations, liability, or access to remedies. They also stress reasoning patterns that are especially sensitive to numerical instability, including close semantic distinctions, rule application under ambiguity, and conflicts between superficially plausible choices.

\MyPara{Finance Domain.}
The 1500-question finance subset is constructed from Financial Sentiment Analysis \cite{maia2018www,malo2014good, malo2014good}, FinQA \cite{chen2021finqa}, and FinClaim \cite{FinClaimRadar2025sksayan01}, organized into 5 use cases: \textbf{financial claim credibility, market sentiment monitoring, numerical reasoning over tables, risk and exposure identification, and classification of financially meaningful language}. These tasks target high-stakes settings where model outputs inform investment analysis, risk monitoring, compliance review, or financial decision support, and where subtle instability can cause models to miscalculate figures, misread disclosures, confuse speculation with evidence, or misclassify claim credibility under uncertainty.

% \parabf{Models and Benchmarks.} We select three representative models for our analysis: Qwen3-8B~\cite{yang2025qwen3}, Llama-3.1-8B-Instruct~\cite{grattafiori2024llama}, and DeepSeek-R1-Distill-Qwen-7B~\cite{guo2025deepseek}. These models were chosen for their widespread industry adoption and their common usage of Grouped Query Attention (GQA)~\todoinline{cite}, the standard Attention mechanism for current LLM, to guarantee the consistency of our discovery. We also standardize the context length to 4096 tokens across all models. 

% Our evaluation employs three diverse datasets covering both prefill and generation phases. For the prefill-only task, we use WikiText-2~\cite{merity2016wikitext2} to measure language modeling proficiency via perplexity (PPL). For generative tasks, we utilize GSM8K~\cite{cobbe2021gsm8k} and IFEval~\cite{zhou2023ifeval} to evaluate the models' capabilities in mathematical reasoning and instruction following, respectively.

% \todoinline{batch sizes, eager/non-eager, Attention backend, things we utilized to create different execution paths.}

% \todoinline{how we measure and approximate the logprobs deviation}

\section{Existing Assets Used}
\label{app:asset}

In accordance with the NeurIPS guidelines for responsible research, we disclose all existing models, software libraries, and datasets utilized in this work. We acknowledge the original creators and adhere to their respective licenses and terms of use.

\subsection{Models}
\begin{itemize}
    \item \textbf{Qwen3 (8B, 14B, 32B)}~\cite{yang2025qwen3}: Developed by Alibaba Cloud. The model weights and architecture are utilized for our numerical divergence analysis. We additionally use the W4A16 checkpoints \texttt{Qwen/Qwen3-\{8B,14B,32B\}-AWQ} (AWQ~\cite{lin2024awq}, released by Qwen) and \texttt{RedHatAI/Qwen3-\{8B,14B,32B\}-quantized.w4a16} (GPTQ~\cite{frantar2023gptq}). \textit{License: Apache 2.0}. URL: \url{https://huggingface.co/collections/Qwen/qwen3}
    \item \textbf{Llama-3.1-8B-Instruct}~\cite{grattafiori2024llama}: Developed by Meta. Utilized for zero-shot and reasoning evaluations. We additionally use the W4A16 checkpoints \texttt{hugging-quants/Meta-Llama-3.1-8B-Instruct-AWQ-INT4} (AWQ) and \texttt{RedHatAI/Meta-Llama-3.1-8B-Instruct-quantized.w4a16} (GPTQ). \textit{License: Llama 3.1 Community License}. URL: \url{https://huggingface.co/meta-llama/Llama-3.1-8B-Instruct}
    \item \textbf{DeepSeek-R1-Distill-Qwen-14B}~\cite{guo2025deepseek}: Developed by DeepSeek. Utilized to evaluate the impact of distillation and reinforcement learning on numerical stability. We additionally use the W4A16 checkpoints \texttt{casperhansen/deepseek-r1-distill-qwen-14b-awq} (AWQ) and \texttt{RedHatAI/DeepSeek-R1-Distill-Qwen-14B-quantized.w4a16} (GPTQ). \textit{License: MIT License}. URL: \url{https://huggingface.co/deepseek-ai/DeepSeek-R1-Distill-Qwen-14B}
\end{itemize}

\subsection{Software and Frameworks}
\begin{itemize}
    \item \textbf{vLLM (v0.19.0)}~\cite{kwon2023vllm}: An open-source high-throughput inference engine serving as the foundation for our experimental testbed. \textit{License: Apache 2.0}. URL: \url{https://github.com/vllm-project/vllm}
    \item \textbf{PyTorch}~\cite{paszke2019pytorch}: Deep learning framework underlying vLLM. \textit{License: BSD 3-Clause}. URL: \url{https://github.com/pytorch/pytorch}
    \item \textbf{FlashAttention-3}~\cite{shah2024flashattention}: Used as the highly optimized exact Attention backend. \textit{License: BSD 3-Clause}. URL: \url{https://github.com/Dao-AILab/flash-attention}
    \item \textbf{Triton}~\cite{tillet2019triton}: An open-source programming language for writing highly efficient custom GPU kernels. \textit{License: MIT License}. URL: \url{https://github.com/triton-lang/triton}
    \item \textbf{CUDA Toolkit (12.8)} \& \textbf{cuBLASLt}~\cite{cublaslt}: Proprietary libraries developed by NVIDIA for GPU acceleration. \textit{License: NVIDIA Software License Agreement}. URL: \url{https://developer.nvidia.com/cuda-toolkit}
    \item \textbf{ROCm (7.2.1)} \& \textbf{hipBLASLt}~\cite{hipBLASLt}: Open-source software stack and BLAS library for AMD GPUs. \textit{License: MIT License}. URL: \url{https://github.com/ROCm/rocm}
    \item \textbf{LM Evaluation Harness (v0.4.9)}~\cite{gao2024lmeval}: Used for WikiText-2, MMLU-Redux, IFEval, and RULER in the model quality evaluation. \textit{License: MIT License}. URL: \url{https://github.com/EleutherAI/lm-evaluation-harness}
    \item \textbf{LightEval (v0.10.0)}~\cite{habib2023lighteval}: Used for MATH-500 and LiveCodeBench in the model quality evaluation. \textit{License: MIT License}. URL: \url{https://github.com/huggingface/lighteval}
\end{itemize}

\subsection{Datasets}
We utilized the following publicly available datasets for system benchmarking and to construct the Mission-Critical Reproducibility Benchmark (MCR-Bench). 

\paragraph{System Profiling Datasets:}
\begin{itemize}
    \item \textbf{ShareGPT V3}~\cite{sharegpt_v3}: Used for evaluating Time Per Output Token~(TPOT) and peak memory overhead. \textit{License: Apache 2.0}. URL: \url{https://huggingface.co/datasets/anon8231489123/ShareGPT_Vicuna_unfiltered}
    \item \textbf{C4 Dataset}~\cite{2019t5}: A cleaned version of Common Crawl used to profile the magnitude of structural outliers in intermediate activations. \textit{License: Apache 2.0}. URL: \url{https://www.tensorflow.org/datasets/catalog/c4}
\end{itemize}

\paragraph{MCR-Bench Datasets (Medical Domain):}
\begin{itemize}
    \item \textbf{MedQA}~\cite{jin2021disease}: USMLE multiple-choice questions. \textit{License: MIT License (original release); CC BY 4.0 (Hugging Face release)}. URL: \url{https://huggingface.co/datasets/GBaker/MedQA-USMLE-4-options}
    \item \textbf{CareQA}~\cite{hpai2025careqa}: Healthcare evaluation benchmark. \textit{License: Apache 2.0}. URL: \url{https://huggingface.co/datasets/HPAI-BSC/CareQA}
    \item \textbf{PubMedQA}~\cite{jin2019pubmedqa}: Biomedical research question answering. \textit{License: MIT License}. URL: \url{https://github.com/pubmedqa/pubmedqa}
\end{itemize}

\paragraph{MCR-Bench Datasets (Legal Domain):}
\begin{itemize}
    \item \textbf{CaseHOLD}~\cite{zheng2021does}: Multiple-choice legal holdings. \textit{License: Apache 2.0}. URL: \url{https://github.com/reglab/casehold}
    \item \textbf{LegalBench}~\cite{guha2023legalbench}: Legal reasoning benchmark. \textit{License: CC BY 4.0}. URL: \url{https://huggingface.co/datasets/nguha/legalbench}
    \item \textbf{BarExamQA}~\cite{zheng2025reasoning}: Reasoning-focused legal retrieval benchmark. \textit{License: CC BY-SA 4.0}. URL: \url{https://huggingface.co/datasets/reglab/barexam_qa}
    \item \textbf{LexGLUE}~\cite{chalkidis-etal-2022-lexglue}: Legal language understanding benchmark; we use the UNFAIR-ToS and CaseHOLD tasks. \textit{License: CC BY 4.0}. URL: \url{https://huggingface.co/datasets/coastalcph/lex_glue}
\end{itemize}

\paragraph{MCR-Bench Datasets (Finance Domain):}
\begin{itemize}
    \item \textbf{Financial Sentiment Analysis}~\cite{malo2014good, maia2018www}: Financial text classification combining Financial PhraseBank and the FiQA sentiment task. \textit{License: CC0 (combined release); CC BY-NC-SA 3.0 (Financial PhraseBank); not specified (FiQA)}. URL: \url{https://www.kaggle.com/datasets/sbhatti/financial-sentiment-analysis}
    \item \textbf{FinQA}~\cite{chen2021finqa}: Numerical reasoning over financial data. \textit{License: MIT License}. URL: \url{https://github.com/czyssrs/FinQA}
    \item \textbf{FinClaim}~\cite{FinClaimRadar2025sksayan01}: Evidentiary reasoning of financial claims. \textit{License: CC BY-SA 4.0}. URL: \url{https://huggingface.co/datasets/ImBinarry/FinClaimRadar-Financial-Reasoning}
\end{itemize}

\paragraph{Model Quality Benchmarks:}
\begin{itemize}
    \item \textbf{WikiText-2}~\cite{merity2016wikitext2}: Language modeling perplexity. \textit{License: CC BY-SA}. URL: \url{https://huggingface.co/datasets/EleutherAI/wikitext_document_level}
    \item \textbf{MMLU-Redux 2.0}~\cite{gema2025mmluredux}: General knowledge. \textit{License: CC BY 4.0}. URL: \url{https://huggingface.co/datasets/edinburgh-dawg/mmlu-redux-2.0}
    \item \textbf{IFEval}~\cite{zhou2023ifeval}: Instruction following. \textit{License: Apache 2.0}. URL: \url{https://huggingface.co/datasets/google/IFEval}
    \item \textbf{MATH-500}~\cite{lightman2024verify}: Mathematical reasoning. \textit{License: MIT License (PRM800K and MATH)}. URL: \url{https://huggingface.co/datasets/HuggingFaceH4/MATH-500}
    \item \textbf{LiveCodeBench}~\cite{jain2025livecodebench}: Code generation (v5). \textit{License: MIT License}. URL: \url{https://huggingface.co/datasets/livecodebench/code_generation_lite}
    \item \textbf{MMMLU}~\cite{openai2024mmmlu} and \textbf{MMLU}~\cite{hendrycks2021measuring}: Multilingual and English knowledge. \textit{License: MIT License}. URL: \url{https://huggingface.co/datasets/openai/MMMLU}, \url{https://huggingface.co/datasets/cais/mmlu}
    \item \textbf{RULER}~\cite{hsieh2024ruler}: Long-context tasks, instantiated with HotpotQA~\cite{yang2018hotpotqa} (\textit{License: CC BY-SA 4.0}) and Paul Graham essays as filler text (\textit{license not specified}). URL: \url{https://github.com/NVIDIA/RULER}, \url{https://huggingface.co/datasets/hotpotqa/hotpot_qa}
\end{itemize}

\section{Limitations}
\label{sec:limit}

While \techname substantially improves cross-hardware reproducibility, we note the following limitations:

\begin{itemize}
\item \textbf{Hardware Dependence:} The performance efficiency of our approach relies on flexible, hardware-optimized GEMM implementations.
\item \textbf{Functional, not Bit-wise:} \techname bounds numerical errors to substantially reduce the probability of output divergence, but it does not provide strict bit-to-bit determinism.
\item \textbf{Model-Dependent Calibration:} The layer-skipping policy of \techname{}-Base+Attn+Skip and the per-channel KV-cache normalization statistics of \techname are calibrated per model, making the optimization model-dependent.
\item \textbf{Performance Trade-offs:} While significantly more efficient than global FP32 upcasting, \techname still incurs a 70--120\% TPOT overhead (about 2.1$\times$ the BF16 TPOT on average) relative to the native BF16 baseline.
\item \textbf{FP8 Formats:} Our current implementation targets 16-bit formats, as native FP8 execution discards too much precision at kernel boundaries to achieve \techname-level reproducibility alone. Extending \techname to FP8 would introduce higher overhead.
\end{itemize}

\section{Broader Impacts}
\label{app:impact}

This paper addresses the critical challenge of non-reproducibility in LLM inference, where heterogeneous hardware and dynamic software configurations often lead to divergent outputs even under supposedly deterministic settings like greedy decoding. Our proposed solution ensures consistent model behavior across diverse infrastructure and multiple runs. This work significantly enhances the reliability of LLMs for mission-critical applications such as medical diagnosis, legal reasoning, and production workflow automation, where reproducibility is a fundamental requirement. Consequently, this research fosters greater trust in LLM systems deployed in complex, real-world datacenter environments.

\section{Acknowledgment of LLM Usage}
\label{app:llm-ack}

We used OpenAI Codex, Claude Code, and Google Gemini as assistive tools during the preparation of this work. OpenAI Codex and Claude Code were used to assist with writing experiment scripts and running evaluations. Codex was also used to assist with implementing the system solution, while Claude Code was additionally used to draft parts of this manuscript. Google Gemini was used for copyediting and refining to improve clarity and readability. All LLM-generated content was reviewed and verified by the authors, who take full responsibility for the final manuscript.

%%%%%%%%%%%%%%%%%%%%%%%%%%%%%%%%%%%%%%%%%%%%%%%%%%%%%%%%%%%%

% \clearpage
% \input{checklist.tex}

\end{document}